\documentclass[letterpaper]{article} 
\usepackage{amsmath,amssymb,mathtools,bm,empheq,amsthm}

\theoremstyle{plain}

\theoremstyle{definition}

\theoremstyle{remark}

\def\bal#1\eal{\begin{align}#1\end{align}} 
\def\suml{\sum\limits}

\newcommand{\pr}[1]{\left(#1\right)} 
\newcommand{\cbr}[1]{\left\{#1\right\}} 
\DeclareMathOperator*{\argmin}{arg\,min} 
\DeclareMathOperator*{\argmax}{arg\,max} 

\def\transp{\mathsf{T}} 
\def\m{\mathbf}
\def\mc{\mathcal}
\def\R{\mathbb{R}}

\newcommand{\norm}[2]{\ensuremath{\left\|#1\right\|_{#2}}}

\newcommand {\bbmtx}{\begin{bmatrix}} 
\newcommand {\ebmtx}{\end{bmatrix}} 

\DeclareMathOperator*{\vctr}{vec} 
\newcommand{\vc}[1]{\vctr{\left(#1\right)}}

\DeclareMathOperator*{\trace}{tr} 
\newcommand{\tr}[1]{\trace\pr{#1}}

\usepackage[draft]{paper_format}  
\usepackage{times}  
\usepackage{helvet}  
\usepackage{courier}  
\usepackage[hyphens]{url}  
\usepackage{graphicx} 
\usepackage{natbib}  
\usepackage{caption} 
\usepackage{adjustbox}
\usepackage{algorithm}
\usepackage{algorithmic}
\usepackage{amsmath}
\usepackage{booktabs}
\usepackage{cuted}
\usepackage{adjustbox}
\usepackage{newfloat}
\usepackage{listings}
\usepackage{array}
\usepackage{arydshln}
\usepackage{pgfplots}
\usepackage{pgfplotstable}
\usepackage{multirow}
\pgfplotsset{compat=1.18}
\usepackage{pifont}
\usepackage{tikz}
\usetikzlibrary{plotmarks}
\pgfplotsset{compat=1.17}
\pgfplotsset{compat=1.17}
\usetikzlibrary{pgfplots.groupplots}
\usepgfplotslibrary{groupplots}
\DeclareCaptionStyle{ruled}{labelfont=normalfont,labelsep=colon,strut=off} 
\floatstyle{ruled}
\newfloat{listing}{tb}{lst}{}
\floatname{listing}{Listing}
\title{\hspace{1ex}\includegraphics[width=.5cm]{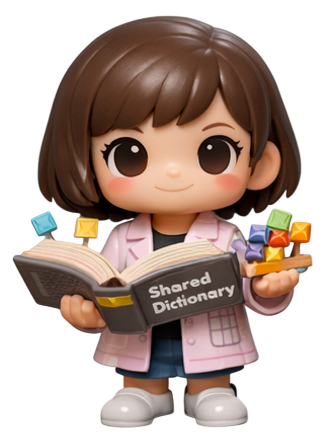} Share Your Attention: Transformer Weight Sharing via Matrix-based Dictionary Learning}

\author{%
  Magauiya Zhussip$^{1}$, Dmitriy Shopkhoev$^{1,2}$, Ammar Ali$^{1,2}$, \\
  Stamatios Lefkimmiatis$^{1}$ \\ 
  \vspace{4mm}
  $^1$MTS AI ,
  $^2$ITMO University
}
\begin{document}

\maketitle
$\def\thefootnote{*}\footnotetext{This work was accepted and presented at AAAI 2026}$
\begin{abstract}
Large language models (LLMs) have revolutionized AI applications, yet their high computational and memory demands hinder their widespread deployment. Existing compression techniques focus on intra-block optimizations (e.g., low-rank approximation or attention head pruning), while the repetitive layered structure of transformers implies significant inter-block redundancy—a dimension largely unexplored beyond key-value (KV) caching. Inspired by dictionary learning in convolutional networks, we propose a framework for structured weight sharing across transformer layers. Our approach decomposes attention projection matrices (Q, K, V, O) into shared dictionary atoms, reducing the attention module’s parameters by 66.7\% (e.g., 226.5M → 75M in a 700M-parameter model) while achieving on-par performance. Unlike complex methods requiring distillation or architectural changes, MASA (Matrix Atom Sharing in Attention) operates as a drop-in replacement—trained with standard optimizers—and represents each layer’s weights as linear combinations of shared matrix atoms. Experiments across scales (100M–700M parameters) show that MASA achieves better benchmark accuracy and perplexity than grouped-query attention (GQA), low-rank baselines and recently proposed Repeat-all-over/Sequential sharing at comparable parameter budgets. Ablation studies confirm robustness to the dictionary size and the efficacy of shared representations in capturing cross-layer statistical regularities. Extending to Vision Transformers (ViT), MASA matches performance metrics on image classification tasks with 66.7\% fewer attention parameters. By combining dictionary learning strategies with transformer efficiency, MASA offers a scalable blueprint for parameter-efficient models without sacrificing performance. Finally, we investigate the possibility of employing MASA on large pretrained models to reduce their number of parameters without experiencing any significant drop in their performance. Code will be available at 
\textbf{https://github.com/mts-ai/MASA}
\end{abstract}
\section{Introduction}
\label{sec:introduction}
Large language models (LLMs) have achieved remarkable capabilities, yet their widespread deployment is hindered by the prohibitive computational and memory demands of transformer architectures. While existing compression techniques predominantly target \textit{intra-block} redundancies through low-rank approximations or attention head pruning~\cite{yu2024effectively,GQA}, a critical dimension remains underexplored: the \textit{inter-block} redundancy inherent in transformers' repetitive layered structure. This overlooked opportunity represents a fundamental inefficiency, as $L$ transformer layers with hidden dimension $d$ require $\mathcal{O}(L\!\cdot\!d^2)$ parameters, with attention alone consuming up to half the parameters in foundational models like LLaMA~\cite{llama} and Mistral~\cite{mistral}.

Recently proposed methods like grouped-query attention (GQA)~\cite{GQA} and QK compression (e.g. LISA~\cite{mu2024cross}) demonstrate the value of parameter reduction but operate within isolated layers or focus on particular projections inside attention module. Meanwhile, emerging approaches exploring cross-layer sharing—such as Repeat-all-over~\cite{MobileLLM} and Sequential parameter~\cite{sequential_sharing} assignment strategies—reveal promising directions but suffer from performance degradation in reasoning tasks~\cite{liao2024beyond}. Crucially, these methods lack a principled framework for capturing the statistical regularities across transformer layers.

Inspired by dictionary learning principles in convolutional networks~\cite{sparse_coding}, we propose \textbf{Matrix Atom Sharing in Attention (MASA)}, a novel framework that systematically exploits inter-block redundancy through structured weight sharing across transformer layers. Unlike prior sharing strategies that either enforce rigid weight tying or require complex distillation procedures, MASA decomposes attention projection matrices (Q, K, V, O) into shared dictionary atoms, enabling each layer's weights to be represented as linear combinations of these atoms. This approach reduces attention module parameters by 66.7\% (e.g., 226.5M → 75M in a 700M-parameter model) while maintaining competitive performance—achieving what previous parameter-sharing methods like Sequential-sharing~\cite{sequential_sharing} and Repeat-all-over Sharing could not: consistent accuracy across diverse benchmarks and on-par (or better) performance than the original Transformer.

In summary, the contributions of this work are:
\begin{enumerate}
    \item \textbf{Theoretical Foundation}: By reframing attention compression as a dictionary learning problem, we establish a principled connection between classical signal processing and transformer efficiency, revealing how shared matrix atoms capture cross-layer statistical regularities and efficiently exploit inter-block redundancies.
    \item \textbf{Parameter Efficiency with Performance Parity}: MASA exceeds the performance of low-rank baselines, GQA, and recent Repeat-all-over/Sequential sharing approaches across language modeling (perplexity), reasoning, and knowledge benchmarks under the same (or higher) compression rate. Moreover, MASA with 66.7\% less parameters in attention can match the performance of vanilla (uncompressed) Transformer for S, M, L sizes.
    \item \textbf{Architectural Simplicity}: Unlike methods requiring distillation, regularization, or architectural modifications (e.g., increasing hidden dimensions), MASA operates as a drop-in replacement trained with standard optimizers—preserving the original training pipeline while eliminating auxiliary components.
    
\end{enumerate}

\begin{figure}[!t]
\begin{center}
\includegraphics[width=0.48\textwidth]{{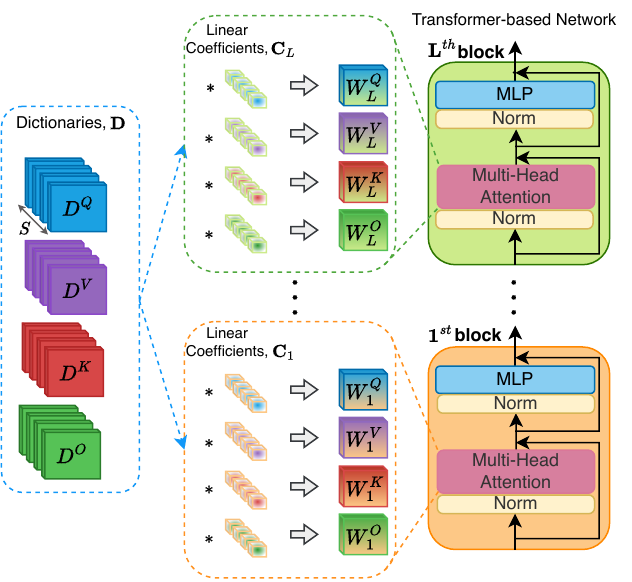}}
\caption{MASA framework: (Left) Independent dictionary pools for Q, K, V, O projections. (Middle) Per-block projection matrices synthesized via weighted combinations of shared dictionaries. All blocks share dictionary pools while using unique linear coefficients for each Transformer block.}
\label{fig:masa_framework}
\end{center}
\end{figure}

Beyond language models, we demonstrate the broad applicability of MASA by extending it to Vision Transformers (ViTs), where it achieves strong performance on image classification tasks while compressing attention modules by 66.7\%. Given the dominance of pretrained models in modern deployment pipelines, we further investigate MASA in training-free adaptation scenarios. Our experiments show that MASA incurs only marginal performance degradation upon parameter pruning, highlighting its robustness and practicality in resource-constrained settings. By unifying dictionary learning with architectural design in Transformers, MASA provides a principled and scalable framework for constructing parameter-efficient models without compromising accuracy. The rest of the paper details our proposed method and presents comprehensive evaluations across model scales (100M--700M parameters) on language and vision tasks. Finally, we conclude with applications to training-free adaptation, highlighting MASA's potential for plug-and-play efficiency in pretrained models.
\section{Related Work}
\label{sec:related_work}

Our work intersects with three primary directions in efficient model design: \textit{structured attention}, \textit{parameter sharing}, and \textit{matrix factorization}. We position the proposed MASA strategy as a unifying and principled advancement that overcomes some of the limitations faced by prior approaches.

\paragraph{Efficient Attention Mechanisms}
The quadratic complexity of self-attention has motivated researchers to discover numerous approximations. For instance, linear attention methods \cite{LA, rwkv} approximate softmax with kernelizable features to achieve linear complexity w.r.t input sequence length. More alternative solutions based on state space models like Mamba \cite{mamba} replace attention with selective recurrence, offering long-context modeling with linear inference. However, these approaches often require pretraining and may show inferior results on tasks requiring global context mixing.

In contrast, MASA preserves the standard attention formulation and instead targets \textit{parameter redundancy} in projection matrices. This ensures compatibility with existing training recipes and pretrained models—a critical advantage for real-world deployment.

\paragraph{Cross-Layer Parameter Sharing}
To reduce inter-block redundancy, several works have explored reusing weight matrices. Weight tying between embedding and output layers is common \cite{press2017using}, and Universal Transformers \cite{dehghaniuniversal} share parameters across time steps. More recently, Repeat-all-over~\cite{MobileLLM} and Sequential-sharing apply deterministic patterns to share attention and FFN weights across layers.

However, such rigid sharing might limit representational flexibility leading to worse performance, particularly in deep models where early and late layers perform distinct functions \cite{liao2024beyond}. Basis Sharing \cite{basis-sharing} improves upon this by sharing singular vectors from SVD of concatenated weights, but lacks fine-grained control over layer-specific adaptation.

MASA generalizes these ideas by introducing \textit{learned, adaptive sharing} via dictionary atoms. Instead of copying or projecting onto fixed bases, MASA learns a compact set of matrix atoms that capture shared patterns across layers, with each layer reconstructing its weights via layer-specific coefficients. This provides a smooth spectrum between full sharing and full independence.

\paragraph{Structured Matrix Factorization and Dictionary Learning}
Our method is inspired by dictionary learning in signal processing \cite{sparse_coding}, where signals are represented as sparse linear combinations of learned basis elements. In deep learning, this idea has been applied to compress/optimize CNNs \cite{liu2018dictionary,xiao2020degradation} and low-rank adaptations \cite{yu2024effectively}.

MASA extends this principle to \textit{Transformer weight matrices}, treating each attention projection as a signal to be reconstructed from a shared dictionary. Unlike low-rank methods that enforce a uniform rank constraint across all layers and projection types, MASA learns separate dictionaries of shared matrix atoms for each attention projection and represents each layer's weights as adaptive linear combinations of these atoms, allowing the effective rank to vary by layer and projection type. This flexible decomposition achieves higher compression ratios without sacrificing performance.

Moreover, MASA integrates seamlessly into standard training—unlike methods requiring distillation \cite{sun2019patient} or auxiliary reconstruction losses. It also avoids architectural inflation (e.g., widening layers to compensate for compression), making it a plug-and-play solution.

By unifying dictionary learning with Transformer architecture, MASA occupies a unique point in the design space: exploiting inter-block redundancy through a theoretically grounded, flexible, and practical framework. Thus, we provide a scalable solution without sacrificing performance.
\section{Matrix Atom Sharing in Attention (MASA)}
\label{sec:MASA}

We consider a deep neural network architecture composed of $ L $ identical transformer blocks, each consisting of a multi-head self-attention module followed by a position-wise feed-forward network (FFN). Let $ \mathbf{W}_\ell \in \mathbb{R}^{d \times h} $ denote any of the (Q, K, V, O) weight projection matrices of the attention component in the $ \ell $-th block, for $ \ell = 1, \dots, L $. The total number of parameters across all $ L $ blocks for this particular projection is thus $ L \cdot d \cdot h $, which can be prohibitively large for deep models.

Our objective is to exploit existing potential redundancies among the weight matrices $ \{\mathbf{W}_\ell\}_{\ell=1}^L $ by introducing a matrix weight-sharing mechanism between blocks. To accomplish this goal and motivated by dictionary learning methods, we propose a representation learning strategy that expresses the input model weights of different blocks in the form of a linear combination of shared basic components. 

In the context of classical dictionary learning, the shared basic components are called atoms and they compose a dictionary, while the linear coefficients indicate the contribution of each atom in the representation of a specific input weight. This modeling (approximation) strategy can be expressed in matrix form as:
\begin{equation}
\mathbf{W} \approx \mathbf{D} \mathbf{C},
\label{eq:DL_approx}
\end{equation}
where in our case $\mathbf{W} = \begin{bmatrix} \operatorname{vec}(\mathbf{W}_1) & \cdots & \operatorname{vec}(\mathbf{W}_L) \end{bmatrix} \in \mathbb{R}^{d \cdot h \times L}$ is composed by stacking horizontally the vectorized versions of the model weights for the $L$ blocks in the network, $\mathbf{D} = \begin{bmatrix} \operatorname{vec}(\mathbf{D}_1) & \cdots & \operatorname{vec}(\mathbf{D}_S) \end{bmatrix} \in \mathbb{R}^{d \cdot h \times S}$ is the dictionary, where $\operatorname{vec}(\mathbf{D}_s) \in \mathbb{R}^{d \cdot h}$, $s = 1, \dots, S$ represents the $s$-th matrix atom, $\mathbf{D}_s \in \mathbb{R}^{d \times h}$, in vectorized form and $S$ is the total number of dictionary atoms, while $\mathbf{C} \in \mathbb{R}^{S \times L}$ represents the linear coefficients of the representation.

By carefully examining Eq.~\eqref{eq:matrix_approx}, we can express each individual weight $\mathbf{W}_l$ as:

\begin{equation}
\hat{\mathbf{W}}_l = \sum_{s=1}^{S} c_{ls} \mathbf{D}_s, \text{ with } \mathbf{D}_s \in \mathbb{R}^{d \times h}, c_{ls} \in \mathbb{R},
\label{eq:matrix_approx}
\end{equation}
where $c_{ls}$ is a scalar entry of the coefficient matrix $\mathbf{C}$. The above formulation describes our weight-sharing mechanism, where each individual weight matrix $\mathbf{W}_l$ is defined by a collection of shared weights (dictionary $\mathbf{D}$) and individual per block mixing coefficients ($\mathbf{c}_l = [c_{l1} \; \cdots \; c_{lS}]^\top \in \mathbb{R}^S$). By employing this strategy in a transformer model, we can significantly reduce the network parameters, with the exact compression rate for a specific type of weight projection matrix computed as $r = 1 - \frac{S(d\cdot h + L)}{L \cdot d \cdot h} \approx 1 - \frac{S}{L}$, with $S < L$ and $L < d \cdot h$. 

In dictionary learning, the optimal pair of the dictionary $\mathbf{D}$ and linear coefficients $\mathbf{C}$ are usually estimated by minimizing the approximation error.

\begin{equation}
\mathbf{D}^*, \mathbf{C}^* = \mathop{\mathrm{arg\,min}}_{\mathbf{D} \in \mathcal{D},\, \mathbf{C} \in \mathcal{C}} \left\| \mathbf{W} - \mathbf{D}\mathbf{C} \right\|_F^2,
\end{equation}
where both the dictionary and the coefficients can be potentially further constrained. Typical constraints that are imposed on the atoms is that they maximize the mutual incoherence property (near-orthogonality condition) and be of unit-norm.

In our case, we propose to learn the shared matrix atoms and the linear coefficients jointly via back-propagation on the network training loss. While it would be possible to enforce similar soft-constraints as those mentioned above by using additional terms in the training loss, we avoid doing so to allow for a more flexible learning process. Our proposed weight sharing-strategy is applied independently to Q, K, V and O projection matrices within attention blocks to promote a better expressivity of the model.

\subsection{Matrix Weight-Sharing for Pretrained Models}
Here, we discuss how we can extend our proposed weight-sharing strategy to existing pretrained transformer models. We begin by providing an overview of the Matrix Principal Component Analysis (Matrix PCA), which plays a key role in our framework. Subsequently, we present a transformer-block grouping method that enables the effective application of matrix PCA within groups of transformer blocks. Additionally, we propose a data-aware, layerwise local optimization criterion that dynamically refines the low-rank residuals. This approach leverages activation statistics extracted from the pretrained model to optimize performance on downstream tasks. Overall, our approach seeks to reduce the number of parameters, while preserving essential performance of the pretrained model.

\paragraph{Matrix PCA}

Similarly to MASA's approach, as described in Eq.~\eqref{eq:matrix_approx}, given a set of pretrained weights $\{\mathbf{W}_l\}_{l=1}^L$ we aim to approximate each one as a linear combination of a collection of shared matrix components. However, unlike our training-from-scratch strategy, we don't rely on the network loss and instead we aim for an analytical solution that minimizes the norm of the approximation error between the pretrained weights and the approximated ones. We further require that the shared matrix components are of unit-norm and orthogonal to each other, so it holds $\operatorname{tr}\left( \mathbf{D}_i^\top \mathbf{D}_j \right) = \delta_{ij}$ with $\delta$ the Kronecker delta function, while the linear coefficients are computed as $c_{ls} = \operatorname{tr}\left( \mathbf{D}_s^\top \mathbf{W}_l \right)$. Thus, we are looking for a matrix basis of a subspace in $\mathbb{R}^{d\times h}$. The basis matrices, which constitute the basis, can be recovered as the minimizer of the following objective:
\small
\begin{equation}
\centering
\mathbf{D}_1^*, .., \mathbf{D}_S^* = \mathop{\mathrm{arg\,min}}_{\substack{
    \mathbf{D}_s \in \mathbb{R}^{d \times h} \\
    \operatorname{tr}\left(\mathbf{D}_i^\top \mathbf{D}_j \right) = \delta_{ij}
}} 
\sum_{l=1}^{L} \left\| \mathbf{W}_l - \sum_{s=1}^{S} \operatorname{tr}\left( \mathbf{D}_s^\top \mathbf{W}_l \right) \mathbf{D}_s \right\|_F^2,
\label{eq:orthogonal_dict_learning}
\end{equation}
\normalsize
where $\operatorname{tr}(\cdot)$ denotes the matrix trace. Fortunately, the above minimization problem admits a closed-form solution (we refer to the appendix for a detailed derivation) that involves the eigenvectors corresponding to the $S$ largest eigenvalues of the matrix product $\mathbf{W} \mathbf{W}^\top$, with $\mathbf{W}$ defined as in Eq.~\eqref{eq:DL_approx}.

\paragraph{Grouping Strategy.}
To apply MASA to pretrained large language models, we first group transformer blocks into shared-weight segments, where each group of blocks has its own shared dictionary. The grouping is based on functional similarity of blocks. First, we calculate output for each transformer block using a small set of calibration data. Then, using the model's final output projection as a semantic probe, we map each block’s averaged (over tokens) hidden state to the output vocabulary space, obtaining a sequence of probability distributions over layers. By computing the Kullback–Leibler divergence between consecutive distributions, we identify segments of blocks that induce minimal semantic change—indicating functional redundancy. We then form groups of consecutive blocks where intra-group distributional drift is small, ensuring that parameter sharing occurs among behaviorally similar layers. This data-driven, training-free strategy enables structured compression while preserving semantic coherence, and facilitates practical adaptation of pretrained LLMs without fine-tuning. Step-by-step description provided in the Appendix.

\paragraph{Local Refinement.} To enhance the fidelity of MASA in pretrained models without fine-tuning, we introduce a data-informed local refinement strategy that captures reconstruction residuals with compact, structured representations. After grouping blocks and computing shared dictionary atoms via Matrix PCA, we reconstruct each layer’s weights and compute the residual $\Delta \mathbf{W}_l = \mathbf{W}_l - \hat{\mathbf{W}}_l$. Instead of modeling $\Delta \mathbf{W}_l$ directly, we apply a Cholesky whitening transform based on calibration data, and approximate $L_l \Delta \mathbf{W}_l$ with a low-rank matrix, where $L_l$ is the Cholesky factor~\cite{cholesky_book} of the input autocorrelation. This accounts for data geometry and improves approximation efficiency.

We further propose an adaptive rank allocation scheme that distributes the residual budget according to the role of each weight matrix in the attention computation. Motivated by the rank inequality $\operatorname{rank}(\mathbf{A}\mathbf{B}) \leq \min\{\operatorname{rank}(\mathbf{A}), \operatorname{rank}(\mathbf{B})\}$, we allocate more residual capacity to matrices with higher intrinsic rank (e.g., $\mathbf{W}_q$, $\mathbf{W}_o$) and less to those with structural constraints (e.g., $\mathbf{W}_k$, $\mathbf{W}_v$ in GQA/MQA). This asymmetric allocation ensures optimal use of the parameter budget under architectural imbalances. Detailed description of the proposed dynamic ranking algorithm is provided in the Appendix.

Our refinement is fully training-free and plug-and-play, significantly reducing approximation error while preserving compatibility with pretrained checkpoints.

\section{Experiments}
\label{sec:experiments}

\subsection{Experimental Setup}

\paragraph{Model Architecture.}
We evaluate our approach within the standard Transformer architecture~\cite{transformers}, which employs multi-head self-attention layers followed by GeLU-activated feed-forward networks (FFNs). As text tokenizer, we adopt a well-known Llama tokenizer~\cite{llama} and conduct experiments across three model scales: small (110M parameters, denoted Transformer-S), medium (335M, Transformer-M), and large (729M, Transformer-L). This scaling allows us to analyze how architectural modifications interact with model capacity.

We focus on structured parameter sharing in the attention module, particularly in the query (Q), key (K), value (V), and output (O) projection matrices. We consider two compression regimes:

1. \textbf{High compression:} 66.7\% reduction in attention parameters, achieved by employing $S = L/3$ shared matrices \textit{separately} across each of Q, K, V, and O projections (denoted MASA-QKVO).
    
2. \textbf{Moderate compression:} 50\% reduction, where only Q, K, and V projections are defined using $S = L/3$ shared weights, while the O projections for each transformer block are left untouched (denoted MASA-QKV).

This design enables a controlled study of the trade-off between representational expressiveness and computational efficiency. In ablation studies, we further investigate: (i) how scaling model size interacts with compression, (ii) the impact of varying the number of shared weight matrices ($S$), and (iii) how the performance is affected if shared dictionary atoms are common for Q, K, V, and O projections.  

Moreover, to enhance the stability and adaptability of the learned mixing factors $\mathbf{c}_l \in \mathbb{R}^S$ for each block $l$, we introduce a block-specific embedding-based parameterization. Each block is assigned a unique trainable embedding vector, which serves as input to a 3-layer MLP that predicts the corresponding coefficients $\mathbf{c}_l$. This parameterized formulation decouples the optimization dynamics of the mixing coefficients from direct, potentially unstable, updates, thereby reducing gradient fluctuations during training and promoting smoother convergence. Importantly, this design acts as an implicit regularization mechanism, guiding the model toward more stable configurations. After training, both MLP and embeddings are discarded, and only the final coefficient matrix $\mathbf{C}$ is retained for inference. This ensures no additional computational overhead at test time while preserving the benefits of smooth and more efficient training.

\paragraph{Training Protocol.}
All models are trained on the RefinedWeb dataset~\cite{refinedweb}, a high-quality web corpus filtered for linguistic and factual coherence. We follow the Chinchilla-optimal training regime~\cite{hoffmann2022empirical}, allocating $20\times$ the number of model parameters in training tokens (e.g., 2.2B tokens for Transformer-S). 

We follow the established scaling laws~\cite{zhang2022opt, hoffmann2022empirical} and set up hyperparameters, such as learning rate, batch size, and learning rate warmup schedule accordingly. Training is performed on A100 40GB GPUs using mixed-precision and optimized attention kernels via FlashAttention~\cite{dao2022flashattention} to handle long sequences efficiently. For reproducibility, all training hyperparameters are listed in the Appendix.

\paragraph{Evaluation Protocol} We assess zero-shot performance across two benchmark families: multiple-choice reasoning and language modeling. We compute the average accuracy across \textbf{8} test sets, including PIQA~\cite{bisk2019piqa0}, HellaSwag~\cite{hellaswag}, and ARC Challenge~\cite{arc-challenge}. Also, we estimate perplexity for LAMBADA~\cite{paperno2016lambada} and WikiText~\cite{wikitext2}. Detailed description for each benchmark can be found in the Appendix. 

\subsection{Results}
\begin{table*}[!t]
\centering
\small
\resizebox{\textwidth}{!}{%
\renewcommand{\arraystretch}{1.3}
\begin{tabular}{l|>{\centering\arraybackslash}p{1.cm}|c>{\centering\arraybackslash}p{0.5cm}>{\centering\arraybackslash}p{1.3cm}>{\centering\arraybackslash}p{0.6cm}>{\centering\arraybackslash}p{0.6cm}ccc|>{\centering\arraybackslash}p{0.5cm}>{\centering\arraybackslash}p{1.45cm}|c}
\hline
\textbf{Model} & \textbf{Attn CR}  & \textbf{PIQA$\uparrow$} & \textbf{Hella Swag$\uparrow$} & \textbf{LAMBADA acc.$\uparrow$}& \textbf{ARC easy$\uparrow$} &\textbf{ARC chall.$\uparrow$}  &\textbf{SciQ$\uparrow$} &\textbf{Race$\uparrow$}& \textbf{MMLU$\uparrow$} & \textbf{Wiki Text$\downarrow$} & \textbf{LAMBADA ppl$\downarrow$}.& \textbf{AVG, \%$\uparrow$} \\ \hline
Transformer-S (110M)     & 0\%                      & 0.593 & 0.279 & 0.195 & 0.340 & 0.202 & 0.585 & 0.254 & 0.229 & 76.11           & 167.39          & 33.48 \\ \hline
GQA                      & 41.7\%                   & 0.600 & 0.282 & 0.193 & 0.329 & 0.217 & 0.571 & 0.243 & 0.229 & 78.41           & 187.71          & 33.34 \\ 
\textbf{MASA-QKV} (ours) & 50.0\%                   & 0.589 & 0.282 & 0.231 & 0.355 & 0.213 & 0.590 & 0.264 & 0.229 & \textbf{72.08}  & \textbf{112.23} & \textbf{34.43} \\ \hdashline
Low-Rank                 & \multirow{4}{*}{66.7\%}  & 0.579 & 0.275 & 0.163	& 0.327 & 0.231 & 0.536 & 0.241 & 0.227 & 83.25          & 264.52          & 32.27 \\
Seq-Sharing              &                          & 0.589 & 0.279 & 0.204	& 0.332 & 0.214 & 0.571 & 0.260 & 0.228 & 80.35          & 171.52          & 33.50 \\
Repeat-all-over          &                          & 0.583 & 0.279 & 0.209	& 0.334 & 0.209 & 0.561 & 0.251 & 0.229 & 78.97          & 162.15          & 33.24 \\
\textbf{MASA-QKVO} (ours)&                          & 0.602 & 0.278 & 0.214  & 0.332 & 0.214 & 0.572 & 0.256 & 0.229 & \textbf{72.82} & \textbf{133.62} & \textbf{33.74} \\ \hline
Transformer-M (335M)     & 0\%                      & 0.631 & 0.323 & 0.289	& 0.372	& 0.221	& 0.636	& 0.272 & 0.238	& 44.49          & 48.76          & 37.31 \\ \hline
GQA                      & 43.8\%& 0.650 & 0.316 & 0.316	& 0.370	& 0.226	& 0.598	& 0.270 & 0.241	& 46.21          & 53.55          & 37.39 \\
\textbf{MASA-QKV} (ours) & 50.0\%                   & 0.632	& 0.330	& 0.295	& 0.382 & 0.224 & 0.638	& 0.282	& 0.245	& 42.31          & \textbf{45.27} & \textbf{37.86} \\ \hdashline
Low-Rank                 & \multirow{4}{*}{66.7\%}  & 0.629 & 0.315 & 0.271	& 0.380	& 0.229	& 0.592	& 0.284 & 0.244	& 47.48          & 59.34          & 36.84 \\
Seq-Sharing              &                          & 0.634 & 0.315 & 0.284	& 0.371	& 0.228	& 0.602	& 0.275 & 0.231	& 47.36          & 55.50          & 36.80 \\
Repeat-all-over            &                          & 0.640 & 0.316 & 0.274	& 0.368	& 0.228	& 0.612	& 0.271 & 0.234	& 47.63          & 60.57          & 36.83 \\
\textbf{MASA-QKVO} (ours)&                          & 0.636 & 0.322 & 0.290	& 0.375	& 0.219	& 0.626	& 0.288 & 0.231	& \textbf{45.00} & \textbf{50.26} & \textbf{37.37} \\ \hline
Transformer-L (729M)     & 0\%                      & 0.675 & 0.397 & 0.397	& 0.422	& 0.240	& 0.696	& 0.296 & 0.243	& 30.88          & 20.73 & 42.12 \\ \hline
GQA                      & 41.7\%                   & 0.675 & 0.394 & 0.374	& 0.422	& 0.239	& 0.674	& 0.290 & 0.232	& 31.74          & 24.10 & 41.29 \\ 
\textbf{MASA-QKV} (ours) & 50.0\%                   & 0.684 & 0.399 & 0.391	& 0.415 & 0.235	& 0.688	& 0.295	& 0.230	& \textbf{30.83} & \textbf{22.08} & \textbf{41.74} \\ \hdashline
Low-Rank                 & \multirow{4}{*}{66.7\%}  & 0.666 & 0.379 & 0.324	& 0.414	& 0.246	& 0.646	& 0.289 & 0.238	& 33.28          & 31.74 & 40.07 \\
Seq-Sharing              &                          & 0.674 & 0.387 & 0.363	& 0.406	& 0.230	& 0.645	& 0.287 & 0.245	& 32.43          & 25.64 & 40.51 \\
Repeat-all-over            &                          & 0.681 & 0.387 & 0.341	& 0.410	& 0.242	& 0.651	& 0.287 & 0.241	& 32.27          & 27.67 & 40.54 \\
\textbf{MASA-QKVO} (ours)&                          & 0.684 & 0.398 & 0.387	& 0.413	& 0.232	& 0.675	& 0.283 & 0.228	& \textbf{31.34} & \textbf{21.21} & \textbf{41.30} \\
\hline
\end{tabular}
}
\caption{Performance of existing attention-block compression techniques on downstream tasks for different sizes of the transformer under the zero-shot setting. We report accuracy ($\uparrow$ is better) results first and then the perplexity ($\downarrow$ is better) performance on WikiText~\cite{wikitext2} validation set and on LAMBADA~\cite{paperno2016lambada}. We report the proposed MASA in two setups: MASA-QKV applies only for Q, K, V projections and MASA-QKVO for all projections in the attention module.
}
\label{tab:main}
\end{table*}

We evaluate MASA against a suite of state-of-the-art attention compression techniques, including \textbf{Grouped Query Attention (GQA)}~\cite{GQA}, \textbf{Sequential-Sharing}~\cite{sequential_sharing}, \textbf{Repeat-all-over}~\cite{MobileLLM}, and \textbf{Low-Rank Attention} inspired by LoRA~\cite{denil2013predicting, structuredffns}. While LoRA was originally proposed for CNNs~\cite{denil2013predicting} and later adapted to compress FFN blocks in Transformers~\cite{structuredffns}, we apply low-rank decomposition exclusively to the attention projections (Q, K, V, O), constraining each to rank $r = d/3$ to achieve a 66.7\% parameter reduction in the attention module. For GQA~\cite{GQA}, we use 8 groups in Transformer-M and 6 groups in Transformer-S and Transformer-L, yielding moderate compression (43.8\% and 41.7\%, respectively), which is known to preserve performance well. The rest of the methods are configured to achieve exactly 66.7\% compression in the attention block for a fair comparison.

As shown in Table~\ref{tab:main}, MASA outperforms all competing methods across both reasoning accuracy and language modeling perplexity, despite matching or exceeding their compression rates. Notably, MASA-QKV, which compresses only Q, K, and V projections (50\% reduction), achieves an average accuracy of 34.43\%, surpassing the full Transformer-S by \textbf{+1.0\%}, while reducing perplexity by \textbf{4.03} on WikiText and \textbf{55.16} on LAMBADA. This demonstrates that representative weight sharing across all blocks can act as an effective compression approach, improving generalization even under parameter reduction.

Meanwhile, MASA-QKVO, which compresses all four projections (66.7\% reduction), achieves slightly higher performance (\textbf{+0.26\%}) than full model in accuracy and significantly outperforms all compressed baselines in perplexity. This confirms our sharing mechanism preserves critical representational capacity while drastically reducing parameters.

\begin{table*}[!t]
\centering
\small
\resizebox{\textwidth}{!}{%
\renewcommand{\arraystretch}{1.35}
\begin{tabular}{l|>{\centering\arraybackslash}p{0.5cm}>{\centering\arraybackslash}p{1.0cm}|c>{\centering\arraybackslash}p{0.5cm}>{\centering\arraybackslash}p{1.3cm}>{\centering\arraybackslash}p{0.6cm}>{\centering\arraybackslash}p{0.6cm}ccc|>{\centering\arraybackslash}p{0.5cm}>{\centering\arraybackslash}p{1.45cm}|c}
\hline
\textbf{Model} & \textbf{Attn CR}  & \textbf{Num. Weights} & \textbf{PIQA$\uparrow$} & \textbf{Hella Swag$\uparrow$} & \textbf{LAMBADA acc.$\uparrow$}& \textbf{ARC easy$\uparrow$} &\textbf{ARC chall.$\uparrow$}  &\textbf{SciQ$\uparrow$} &\textbf{Race$\uparrow$}& \textbf{MMLU$\uparrow$} & \textbf{Wiki Text$\downarrow$} & \textbf{LAMBADA ppl$\downarrow$}.& \textbf{AVG, \%$\uparrow$} \\ \hline
Transformer-S    & 0\%        & N/A & 0.593 & 0.279 & 0.195 & 0.340 & 0.202 & 0.585 & 0.254 & 0.229 & 76.11           & 167.39          & 33.48 \\ \hline
MASA-QKV (ours)  & 62.5\%     & 2   & 0.594	& 0.276	& 0.217	& 0.337	& 0.208 & 0.585	& 0.253 & 0.229 & 73.40           & 133.03          & 33.77 \\
MASA-QKV (ours)  & 50.0\%     & 4   & 0.590	& 0.282	& 0.231	& 0.355	& 0.213 & 0.590	& 0.264 & 0.228 & 72.08           & \textbf{112.23} & \textbf{34.43} \\
MASA-QKV (ours)  & 37.5\%     & 6   & 0.595 & 0.282 & 0.228 & 0.337 & 0.205 & 0.569 & 0.259 & 0.229 & 71.32           & 121.43	        & 33.81 \\
MASA-QKV (ours)  & 25.0\%     & 8   & 0.608	& 0.277	& 0.228	& 0.343	& 0.212 & 0.561	& 0.248 & 0.228 & \textbf{70.77}  & 114.19          & 33.84 \\ 

\hline
MASA-QKVO (ours) & 83.3\%     & 2   & 0.594	& 0.276	& 0.219	& 0.332	& 0.224	& 0.573	& 0.257	& 0.230	& 74.79	          & 139.44          & 33.82 \\
MASA-QKVO (ours) & 66.7\%     & 4   & 0.602 & 0.278 & 0.214 & 0.332 & 0.214 & 0.572 & 0.256 & 0.229 & 72.82           & 133.62          & 33.74 \\ 
MASA-QKVO (ours) & 50.0\%     & 6   & 0.585	& 0.280	& 0.229	& 0.341	& 0.212	& 0.568	& 0.264	& 0.229	& 71.70	          & \textbf{120.86} & 33.86 \\
MASA-QKVO (ours) & 33.3\%     & 8   & 0.602	& 0.284	& 0.225	& 0.345 & 0.224	& 0.552	& 0.255	& 0.230	& \textbf{70.66}  & 123.80		    & \textbf{33.94} \\
\hline
\end{tabular}
}
\caption{The results for various number of representative weights that are shared over all blocks on the downstream tasks. MASA is evaluated under two setups: shared matrices for each Q, K, V (denoted as MASA-QKV) and shared matrices for each Q, K, V, and O separately (QKVO)}
\label{tab:ablation_num_basis}
\end{table*}

\paragraph{Model Scaling Analysis.}
We further investigate how MASA behaves across model scales (S: 110M, M: 335M, L: 729M). As shown in Table~\ref{tab:main}, MASA consistently outperforms existing compression methods across all sizes.

For the small model (Transformer-S), MASA-QKVO exhibits the largest relative gains: it outperforms the second-best method (Repeat-all-over Sharing) by \textbf{6.15} lower perplexity on WikiText and \textbf{28.53} on LAMBADA, along with a \textbf{+0.5\%} improvement in average accuracy. This suggests that in low-capacity regimes, the inductive bias introduced by MASA-QKVO is particularly beneficial, compensating for limited model expressiveness.

As model size increases, the absolute perplexity gap between MASA-QKVO and baselines narrows — for example, Repeat-all-over Sharing lags by \textbf{6.46} on LAMBADA at the large scale — but remains substantial. In contrast, accuracy gap slightly increases with scale: at the large model level, MASA-QKVO exceeds the second-best method by \textbf{+0.7\%} in average accuracy. This indicates our method better leverages increased model capacity under parameter constraints.

When compared to the \textit{uncompressed} Transformer baseline, MASA-QKVO performs exceptionally well at smaller scales but shows a small performance gap at larger ones. Specifically, MASA-QKV (50\% compression) achieves \textbf{0.05} lower perplexity on WikiText and \textbf{0.38\%} lower average accuracy than the vanilla Transformer-L. Under higher compression (66.7\%), the gap widens to \textbf{0.46} in perplexity and \textbf{0.82\%} in accuracy. This trend suggests that larger models benefit more from layer-wise diversity — a finding consistent with scaling laws~\cite{kaplan2020scaling}. Even so, the fact that a two-thirds compressed attention module (MASA-QKVO) remains within \textbf{1\%} of a full 729M-parameter model and superior results over SOTA approaches, demonstrates the efficiency and consistent scaling abilities of our method.

\begin{table*}[!ht]
\centering
\small
\resizebox{\textwidth}{!}{%
\renewcommand{\arraystretch}{1.3}
\begin{tabular}{l|>{\centering\arraybackslash}p{0.5cm}|c>{\centering\arraybackslash}p{0.5cm}>{\centering\arraybackslash}p{1.4cm}>{\centering\arraybackslash}p{0.6cm}>{\centering\arraybackslash}p{0.6cm}ccc|>{\centering\arraybackslash}p{0.5cm}>{\centering\arraybackslash}p{1.4cm}|c}
\hline
\textbf{Model} & \textbf{Attn CR}  & \textbf{PIQA$\uparrow$} & \textbf{Hella Swag$\uparrow$} & \textbf{LAMBADA acc.$\uparrow$}& \textbf{ARC easy$\uparrow$} &\textbf{ARC chall.$\uparrow$}  &\textbf{SciQ$\uparrow$} &\textbf{Race$\uparrow$}& \textbf{MMLU$\uparrow$} & \textbf{Wiki Text$\downarrow$} & \textbf{LAMBADA ppl$\downarrow$}.& \textbf{AVG, \%$\uparrow$} \\ \hline
Llama 3.2 1B        & N/A  & 0.745  & 0.637  & 0.629  & 0.605  & 0.362  & 0.883  & 0.378  & 0.370  & 11.57          & 5.73           & 57.61 \\ \hline
SVD-LLM             & 20\% & 0.733  & 0.597  & 0.554  & 0.533  & 0.337  & 0.827  & 0.373  & 0.295  & 15.08          & 9.55           & 53.11 \\
\textbf{Matrix PCA}(ours) & 20\% & 0.742  & 0.610  & 0.599  & 0.573  & 0.344  & 0.873  & 0.356  & 0.330  & \textbf{12.61} & \textbf{6.65}  & \textbf{55.34}  \\ \hdashline
SVD-LLM             & 30\% & 0.712  & 0.551  & 0.482  & 0.505  & 0.296  & 0.808  & 0.365  & 0.276  & 17.89          & 14.20          & 49.94  \\
\textbf{Matrix PCA}(ours) & 30\% & 0.732  & 0.561  & 0.545  & 0.537  & 0.324  & 0.830  & 0.342  & 0.288  & \textbf{14.91} & \textbf{8.79}  & \textbf{52.00}  \\ \hline \hline
Llama 3.2 3B        & N/A  & 0.775  & 0.736  & 0.705  & 0.716  & 0.460  & 0.927  & 0.400  & 0.543  & 9.26           & 3.94           & 65.78 \\ \hline
SVD-LLM             & 20\% & 0.768  & 0.705  & 0.651  & 0.668  & 0.436  & 0.906  & 0.386  & 0.509  & 11.50          & 5.57           & 62.86 \\
\textbf{Matrix PCA}(ours) & 20\% & 0.771  & 0.713  & 0.690  & 0.703  & 0.438  & 0.926  & 0.393  & 0.506  & \textbf{10.08} & \textbf{4.39}  & \textbf{64.25}  \\ \hline \hline
Llama 3.1 8B        & N/A  & 0.812  & 0.791  & 0.754  & 0.813  & 0.538  & 0.944  & 0.393  & 0.629  & 7.33           & 3.13           & 70.93 \\ \hline
SVD-LLM             & 20\% & 0.797  & 0.775  & 0.705  & 0.763  & 0.508  & 0.939  & 0.396  & 0.590  & 9.05           & 4.63           & 68.41 \\
\textbf{Matrix PCA}(ours) & 20\% & 0.811  & 0.780  & 0.739  & 0.800  & 0.529  & 0.943  & 0.400  & 0.605  & \textbf{7.84}  & \textbf{3.35}  & \textbf{70.09}  \\
\hline
\end{tabular}
}
\caption{Comparison of our method against SVD-LLM on different compression ratios and different model sizes.}
\label{tab:compression_comparison_pretrained}
\end{table*}

\paragraph{Impact of Number of Weight Sharing Matrices} 
We evaluate MASA with varying numbers of shared matrices ($S = 2, 4, 6, 8$) to analyze the trade-off between compression and performance. As mentioned before, we consider two configurations: MASA-QKV and MASA-QKVO. Compression rate (CR) decreases as $S$, total number of dictionary atoms, increases. All models are trained under the same Chinchilla-optimal regime. The results in Table~\ref{tab:ablation_num_basis} show that:

- \textbf{Larger the dictionary size, better the performance} For MASA-QKVO, increasing $S$ (i.e., reducing compression) consistently improves perplexity and average accuracy over multiple-choice reasoning benchmarks. This confirms that richer representational capacity in the dictionary enhances long-range modeling and reasoning.

- \textbf{Accuracy is robust to compression:} Average accuracy remains stable across all settings, varying by less than 0.5\%. Notably, MASA-QKVO with $S=8$ achieves the highest average accuracy (33.94\%), suggesting that moderate sharing with sufficient dictionary diversity might act as a regularizer.
    
- \textbf{The output (O) projection matters:} Comparing the two setups, MASA-QKV (unshared O) outperforms MASA-QKVO at similar compression rates. For example, at $S=4$, MASA-QKV achieves 121.4 perplexity on LAMBADA vs. 133.6 for MASA-QKVO. Thus, compressing the output projection (O) introduces a bottleneck that harms language modeling more than compressing Q, K, V.

- \textbf{QKV projections are more compressible:} Even with high compression (e.g., $S=2$, 62.5\% reduction), MASA-QKV maintains performance similar (or better) to the vanilla (uncompressed) model. In contrast, compressing O—even with more shared matrices—fails to recover the same level of performance. This supports the idea that Q, K, V are more redundant across layers, while O plays a more specialized role in information transformation.

Thus, these findings suggest a practical design principle: \textit{prioritize compression on Q, K, V projections and preserve parameter independence in the output projection}. In the Appendix we further explore how utilizing a common dictionary across Q, K, V, and O affects model performance.

- \textbf{Computational Overhead:} MASA-QKVO achieves 1240 tokens/sec vs. 1352 tokens/sec for Transformer-S (sequence length $n$=256, batch 16, A100-40GB), a ${\sim}8.3\%$ drop due to atom-based weighting. FLOPs increase from $4n^2d + 8d^2n$ to $4n^2d + 8d^2(n + k)$, where, $d$ is hidden dimension, Given that $k \ll n$ is the number of atoms, the additional computational cost remains negligible.

\paragraph{Extension to Vision Transformers.}
To ensure the scalability of the proposed method, we trained different versions of vision transformers on CIFAR10 \cite{cifar10}, CIFAR100 \cite{cifar100} and TinyImageNet \cite{tinyimagenet} datasets. 

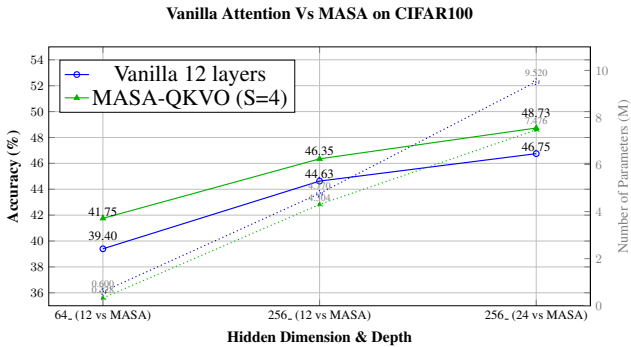
\begin{figure}[!h]
    \centering
    \begin{adjustbox}{width=\columnwidth}
    \begin{tikzpicture}
    \begin{axis}[
        width=15cm, height=8cm,
        xlabel={Hidden Dimension \& Depth},
        ylabel={Accuracy (\%)},
        ymajorgrids=true,
        ylabel style={black},
        xmin=0.75, xmax=3.25,
        ymin=35, ymax=55,
        xtick={1,2,3},
        xticklabels={
            \fontsize{14}{14}\selectfont 64\_ (12 vs MASA),
            \fontsize{14}{14}\selectfont 256\_ (12 vs MASA),
            \fontsize{14}{14}\selectfont 256\_ (24 vs MASA)
        },
        legend style={at={(0.02,0.95)},anchor=north west,font=\fontsize{16}{17}\selectfont},
        tick label style={font=\large},
        label style={font=\bfseries\huge},
        title style={yshift=10pt,font=\bfseries\Large},
        ytick distance=2,
        grid=both,
    ]
    
    \addplot [
        color=blue, mark=o, solid, thick
    ]
    coordinates {
        (1,39.40)
        (2,44.63)
        (3,46.75)
    };
    
    \addplot [
        color=green!70!black, mark=triangle*, solid, thick
    ]
    coordinates {
        (1,41.75)
        (2,46.35)
        (3,48.73)
    };
      
    \node at (axis cs:1,40.4) {\fontsize{13}{13}\selectfont 39.40};
    \node at (axis cs:1,42.3) {\fontsize{13}{13}\selectfont 41.75};
    \node at (axis cs:2,45.2) {\fontsize{13}{13}\selectfont 44.63};
    \node at (axis cs:2,47.0) {\fontsize{13}{13}\selectfont 46.35};
    \node at (axis cs:3,47.3) {\fontsize{13}{13}\selectfont 46.75};
    \node at (axis cs:3,49.9) {\fontsize{13}{13}\selectfont 48.73};
    \addlegendentry{Vanilla 12 layers}
    \addlegendentry{MASA-QKVO (S=4)}
    \addlegendimage{color=green!70!black, mark=triangle*, solid, thick}
\end{axis}
    
    \begin{axis}[
        width=15cm, height=8cm,
        axis y line*=right,
        axis x line=none, 
        ylabel={Number of Parameters (M)},
        ylabel style={gray},
        label style={font=\huge},
        ymin=0, ymax=11,
        xmin=0.75, xmax=3.25,
        ytick={0,2,4,6,8,10},
        yticklabel style={gray, /pgf/number format/fixed, /pgf/number format/precision=1},
        xtick=\empty,
        hide x axis,
        y tick label style={font=\large, color=gray}]

    \addplot [
        color=blue!70!black, mark=o, dotted, thick
    ]
    coordinates {
        (1,0.6)
        (2,4.77)
        (3,9.5195)
    };
    
    \addplot [
        color=green!70!black, mark=triangle*, dotted, thick
    ]
    coordinates {
        (1,0.328)
        (2,4.304)
        (3,7.476)
    };
    
    \node [gray] at (axis cs:1,1.3) {\fontsize{13}{13}\selectfont 0.600};
    \node [gray] at (axis cs:1.2,0.67) {\fontsize{13}{13}\selectfont 0.328};
    \node [gray] at (axis cs:2.12,4.6)  {\fontsize{13}{13}\selectfont 4.770};
    \node [gray] at (axis cs:2.12,3.8)  {\fontsize{13}{13}\selectfont 4.304};
    \node [gray] at (axis cs:3,9.9)  {\fontsize{13}{13}\selectfont 9.520};
    \node [gray] at (axis cs:3.11,7.5) {\fontsize{13}{13}\selectfont 7.476};
    
    \end{axis}
    \end{tikzpicture}
    \end{adjustbox}
    \caption{Evaluation results of different ViT models trained on CIFAR100 trainset, the blue and green solid plots represent the Top1-Accuracy of the vanilla attention models and MASA respectively. The dotted lines represent the parameter count of the full models respectively.} 
    \label{cifar100_results}
\end{figure}


As illustrated in Fig. \ref{cifar100_results}, our proposed method consistently surpasses vanilla attention across all scales. In particular, we investigated three experimental configurations: a compact architecture with $L$=12 layers and reduced width (both hidden and MLP dimensions); a fixed-depth model with ($L$=12) with $4$ times larger width in both hidden and MLP dimensions; and an increased-depth variant with ($L$=24 layers) layers while preserving the original width.

In all configurations, proposed MASA-QKVO with S=4 maintains a significant performance advantage over vanilla attention on both CIFAR-100, CIFAR-10, and TinyImageNet datasets. The training details as well as results for CIFAR-10 and TinyImageNet can be found in the Appendix.
\subsection{Pretrained LLMs}

In this section, we evaluate the proposed method across large language models with varying architectural scales, conducting a comparative analysis against SVD-LLM as the primary baseline. To reiterate the methodology: for a predefined number of groups and a fixed number of basis matrices per group, we first apply Matrix Principal Component Analysis (Matrix PCA) to achieve a global low-rank approximation of the weight matrices within each group. This is followed by a local, data-aware refinement stage that operates on the residual components—defined as the difference between the original and reconstructed weights—leveraging calibration data to enhance reconstruction accuracy.

In Table~\ref{tab:compression_comparison_pretrained}, our method demonstrates consistent superiority over SVD-LLM across different model scales and compression ratios, achieving higher average accuracy across diverse downstream benchmarks while maintaining better language modeling performance as measured by perplexity. For larger architectures such as LLama 3.1 8B, our approach enables up to $20\%$ compression of the attention weight matrices while preserving approximately $99\%$ of the original model's accuracy, indicating minimal degradation in semantic and reasoning capabilities despite parameter reduction.

\section{Conclusion}
\label{sec:conclusion}

To conclude, we introduce a novel strategy, named MASA, that leverages dictionary learning to reduce redundancy in attention projections of the Transformer-based networks. By decomposing weight matrices into shared matrix atoms and reconstructing them via linear combinations, MASA achieves 66.7\% parameter reduction in attention modules without sacrificing performance. Unlike prior works constrained by rigid sharing schemes or complex retraining, MASA operates as a plug-and-play solution within standard optimization frameworks, maintaining training efficiency and significantly improving model compactness.

Our extensive empirical results demonstrate that MASA surpasses existing techniques—including GQA, low-rank approximations, and layer-wise sharing across language, reasoning, and vision tasks. Notably, MASA's compatibility with pretrained models enables training-free compression with minimal accuracy degradation, offering a practical pathway for real-world deployment.

By bridging classical signal processing principles with modern neural architecture design, MASA establishes a scalable, theoretically grounded paradigm for building parameter-efficient Transformers. We believe this work will serve as an indication of how powerful inter-layer matrix decomposition is and foster the research community to explore more inter-layer compression methods.
\bibliography{references}

\clearpage
\section{Supplementary Material for $``$Share Your Attention: Transformer Weight Sharing via Matrix-based Dictionary Learning$"$}
\label{sec:appendix}

\subsection{Matrix PCA}
\paragraph{Recovery of Principal Matrices - Proof}
As we discussed in the main paper, for a set of pretrained weights $\cbr{\m W_l}_{l=1}^L$ we want to estimate the $S$-principal matrix components $\cbr{\m D_s}_{s=1}^S$ that minimize the objective loss:
\bal
\mc{L} = \suml_{l=1}^L\norm{\m W_l-\suml_{s=1}^S \tr{\m D_s^\transp\m W_l}\m D_s}{F}^2,
\label{eq:mpca_loss}
\eal
under the additional constraint $\tr{\m D_i^\transp\m D_j}=\delta_{ij}$.

To do so, we fist rewrite the loss in Eq.~\eqref{eq:mpca_loss} in the equivalent form:
\bal
\mc{L} &= \suml_{l=1}^L\norm{\vc{\m W_l}-\suml_{s=1}^S \vc{\m D_s}^\transp\vc{\m W_l}\vc{\m D_s}}{2}^2 \nonumber\\
&= \suml_{l=1}^L\norm{\vc{\m W_l}-\pr{\suml_{s=1}^S \vc{\m D_s}\vc{\m D_s}^\transp}\vc{\m W_l}}{2}^2 \nonumber\\
&= \suml_{l=1}^L\norm{\vc{\m W_l}-\m D\m D^\transp\vc{\m W_l}}{2}^2\nonumber\\
&=\norm{\m W -\m D\m D^\transp\m W}{F}^2,
\eal
where $\m D=\bbmtx \vc{\m D_1}&\ldots&\vc{\m D_S}\ebmtx \in \R^{d\cdot h\times S}$ and $\m W=\bbmtx \vc{\m W_1} &\ldots&\vc{\m W_L}\ebmtx\in\R^{d\cdot h\times L}$. We note that in the above reformulation we have used the property: $\tr{\m A^\transp\m B} =\vc{\m A}^\transp\vc{\m B}$. In addition, we can compactly represent the constraint $\tr{\m D_i^\transp\m D_j}=\delta_{ij}$ as $\m D^\transp\m D=\m I$, where $\m I\in\R^{S\times S}$ is the Identity matrix. 

Based on the above, the principal matrix components can be recovered as:
\bal
\m D^\star = \argmin_{\m D}\norm{\m W-\m D\m D^\transp\m W}{F}^2\,\,\mbox{s.t}\,\, \m D^\transp\m D=\m I.
\label{eq:argmin}
\eal

Next, we observe that the objective loss can be further simplified as:
\bal
\mc L &= \norm{\m W}{F}^2 + \tr{\m W^\transp\m D\m D^\transp\m D\m D^\transp\m W} -2\tr{\m W^\transp\m D\m D^\transp\m W}\nonumber\\
&\overset{\m D^\transp\m D=I}{=} \norm{\m W}{F}^2 -\tr{\m W^\transp\m D\m D^\transp\m W}\nonumber\\
& = \norm{\m W}{F}^2 -\tr{\m D^\transp\m W \m W^\transp\m D}.
\label{eq:L_simple}
\eal

Combining Eqs.~\eqref{eq:argmin} and \eqref{eq:L_simple} we can obtain the principal matrices as the maximizer of the following problem:
\bal
\m D^\star = \argmax_{\m D}\tr{\m D^\transp\m W \m W^\transp\m D}\,\,\mbox{s.t}\,\, \m D^\transp\m D=\m I.
\label{eq:argmax}
\eal
Tha above maximization problem has a closed-form solution, which is fully defined by the eigenvalues of the matrix $\m P=\m W\m W^\transp$. Specifically, the matrix $\m P\in\R^{d\cdot h\times d\cdot h}$, which is symmetric and positive definite, admits the eigenvalue decomposition $\m P=\m U\bm\Lambda\m U^\transp$, with $\m U\in \R^{d\cdot h\times d\cdot h}$ holding the eigenvectors of $\m P$ in its columns. Then the maximizer of Eq.~\eqref{eq:argmax} is recovered as $\m D^\star=\m U_S$ where $\m U_S\in \R^{d\cdot h\times S}$ is a cropped version of $\m U$ formed with the $S$ eigenvectors corresponding to the largest eigenvalues of $\m P$. Based on the above, we can finally recover the $s$-th principal matrix as:
\bal
\m D^\star_s = \vc{\m U_S^s}^{-1},\,\, s=1,\ldots, S
\eal
where $\m U_S^s$ denotes the $s$-th column of the matrix $\m U_S$ and $\vc{\cdot}^{-1}$ refers to the inverse operation of $\vc{\cdot}$, that is it performs the mapping $\vc{\cdot}^{-1}:\R^{d\cdot h}\mapsto \R^{d\times h}$.

One potential issue with the described solution is that the matrix $\m P\in\R^{d\cdot h\times d\cdot h}$ has huge dimensions and the computation of its eigenvalue decomposition can be practically infeasible. To overcome this difficulty, we notice that the eigenvectors of $\m P$ exactly match the left-singular vectors of $\m W\in\R^{d\cdot h\times L}$. Indeed, if $\m W$ admits the singular value decomposition (SVD) $\m W=\m U\bm\Sigma\m V^\transp$, then we have that: $\m P=\m W\m W^\transp=\m U\bm\Sigma^2\m U^\transp\equiv \m U\bm\Lambda\m U^\transp$, with $\bm\Lambda=\bm\Sigma^2$. Therefore, instead of performing the eigenvalue decomposition on $\m P$ we can recover $\m U$ by computing the SVD on $\m W$. Given that the second dimension of $\m W$ is significantly smaller than its first dimension, that is $L << d\cdot h$, the SVD of $\m W$ can be computed very efficiently. 

\paragraph{Grouping Method}
Applying MASA to pretrained large language models requires a principled strategy for assigning transformer blocks to shared-weight groups. While global optimization over all possible groupings is intractable, we adopt a greedy, data-driven approach that leverages the model’s own output semantics to identify functionally redundant layers.

Consider a standard decoder-only LLM consisting of an embedding layer, $L$ transformer blocks $\{\text{Block}_l\}_{l=1}^L$, and a final output projection (unembedding) layer $\m W_{\text{out}} \in \mathbb{R}^{d \times |\Sigma|}$, where $d$ is the hidden dimension and $|\Sigma|$ is the vocabulary size. Let $\m Y_l \in \mathbb{R}^{T \times d}$ denote the output hidden states of block $l$ across $T$ tokens in a sequence. We compute the layer-wise averaged representation:
\[
\bar{\m y}_l = \frac{1}{T} \sum_{i=1}^{T} \m Y_{li} \in \mathbb{R}^d,
\]
which serves as a global summary of the model’s internal state after block $l$.

We then apply the pretrained output projection as a natural mapping function $f(\cdot)$ into the output probability space:
\[
\m p_l = f(\bar{\m y}_l) = \text{softmax}\left( \m W_{\text{out}}^\top \bar{\m y}_l \right) \in \mathbb{R}^{|\Sigma|},
\]
yielding a probability mass function (pmf) over the vocabulary. This pseudo-output reflects the model’s current predictive distribution before subsequent blocks refine it. We compute $\m P = [\m p_1, \dots, \m p_L]$ over a small, diverse calibration dataset (1024 samples from RefinedWeb) and average across samples to obtain stable layer-wise distributions.

To quantify functional similarity between consecutive blocks, we compute the Kullback–Leibler (KL) divergence:
\[
\mathcal{D}_{\text{KL}}(\m p_l \parallel \m p_{l+1}) = \sum_{k=1}^{|\Sigma|} \m p_l^{(k)} \log \left( \frac{\m p_l^{(k)}}{\m p_{l+1}^{(k)}} \right),
\]
which measures the distributional shift induced by block $l+1$. A small KL divergence suggests that the later block performs minimal semantic refinement, indicating potential redundancy.

We form groups of \textit{consecutive} blocks such that the cumulative KL divergence within each group is minimized. Specifically, we define group boundaries $\{g_0 = 1, g_1, \dots, g_K = L+1\}$ by placing splits at local maxima of $\mathcal{D}_{\text{KL}}(\m p_l \parallel \m p_{l+1})$. This ensures that blocks with similar behavioral impact—i.e., those that collectively stabilize the output distribution—are grouped together.

Within each group, all blocks share the same dictionary atoms in MASA while maintaining unique coefficient vectors, enabling structured weight sharing with preserved expressivity. Although this greedy, sequential grouping is not globally optimal, it is computationally efficient, reproducible, and leverages the model’s intrinsic semantics without requiring gradients or fine-tuning. It thus provides a practical and effective solution for training-free adaptation of pretrained models. More sophisticated clustering methods (e.g., hierarchical or spectral) are left for future work.

\paragraph{Local Refinement}
We investigate the feasibility of employing the proposed weight-sharing mechanism without relying on post-compression fine-tuning (i.e., "healing"), by introducing a data-informed local refinement strategy applied to the approximation residuals. Specifically, after grouping the transformer blocks and estimating the subspace basis matrices for all weights within each group via Matrix PCA, we reconstruct the approximated weight matrix $\hat{\m W}_l$
for each individual layer. Then we compute the residual, defined as the discrepancy between the original and reconstructed weights, as:
\bal
\Delta \m W_l = \m W_l - \hat{\m W}_l,
\eal
where $\m W_l$ denotes the pretrained weight matrix of the $l$-th layer in the  model. We model these residuals as if they exhibit low-rank structure, suggesting that the reconstruction error can be efficiently captured using a compact representation. More specifically, similar to the strategy followed in SVD-LLM, rather than opting for a low-rank representation of $\Delta \m W_l$ itself, we consider the product $\m L_l\Delta \m W_l$ to be of low-rank, where $\m L_l$ denotes the Cholesky factor of the autocorrelation matrix computed from the respective calibration data. Under this strategy the overall approximation of a pretrained weight $\m W_l$ can be expressed as:
\bal
\hat{\m W}_l = \suml_{s=1}^S c_{ls}\m D_s + \m L_l^{-1}g\pr{\m L_l\Delta\m W_l; r},
\eal
where $g\pr{\cdot; r}$ denotes the r-rank approximation of the input argument.

Given a target overall compression ratio $\alpha$ , and considering that $B$ basis matrices are retained per group during the Matrix PCA stage across $L$ layers, we compute the parameter budget allocated to the residual components $\beta$ as:
\bal
\beta \approx \frac{\alpha \cdot L - B}{L - B}
\eal
This expression ensures a consistent total parameter count while enabling adaptive distribution of compression between the shared basis and the residual correction terms.
To further refine our approach, we revisit the mathematical formulation of the attention mechanism:
\bal
\text{out} = \text{softmax}\left(\frac{ \m H  \m W_q  \m W_k^T  \m H^T}{\sqrt{d}}\right) \m H \m W_v  \m W_o
\label{eq:attention}
\eal
In conventional multi-head attention (MHA), the projection matrices $\m W_q,\m W_k,\m W_v,\m W_o$
—corresponding to queries, keys, values, and output projection, respectively—are typically of compatible and balanced dimensions. However, modern large language models (LLMs) increasingly adopt grouped-query attention (GQA) or multi-query attention (MQA), where $\m W_k$ and $\m W_v$ are shared across multiple heads, resulting in significantly reduced dimensionality for keys and values compared to queries. Consequently, the intermediate outputs $\m H\cdot \m W_v$ and $\m H\cdot \m W_k$
are broadcasted or repeated to match the dimensionality required in subsequent operations.
This architectural asymmetry has important implications for rank behavior in matrix products. Recall the fundamental inequality from linear algebra:
\bal
\text{rank}(\m A \m B) \leq \min(\text{rank}(\m A), \text{rank}(\m B)).
\eal
An equivalent but more insightful version of this inequality is the following:
\bal
\text{rank}(\m A \m B) &\leq \text{rank}(\m A) + \text{rank}(\m B)\nonumber \\
&- \max(\text{rank}(\m A), \text{rank}(\m B)).
\eal

This inequality highlights that the rank of a product is constrained not only by the minimum rank but also by the disparity between the ranks of the operands. Motivated by this property, we propose an adaptive rank allocation strategy for residual decomposition, where the rank of the residual approximation is dynamically adjusted based on the type and role of the weight matrix (e.g., $\m W_q$ vs. $\m W_k$ ) and its intrinsic rank constraints within the attention computation graph. This allows for more efficient use of the parameter budget, particularly in asymmetric architectures where uniform rank assignment would be suboptimal.
By integrating structural awareness with residual refinement, our method enhances approximation without requiring retraining, making it suitable for efficient, plug-in compression of large-scale pretrained models.

We implement Algorithm \ref{alg:balanced_compression}, in which the residuals of the output and values projections are processed jointly. The same holds true for the   residuals of keys and queries. Then, we apply a local whitening transform (computed using Cholesky Decomposition on calibration data) to these residuals. To quantitatively assess the reconstruction error, we invoke the Eckart–Young–Mirsky theorem, according to which the Frobenius norm of the reconstruction error corresponds to the sum of the squared singular values that have been omitted during the truncation process, that is:
\bal
\|\mathbf{A} - \mathbf{A}_k\|_F^2 = \sum_{i=k+1}^{r_A} (\sigma_i^A)^2,
\eal
where $\sigma_i^A$ denotes the $i-th$ singular value of matrix $\m A$ and $r_A$ is its rank.

\begin{algorithm}[!h]
\caption{Balanced SVD-Based Matrix Compression with Adaptive Ratio Adjustment}
\label{alg:balanced_compression}
\begin{algorithmic}
\REQUIRE Two matrices $\mathbf{A} \in \mathbb{R}^{m \times n}$, $\mathbf{B} \in \mathbb{R}^{m \times k}$, initial compression ratio $\beta \in (0,1)$, tolerance $\epsilon > 0$
\ENSURE Optimal rank selection $(r_A^*, r_B^*)$ and total approximation error

\STATE Perform SVD on $\mathbf{A}$: 
    \[
    \mathbf{A} = \mathbf{U}_A \mathbf{\Sigma}_A \mathbf{V}_A^\top, \quad \mathbf{\Sigma}_A = \text{diag}(\sigma_1^A, \sigma_2^A, \dots, \sigma_{r_A}^A)
    \]
\STATE Perform SVD on $\mathbf{B}$: 
    \[
    \mathbf{B} = \mathbf{U}_B \mathbf{\Sigma}_B \mathbf{V}_B^\top, \quad \mathbf{\Sigma}_B = \text{diag}(\sigma_1^B, \sigma_2^B, \dots, \sigma_{r_B}^B)
    \]

\STATE Flip singular values of $\mathbf{\Sigma}_A$: $\tilde{\boldsymbol{\sigma}}_A = \text{flip}(\boldsymbol{\sigma}_A)$
\STATE Compute cumulative sum of squared singular values: $\boldsymbol{c}_A = \text{cumsum}(\tilde{\boldsymbol{\sigma}}_A^2)$
\STATE Flip back: $\boldsymbol{s}_A = \text{flip}(\boldsymbol{c}_A)$

\STATE Flip singular values of $\mathbf{\Sigma}_B$: $\tilde{\boldsymbol{\sigma}}_B = \text{flip}(\boldsymbol{\sigma}_B)$
\STATE Compute cumulative sum of squared signular values: $\boldsymbol{c}_B = \text{cumsum}(\tilde{\boldsymbol{\sigma}}_B^2)$
\STATE Flip back: $\boldsymbol{s}_B = \text{flip}(\boldsymbol{c}_B)$

\STATE Set initial ranks based on $\beta$:
    \[
    r_A = \lfloor \frac{(1 - \beta) \cdot m \cdot n}{m + n}  \rfloor, \quad
    r_B =\lfloor \frac{(1 - \beta) \cdot m \cdot k}{m + k}  \rfloor
    \]
\STATE Compute initial total singular value sum:
    \[
    S_{\text{total}} = \sum_{i=r_A+1}^{\min(m,n)} (\sigma_i^A)^2 + \sum_{j=r_B+1}^{\min(m,k)} (\sigma_j^B)^2 
    \]
     \[
     = \boldsymbol{s}_A[min(m,n) - r_A + 1] + \boldsymbol{s}_B[min(m,k) - r_B + 1] 
    \]
\STATE Initialize optimal ranks: $r_A^* \gets r_A$, $r_B^* \gets r_B$
\STATE Initialize minimum error: $E_{\min} \gets S_{\text{total}}$

\WHILE{change in $r_A$ and $r_B$ exceeds $\epsilon$}
    \STATE Increase compression on $\mathbf{A}$: $r_A \gets r_A - 1$
    \STATE Decrease compression on $\mathbf{B}$: $r_B \gets r_B + 1$
    
    \IF{$r_A < 1 \;||\; r_B > \min(m,k)$}
        \STATE \textbf{break}
    \ENDIF

    \STATE Compute current error:
        \[
        E = \sum_{i=r_A+1}^{\min(m,n)} (\sigma_i^A)^2 + \sum_{j=r_B+1}^{\min(m,k)} (\sigma_j^B)^2
        \]
         \[
     = \boldsymbol{s}_A[min(m,n) - r_A + 1] + \boldsymbol{s}_B[min(m,k) - r_B + 1] 
    \]

    \IF{$E < E_{\min}$}
        \STATE $E_{\min} \gets E$
        \STATE $r_A^* \gets r_A$
        \STATE $r_B^* \gets r_B$
    \ENDIF
\ENDWHILE

\RETURN $r_A^*, r_B^*, E_{\min}$
\end{algorithmic}
\end{algorithm}

\subsection{MASA Training details}
In our experiments, we train three backbone sizes—Small (S), Medium (M) and Large (L)—across four variants: the standard Transformer, the GQA ablation, and two MASA variants (QKV and QKVO factorized bases). All models use a shared 32K‐word vocabulary and identical feed-forward scaling (×4) per layer; the only architectural differences are in hidden dimensionality, number of layers/heads, and whether keys, queries and/or values are factorized. Table \ref{tab:masa_training} summarizes these specifications.

For each size class, we train with the same total token budget, batch size, and peak learning rate, as shown in Table \ref{tab:hyperparams}. We use AdamW ($\beta_1$=0.9, $\beta_2$=0.999, weight decay=0.1) with linear warmup over the first 10\% of training steps and cosine decay thereafter. Gradients are clipped to a global norm of 1.0.

\begin{table*}[ht]
  \centering
  \caption{Architectural Details of vanilla and compressed Transformer Models by Size}
  \label{tab:masa_training}
    \resizebox{\textwidth}{!}{%
    \renewcommand{\arraystretch}{1.3}
    \begin{tabular}{ l  |l  |c  |c |c  |c  |c  |c  |c }
        \hline 

Model & Size (M) & Layers & Hidden dim. & FFN scale & Att. head dim. & Num heads & Num. KV heads & Vocab size \\
        \hline 

\textbf{Transformer-S}  & 109.5 & 12 & 768 & 4 & 64 & 12 & 12 & 32000 \\ \hline
\textbf{GQA} & 97.7     & 12 & 768 & 4 & 64 & 12 & 2 & 32000 \\
\textbf{MASA-QKV S=4}   & 95.4 & 12 & 768 & 4 & 64 & 12 & 12 & 32000 \\ \hdashline
\textbf{Seq.-Sharing}   & 90.6 & 12 & 768 & 4 & 64 & 12 & 12 & 32000 \\  
\textbf{Repeat-all-over}& 90.6 & 12 & 768 & 4 & 64 & 12 & 12 & 32000 \\  
\textbf{Low-Rank r=128} & 90.6 & 12 & 768 & 4 & 64 & 12 & 12 & 32000 \\  
\textbf{MASA-QKVO S=4}  & 90.6 & 12 & 768 & 4 & 64 & 12 & 12 & 32000 \\ \hline 

\textbf{Transformer-M}  & 334.8 & 24 & 1024 & 4 & 64 & 16 & 16 & 32000 \\ \hline
\textbf{GQA}            & 290.7 & 24 & 1024 & 4 & 64 & 16 & 2  & 32000 \\
\textbf{MASA-QKV S=8}   & 284.4 & 24 & 1024 & 4 & 64 & 16 & 16 & 32000 \\ \hdashline
\textbf{Seq.-Sharing}   & 267.6 & 24 & 1024 & 4 & 64 & 16 & 16 & 32000 \\ 
\textbf{Repeat-all-over}& 267.6 & 24 & 1024 & 4 & 64 & 16 & 16 & 32000 \\ 
\textbf{Low-Rank r=176} & 268.7 & 24 & 1024 & 4 & 64 & 16 & 16 & 32000 \\ 
\textbf{MASA-QKVO S=8}  & 267.6 & 24 & 1024 & 4 & 64 & 16 & 16 & 32000 \\ \hline 

\textbf{Transformer-L}  & 728.6 & 24 & 1536 & 4 & 128 & 12 & 12 & 32000 \\ \hline
\textbf{GQA}            & 634.3 & 24 & 1536 & 4 & 128 & 12 & 2 & 32000 \\
\textbf{MASA-QKV S=8}   & 615.4 & 24 & 1536 & 4 & 128 & 12 & 12 & 32000 \\ \hdashline
\textbf{Seq.-Sharing}   & 577.6 & 24 & 1536 & 4 & 128 & 12 & 12 & 32000 \\
\textbf{Repeat-all-over}& 577.6 & 24 & 1536 & 4 & 128 & 12 & 12 & 32000 \\ 
\textbf{Low-Rank r=256} & 577.6 & 24 & 1536 & 4 & 128 & 12 & 12 & 32000 \\
\textbf{MASA-QKVO S=8}  & 577.6 & 24 & 1536 & 4 & 128 & 12 & 12 & 32000 \\ \hline 

\end{tabular}
}
\end{table*}

\begin{table}[!h]
    \centering
    \caption{Hyperparameter settings are standardized across all models of a given size. Compression methods are trained using the same configuration (e.g., number of tokens, effective batch size, initial learning rate, scheduler) as the baseline Transformer with which they are compared, ensuring a fair evaluation.}
    \resizebox{0.48\textwidth}{!}{%
    \renewcommand{\arraystretch}{1.2}
    \begin{tabular}{l|c|c|c}
        \hline 
        Model & Tokens & Effective Batch Size & Learning Rate \\
        \hline 
        Transformer-S & 2.2B & 512 & 6.0e-4 \\
        Transformer-M & 6.7B & 512 & 3.0e-4 \\
        Transformer-L & 14.6B & 512 & 2.5e-4 \\
        \hline 
    \end{tabular}
    }
\label{tab:hyperparams}
\vspace{-3mm}
\end{table}

\subsection{Pretrained LLMs ablations}
\paragraph{Ablation on the number of groups}
To further analyze the sensitivity of performance to grouping, we conduct an ablation study varying the number of layer groups, with a fixed one basis per group. 

\begin{table}[H]
    \centering
    \caption{Analysis of grouping for compressing the attention blocks of Llama 3.2 1B model with $20\%$.}
    \resizebox{0.48\textwidth}{!}{%
    \renewcommand{\arraystretch}{1.3}
    \begin{tabular}{c|>{\centering\arraybackslash}p{1.cm}|>{\centering\arraybackslash}p{2cm}|c|c}
        \hline 
         Model & Num. groups & \textbf{Wiki Text$\downarrow$} & \textbf{LAMBADA ppl$\downarrow$} & \textbf{AVG, \%$\uparrow$} \\ \hline 
         \textbf{Llama 3.2 1B} & N/A & \textbf{11.56} & \textbf{5.72} & 0.576 \\ \hdashline 
         \textbf{MASA} & 3 & \textbf{12.55} & \textbf{6.39} & 0.552 \\
         \textbf{MASA} & 4 & 12.61          & 6.42          & 0.550 \\
         \textbf{MASA} & 5 & 12.61          & 6.51          & 0.552 \\
         \textbf{MASA} & 6 & 12.61          & 6.65          & \textbf{0.553}  \\
         \textbf{MASA} & 7 & 12.65          & 6.47          & 0.551 \\
        \hline
    \end{tabular}
    }
    \label{tab:group_ablation}
    \vspace{-3mm}
\end{table}

\begin{table}[H]
    \centering
    \caption{Analysis of number of \textbf{basis} and \textbf{groups} for compressing the attention blocks of Llama 3.2 1B model with $20\%$.}
    \resizebox{0.48\textwidth}{!}{%
    \renewcommand{\arraystretch}{1.3}
    \begin{tabular}{c|>{\centering\arraybackslash}p{.8cm}|>{\centering\arraybackslash}p{.9cm}|c|c|c}
        \hline 
         Model & Num. Basis & Num. Groups & \textbf{Wiki Text$\downarrow$} & \textbf{LAMBADA ppl$\downarrow$} & \textbf{AVG, \%$\uparrow$} \\ \hline 
         \textbf{Llama 3.2 1B} & N/A & N/A & \textbf{11.56} & \textbf{5.72} & 0.576 \\ \hdashline
         MASA                  & 1   & 6   & 12.61          & 6.65          & \textbf{0.553} \\
         MASA                  & 2   & 5   & 12.72          & 6.71          & 0.552 \\
         MASA                  & 2   & 6   & 18.03          & 12.69         & 0.486 \\
         MASA                  & 2   & 4   & \textbf{12.46} &\textbf{6.40}  & \textbf{0.553} \\
        \hline
    \end{tabular}
    }
    \label{tab:basis_ablation}
\end{table}

 The results, presented in Table \ref{tab:group_ablation}, reveal that while performance remains relatively stable across different grouping configurations, optimal setup depends on the evaluation metric. Specifically, three groups yield the lowest perplexity, whereas six groups achieve the highest average accuracy across the benchmarks.

An analysis of the resulting groupings provides insight into the functional hierarchy within the transformer architecture. With three groups, the optimal partition isolates the first layer as Group 1, the last layer as Group 2, and all intermediate layers as Group 3. Notably, for any group consisting of a single layer, local residual refinement is omitted, as Matrix PCA reduces to an identity mapping in such cases (i.e., full-rank reconstruction with one basis).
Extending to four groups, the previous partitioning preserves the first and last layer while introducing a new group for the second-to-last layer. A similar incremental strategy is observed for five groups, which further isolates the third-to-last layer. When increasing to six groups, the large middle block is bisected assigning layers from the second to the seventh into one subgroup and the remaining middle layers into another.
This grouping pattern underscores the importance of the first and last layer, which are consistently isolated across all configurations. In contrast, the internal layers demonstrate greater uniformity, enabling effective parameter sharing.
\paragraph{Ablation on the number of basis}
As previously discussed, the layer grouping strategy results in different group sizes ranging from singleton groups containing a single layer to significantly larger groups encompassing multiple consecutive layers. Given this imbalance, we investigate whether allocating additional basis matrices to larger groups rather than further partitioning them into more groups with a single basis each leads to improved approximation.

The results, summarized in Table ~\ref{tab:basis_ablation}, show that the optimal configuration is achieved with 4 groups and 2 basis matrices assigned to the largest group, while the remaining groups retain a single basis. In terms of total parameter count, this setup is approximately equivalent to a uniform configuration of 5 groups with 1 basis per group. However, the former yields superior performance in both perplexity and downstream task accuracy, indicating that increasing representational capacity within larger groups is more effective than increasing the number of groups under a fixed parameter budget. However, this benefit is subject to diminishing returns: as the number of basis matrices increases, the available parameter budget for the local data-aware refinement stage is correspondingly reduced. This trade-off implies an optimal balance between global basis expressiveness and local correction capacity.  

\subsection{Benchmarks for LLM Evaluation}
\paragraph{Multiple-Choice Reasoning.}
These tasks evaluate a model's ability to perform contextual reasoning and knowledge retrieval without fine-tuning. For each task, we estimate an accuracy of correctly chosen cases.
\begin{itemize}
\item{\textbf{PIQA}}~\cite{bisk2019piqa0}: Measures physical commonsense reasoning by selecting the most plausible method to accomplish everyday tasks.
\item{\textbf{HellaSwag}}~\cite{hellaswag}: Assesses commonsense reasoning in context by asking the model to find the most suitable continuation of a given short narrative.
\item{\textbf{MMLU}}~\cite{mmlu}: A comprehensive benchmark that covers 57 subjects across humanities, social sciences, and STEM. It provides a four-way multiple-choice questions to test academic knowledge.
\item{\textbf{ARC Challenge}}~\cite{arc-challenge}: A challenging grade-school science questions designed to assess deep reasoning and commonsense understanding of LLMs. It consists of two sections: easy and challenging.
\item \textbf{ARC-Easy}~\cite{arc-challenge}: A subset of the AI2 Reasoning Challenge dataset consisting of science exam questions for which the correct answer can be inferred from a single sentence or common knowledge, formatted as multiple-choice problems.
\item \textbf{SciQ}~\cite{sciq}: Evaluates scientific question answering by requiring models to select the correct answer from four options, based on science curriculum content typically encountered in middle and high school.
\item \textbf{RACE}~\cite{race}: Assesses reading comprehension through multiple-choice questions derived from English exams for middle and high school students in China, covering diverse topics and requiring deep reasoning over long passages.
\end{itemize}

\paragraph{Language modeling} Here, we calculate perplexity on the test splits of both datasets.
\begin{itemize}
\item \textbf{LAMBADA}~\cite{paperno2016lambada}: Evaluates the ability to predict the last word in a narrative passage requiring models to perform a context-level understanding.
\item \textbf{WikiText2}~\cite{wikitext2}: Measures language model perplexity on a large corpus of high-quality, authentic Wikipedia articles, assessing performance in open-domain, long-form text modeling.
\end{itemize}

\subsection{Analysis of Common Atom Matrix Sharing}
\label{sec:ablation_common_basis}

We next investigate whether Q, K, V, and O projections can share a \textit{common} dictionary—i.e., operate in the same subspace. This would further reduce memory footprint by eliminating a separate sets of matrix atoms per projection. We evaluate hybrid configurations where certain projections share a common dictionary (e.g., Q and K use the same atoms), while others maintain separate dictionaries. Our findings are summarized below:

- \textbf{Independent dictionaries are better.} The QKVO-Separate configuration (i.e., independent sharing per projection) achieves the best performance, confirming that Q, K, V, and O serve functionally distinct roles and benefit from specialized dictionaries.
    
- \textbf{QV have more similar dictionaries} Among shared configurations, \texttt{QV-Common} performs best (33.95\% average accuracy, 73.62 on Wikitext2, 138.71 on LAMBADA), suggesting that query and value transformations may operate in more similar subspaces—possibly because both are used to compute attention-weighted outputs. Also, jointly sharing Q, K, V (while keeping O separate) performs worse than any pairwise sharing, indicating that forcing all three to share a single dictionary over-constrains the model.

These findings reinforce that while dictionary sharing improves efficiency, \textit{preserving functional specialization} is critical for maintaining performance. A one-size-fits-all strategy is suboptimal; instead, per-projection structured sharing offers the best trade-off between compression and performance. These insights not only validate the design choice of the proposed MASA, but we hope it may provide general guidance for future structured compression methods in Transformers. The detailed tabular results are presented in Table~\ref{tab:ablation_common_separate_base}.

\begin{table*}[!h]
\centering
\small
\caption{Performance of MASA with common vs. separate representative matrices. "Common" indicates shared dictionary across projections; "Separate" uses an independent dictionary for each of the specified projections. Best performance is achieved when all projections use separate dictionary.}
\label{tab:ablation_common_separate_base}
\resizebox{\textwidth}{!}{%
\renewcommand{\arraystretch}{1.35}
\begin{tabular}{l|>{\centering\arraybackslash}p{1cm}c|c>{\centering\arraybackslash}p{1cm}>{\centering\arraybackslash}p{1.4cm}>{\centering\arraybackslash}p{0.6cm}>{\centering\arraybackslash}p{0.6cm}ccc|>{\centering\arraybackslash}p{0.5cm}>{\centering\arraybackslash}p{1.45cm}|c}
\hline
\textbf{Methods} & \textbf{Common} & \textbf{Separate} & \textbf{PIQA} & \textbf{Hella Swag} & \textbf{LAMBADA acc.}& \textbf{ARC easy} &\textbf{ARC chall.}  &\textbf{SciQ} &\textbf{Race}& \textbf{MMLU} & \textbf{Wiki Text} & \textbf{LAMBADA ppl}.& \textbf{AVG, \%} \\ \hline
Transformer-S           & N/A & N/A  & 0.593 & 0.279 & 0.195 & 0.340 & 0.202 & 0.585 & 0.254 & 0.229 & 76.11  & 167.39  & 33.48 \\ \hline
MASA (QKVO-Separate)    & -   & 4    & 0.589 & 0.282 & 0.231 & 0.355 & 0.213 & 0.590 & 0.264 & 0.229 & 72.08  & 112.23 & 34.43 \\ \hdashline
QK-Common VO-Separate   & 8   & 4    & 0.589 & 0.280 & 0.207 & 0.327 & 0.222 & 0.552 & 0.240 & 0.229 & 78.13  & 152.14 & 33.09 \\
KV-Common QO-Separate   & 8   & 4    & 0.594 & 0.280 & 0.212 & 0.344 & 0.219 & 0.583 & 0.247 & 0.230 & 74.98  & 146.27 & 33.87 \\
QV-Common KO-Separate   & 8   & 4    & 0.597 & 0.282 & 0.214 & 0.345 & 0.223 & 0.565 & 0.261 & 0.229 & 73.62  & 138.71 & 33.95 \\
QKV-Common O-Separate   & 12  & 4    & 0.590 & 0.279 & 0.178 & 0.332 & 0.218 & 0.563 & 0.238 & 0.229 & 85.01  & 220.90 & 32.86 \\
\hline
\end{tabular}
}
\vspace{-2mm}
\end{table*}

\begin{table*}[!h]
\centering
\small
\caption{The comparison of MASA-QKV and Transformer-S trained on RefinedWeb training dataset tokens, which is 600 times larger than the model size. The aim is to see the performance under large corpus of dataset.}
\label{tab:ablation_large_training}
\resizebox{\textwidth}{!}{%
\renewcommand{\arraystretch}{1.35}
\begin{tabular}{l|>{\centering\arraybackslash}p{1.9cm}|c>{\centering\arraybackslash}p{1cm}>{\centering\arraybackslash}p{1.4cm}>{\centering\arraybackslash}p{0.6cm}>{\centering\arraybackslash}p{0.6cm}ccc|>{\centering\arraybackslash}p{0.5cm}>{\centering\arraybackslash}p{1.45cm}|c}
\hline
\textbf{Methods} & \textbf{Training Tokens ($\times10^9$)} & \textbf{PIQA} & \textbf{Hella Swag} & \textbf{LAMBADA acc.}& \textbf{ARC easy} &\textbf{ARC chall.}  &\textbf{SciQ} &\textbf{Race}& \textbf{MMLU} & \textbf{Wiki Text} & \textbf{LAMBADA ppl}.& \textbf{AVG, \%} \\ \hline
Transformer-S &65  & 0.646	& 0.342	& 0.338	& 0.381	& 0.243 & 0.669	& 0.271	& 0.245	& 42.44	          & \textbf{34.85}  & \textbf{39.20} \\ 
MASA-QKV      &65  & 0.648	& 0.341	& 0.332	& 0.391	& 0.241 & 0.652	& 0.283	& 0.231	& \textbf{41.11}  & 37.02           & 38.97 \\ \hline
\end{tabular}
}
\vspace{-2mm}
\end{table*}

\subsection{Correlation of Dictionary Atoms}
\label{sec:ablation_dictionary_correlation}
\begin{figure*}[!h]
    \centering
    \begin{minipage}{0.23\textwidth}
        \centering
        \includegraphics[width=\linewidth]{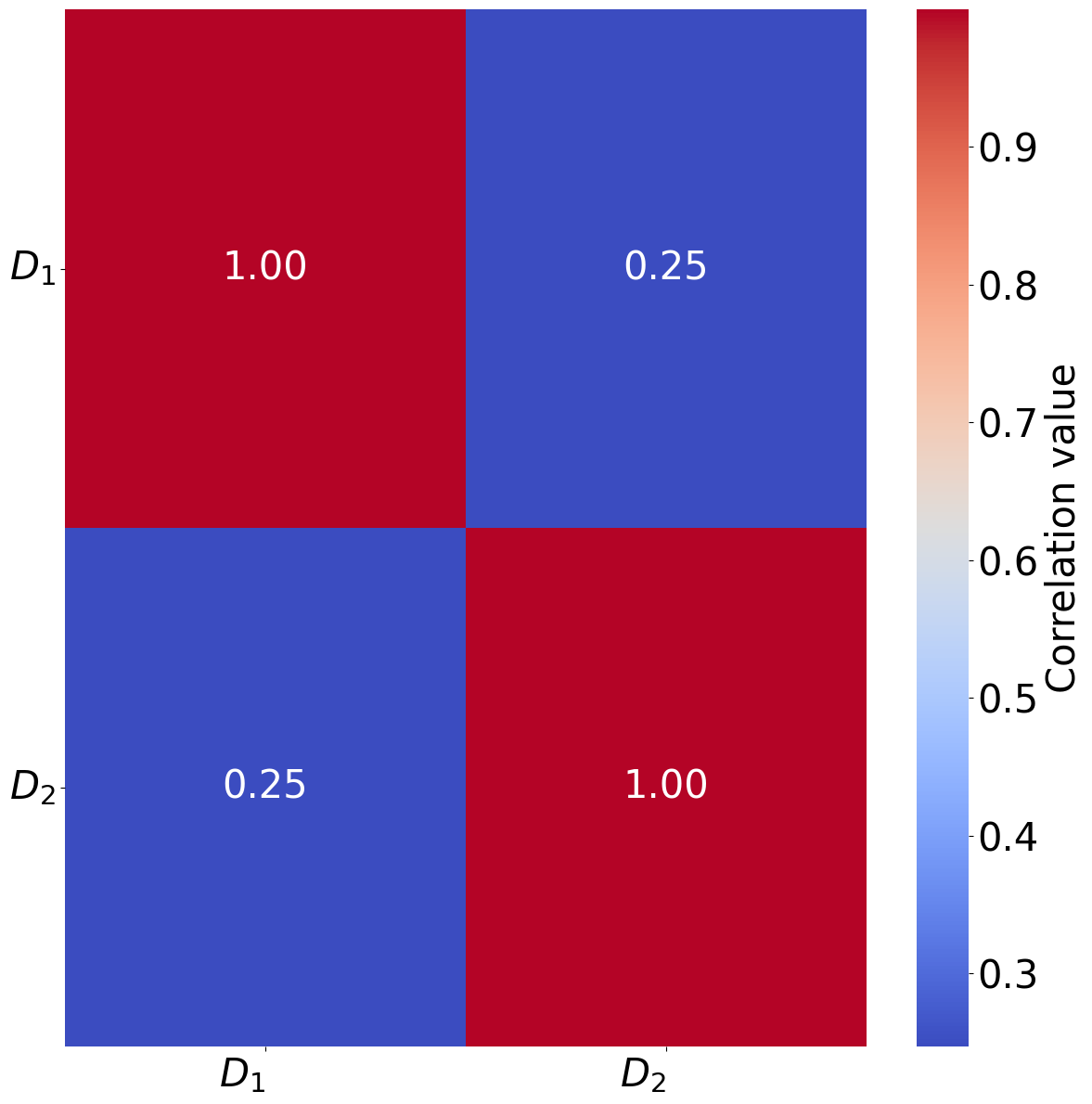}
        \caption*{(A)\quad $D^Q$}
    \end{minipage}
    \hfill
    \begin{minipage}{0.23\textwidth}
        \centering
        \includegraphics[width=\linewidth]{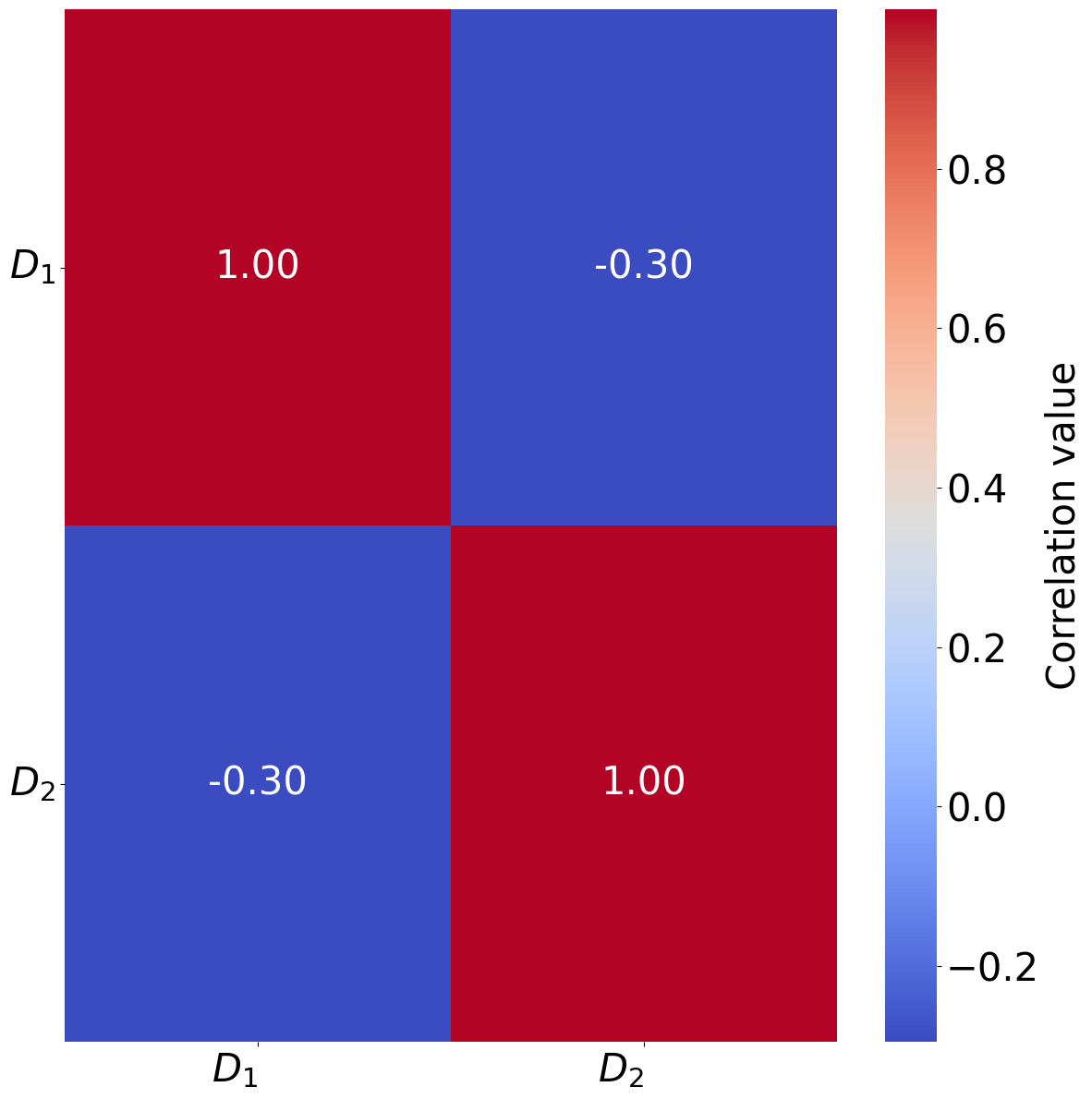}
        \caption*{(B)\quad $D^K$}
    \end{minipage}
    \hfill
    \begin{minipage}{0.23\textwidth}
        \centering
        \includegraphics[width=\linewidth]{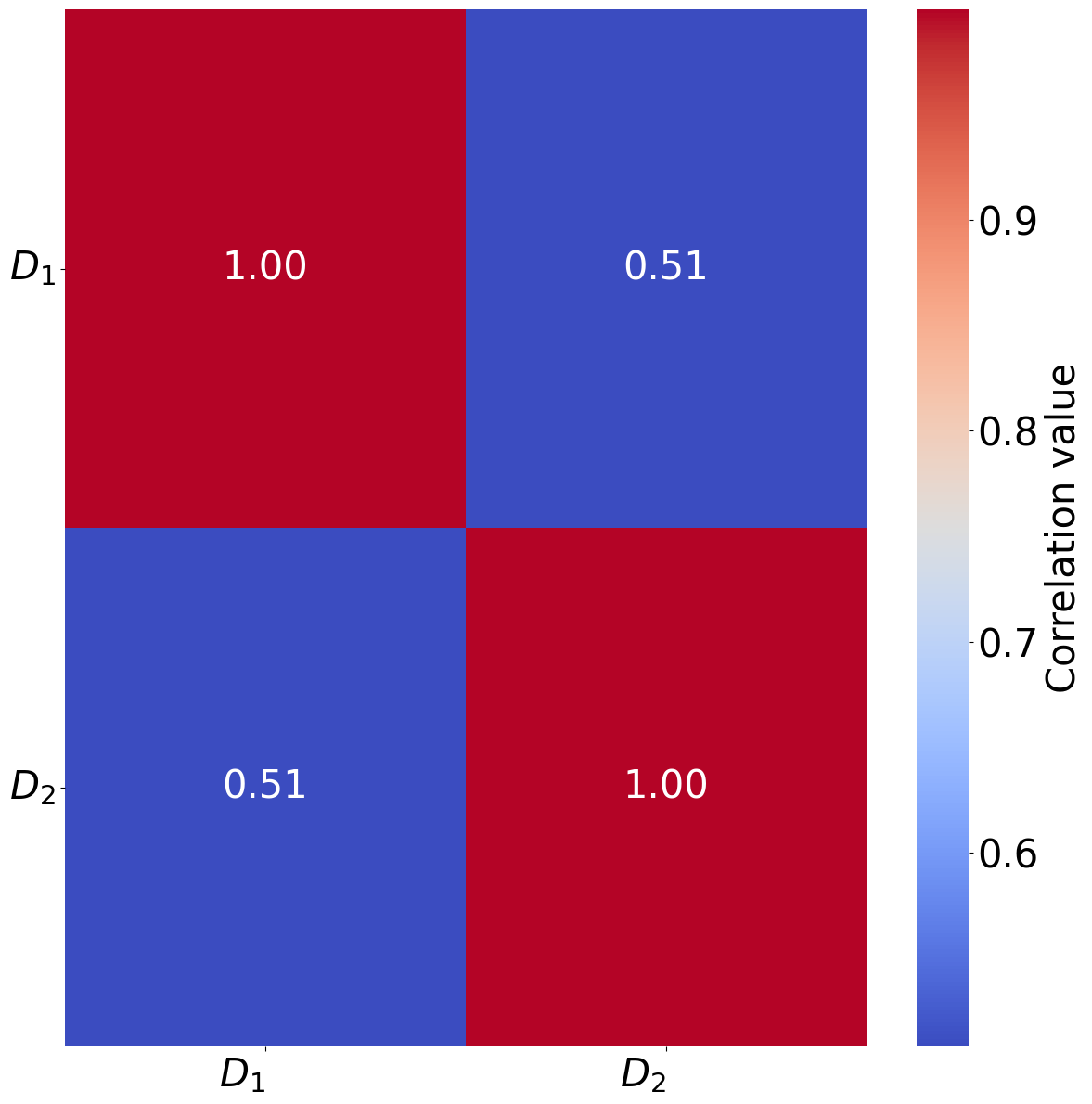}
        \caption*{(C)\quad $D^V$}
    \end{minipage}
    \hfill
    \begin{minipage}{0.23\textwidth}
        \centering
        \includegraphics[width=\linewidth]{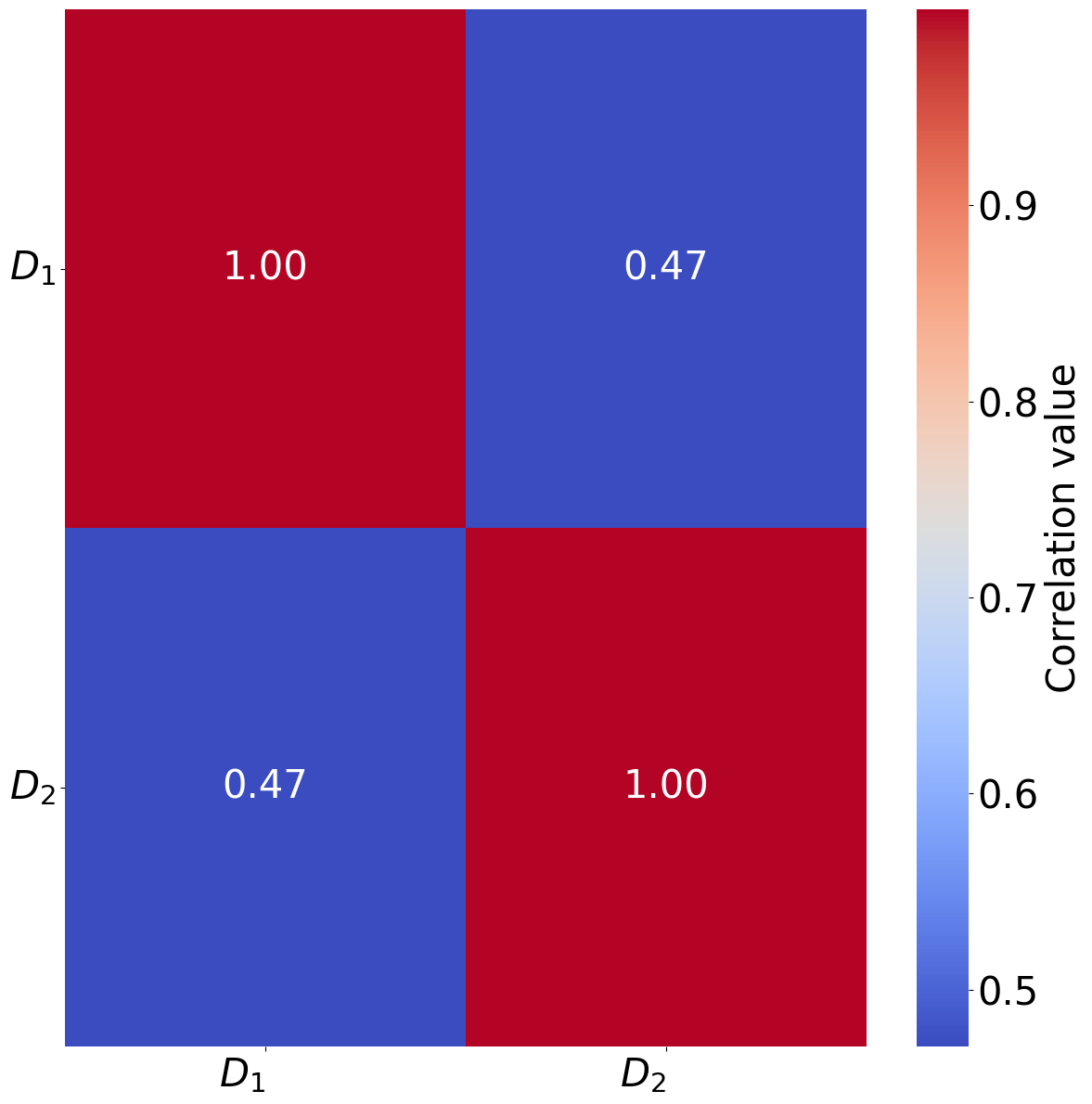}
        \caption*{(D)\quad $D^O$}
    \end{minipage}
    \caption{Cosine similarity between atoms in each dictionary for Q, K, V, and O projections (left to right). Higher absolute values indicate stronger atom correlations. Results shown for MASA-QKVO (small transformer, S=2)}
    \label{fig:correlation_basis_2_qkvo}
    \vspace{-3mm}
\end{figure*}

Here we provide the visualization of correlation between atoms of the learned dictionary for different sizes of the dictionary (number of weights $S$). The correlation is a cosine similarity and calculated with the following formula:
\bal
\Phi(\m D_i, \m D_j) = \frac{\operatorname{trace}(\mathbf{D}_{i}^\top \textbf{D}_j)}{\|\mathbf{D}_i\|_F \|\mathbf{D}_{j}\|_F},
\eal
where $i, j = 1, \dots, S$. Figure~\ref{fig:correlation_basis_2_qkvo} reveals low pairwise correlation among the learned matrix atoms in the $S=2$ setting across all attention projections (Q, K, V, O), indicating that the dictionary components capture distinct, complementary patterns. As the dictionary size increases ($S \uparrow$), we observe a growing number of correlated atoms, suggesting increasing redundancy within the learned dictionary. This implies a potential for further compression through dictionary sparsification or rank-constrained atom learning. This trend aligns with the results in Table~\ref{tab:ablation_num_basis}, where performance improves with larger $S$, peaking around $S=8$. The initial gains reflect enhanced expressivity, while the onset of correlation at higher $S$ suggests a trade-off between representational capacity and parameter efficiency. In contrast, note that observed redundancy in the learned dictionary positively affects the language modeling abilities of the model (see WikiText perplexity column in Table~\ref{tab:ablation_num_basis}). 

\begin{figure*}[!h]
    \centering
    \begin{minipage}{0.23\textwidth}
        \centering
        \includegraphics[width=\linewidth]{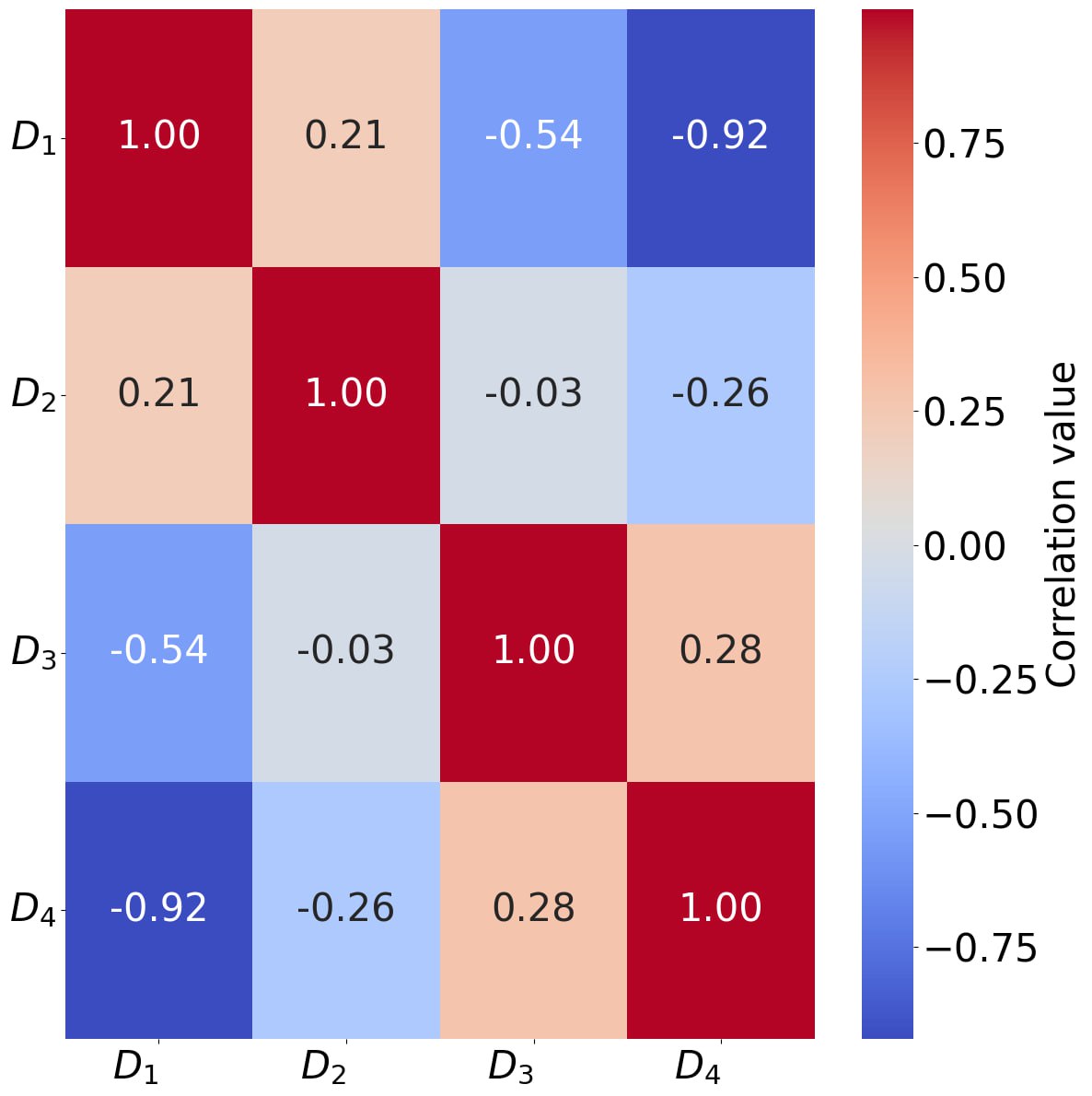}
        \caption*{(A)\quad $D^Q$}
    \end{minipage}
    \hfill
    \begin{minipage}{0.23\textwidth}
        \centering
        \includegraphics[width=\linewidth]{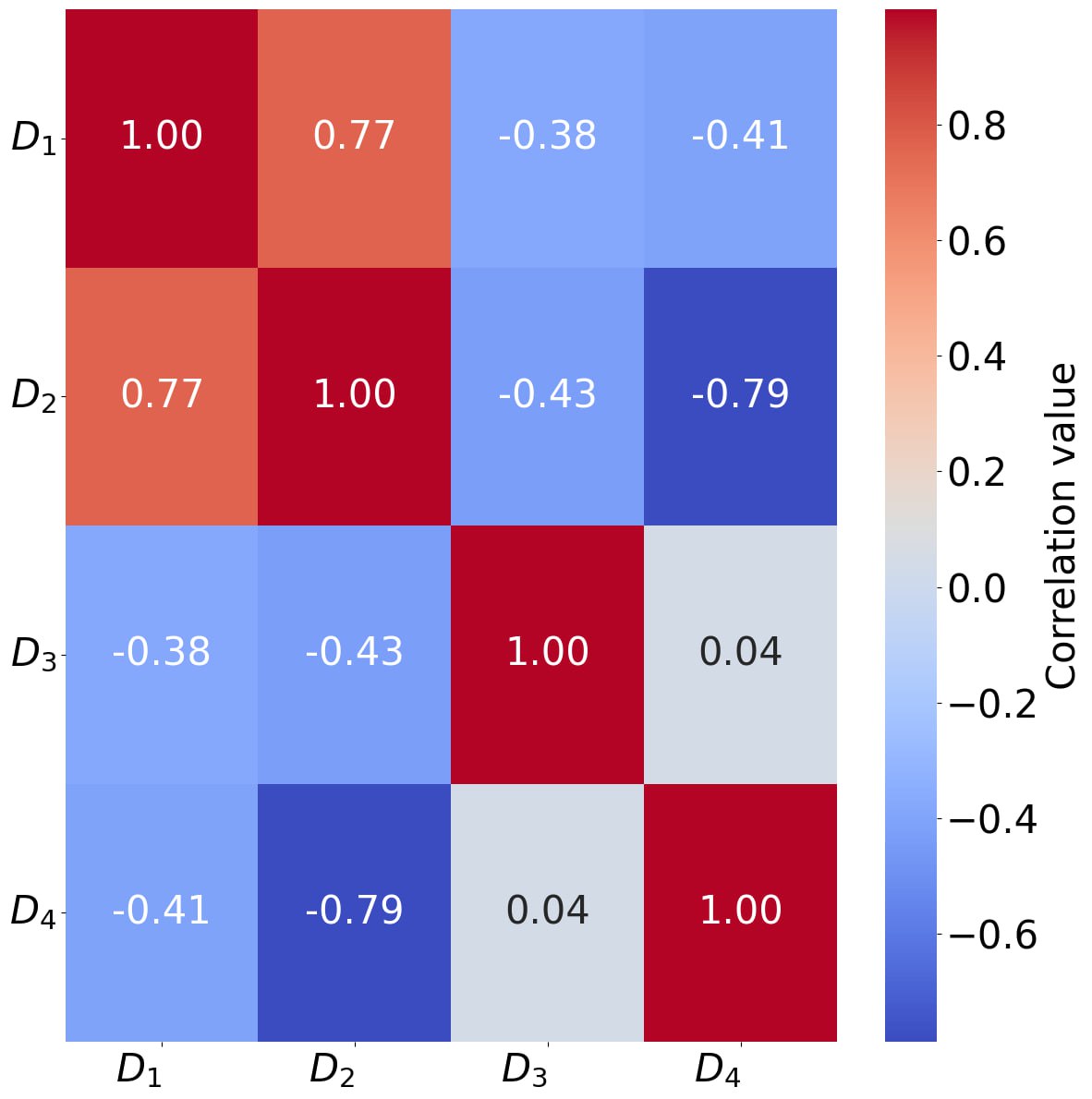}
        \caption*{(B)\quad $D^K$}
    \end{minipage}
    \hfill
    \begin{minipage}{0.23\textwidth}
        \centering
        \includegraphics[width=\linewidth]{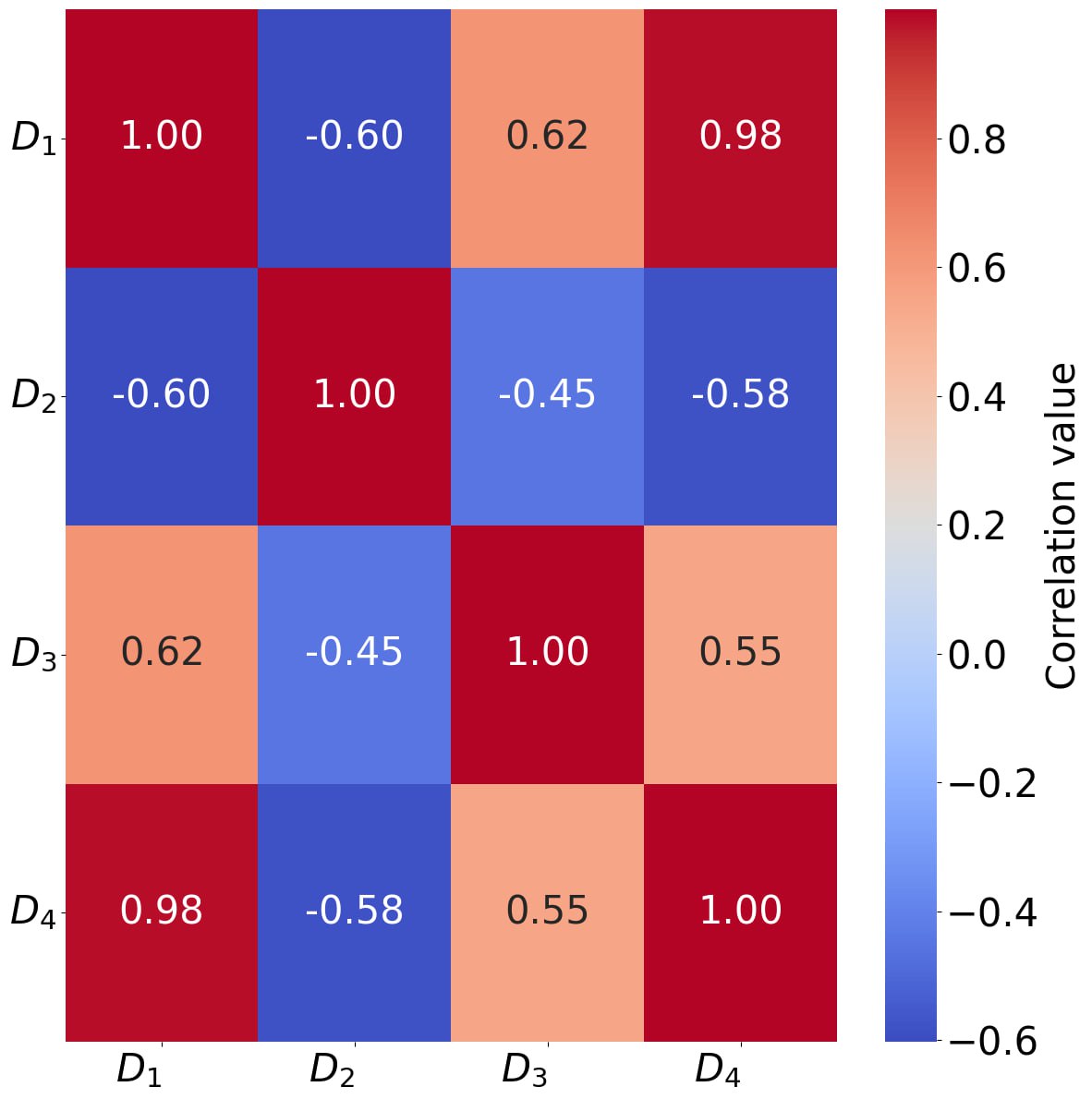}
        \caption*{(C)\quad $D^V$}
    \end{minipage}
    \hfill
    \begin{minipage}{0.23\textwidth}
        \centering
        \includegraphics[width=\linewidth]{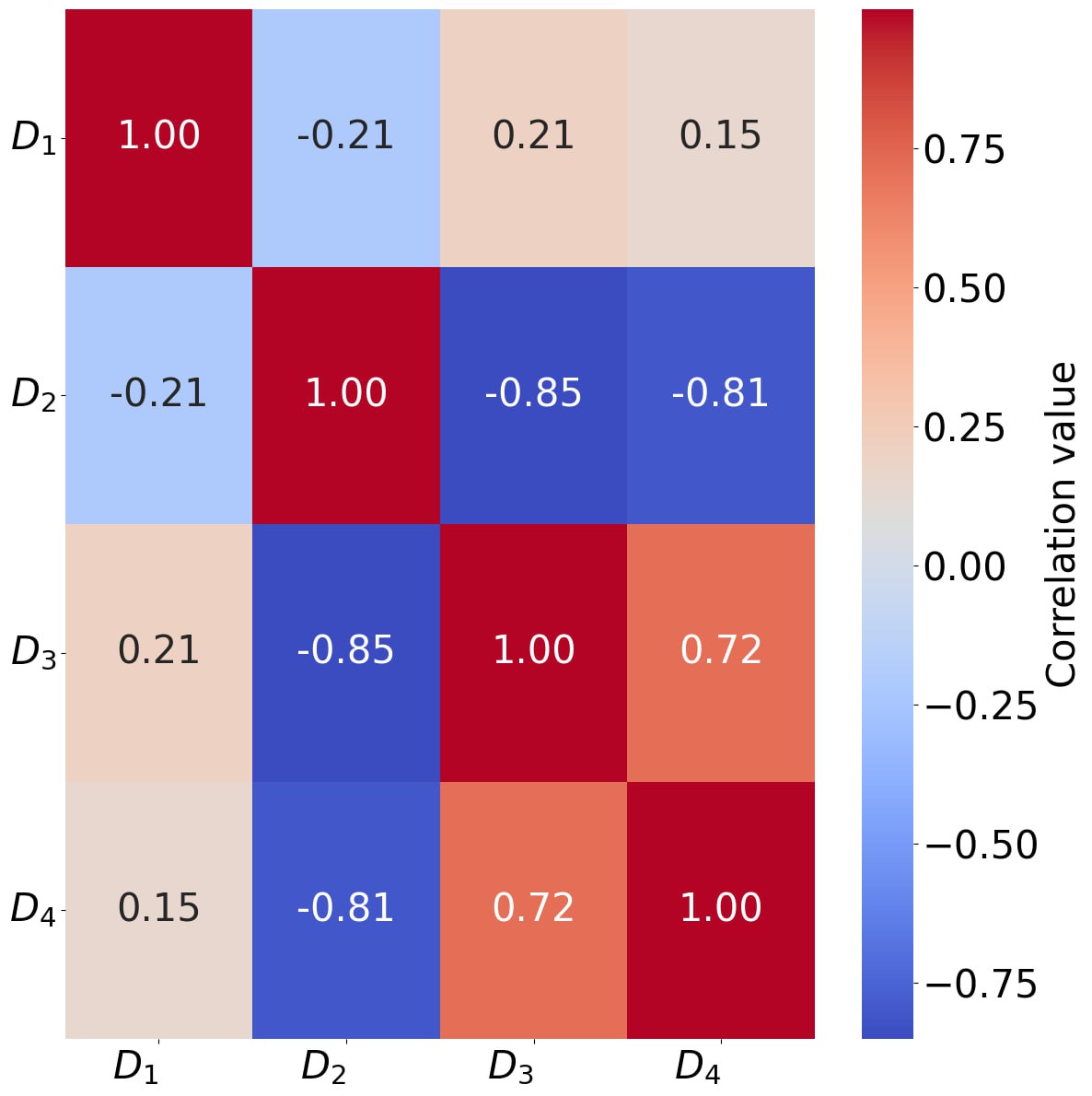}
        \caption*{(D)\quad $D^O$}
    \end{minipage}
    \caption{Cosine similarity between atoms in each dictionary for Q, K, V, and O projections (left to right). Higher absolute values indicate stronger atom correlations. Results shown for MASA-QKVO (small transformer, S=4)}
    \label{fig:correlation_basis_4_qkvo}
    \vspace{-3mm}
\end{figure*}

\begin{figure*}[!h]
    \centering
    \begin{minipage}{0.23\textwidth}
        \centering
        \includegraphics[width=\linewidth]{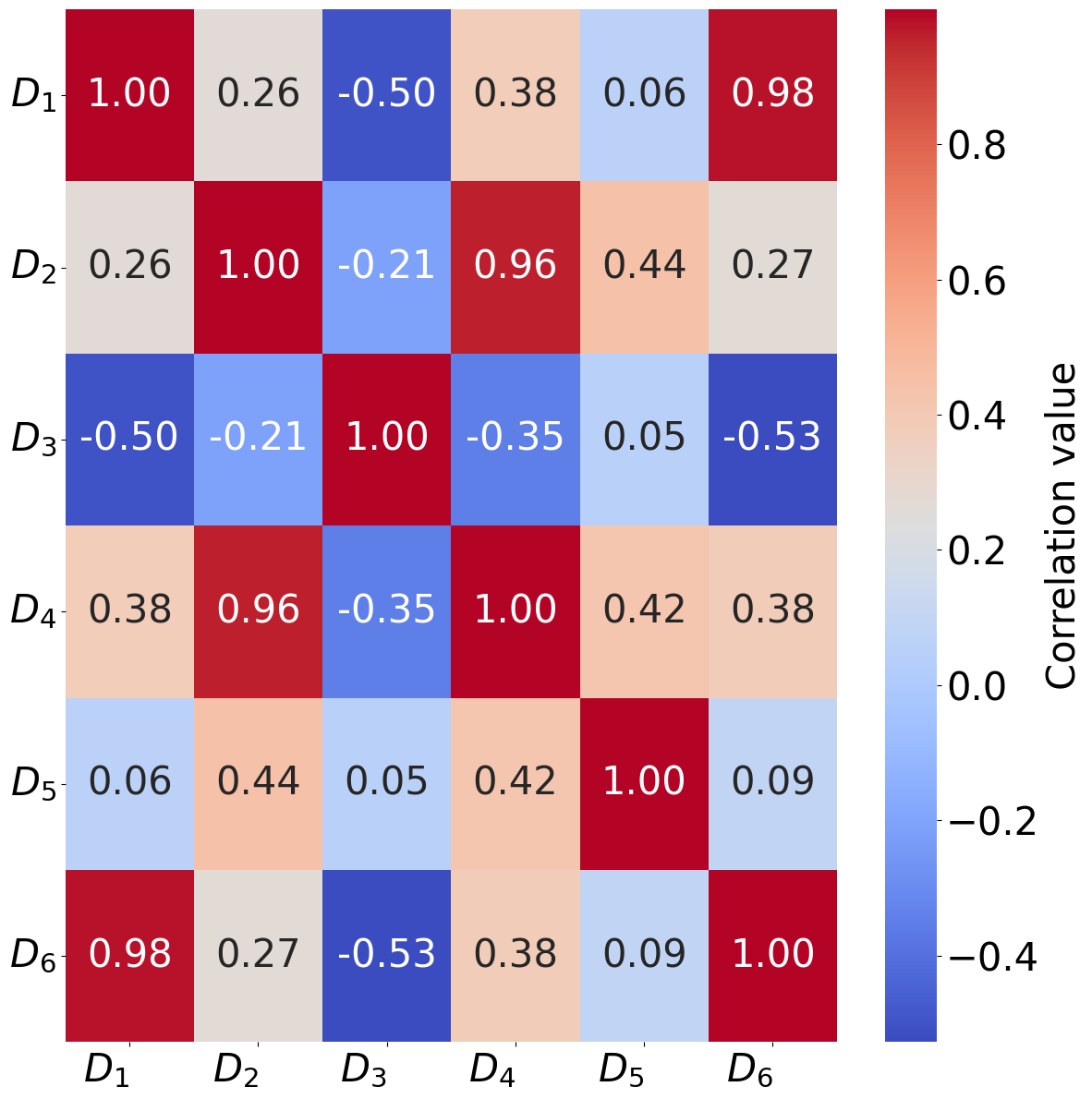}
        \caption*{(A)\quad $D^Q$}
    \end{minipage}
    \hfill
    \begin{minipage}{0.23\textwidth}
        \centering
        \includegraphics[width=\linewidth]{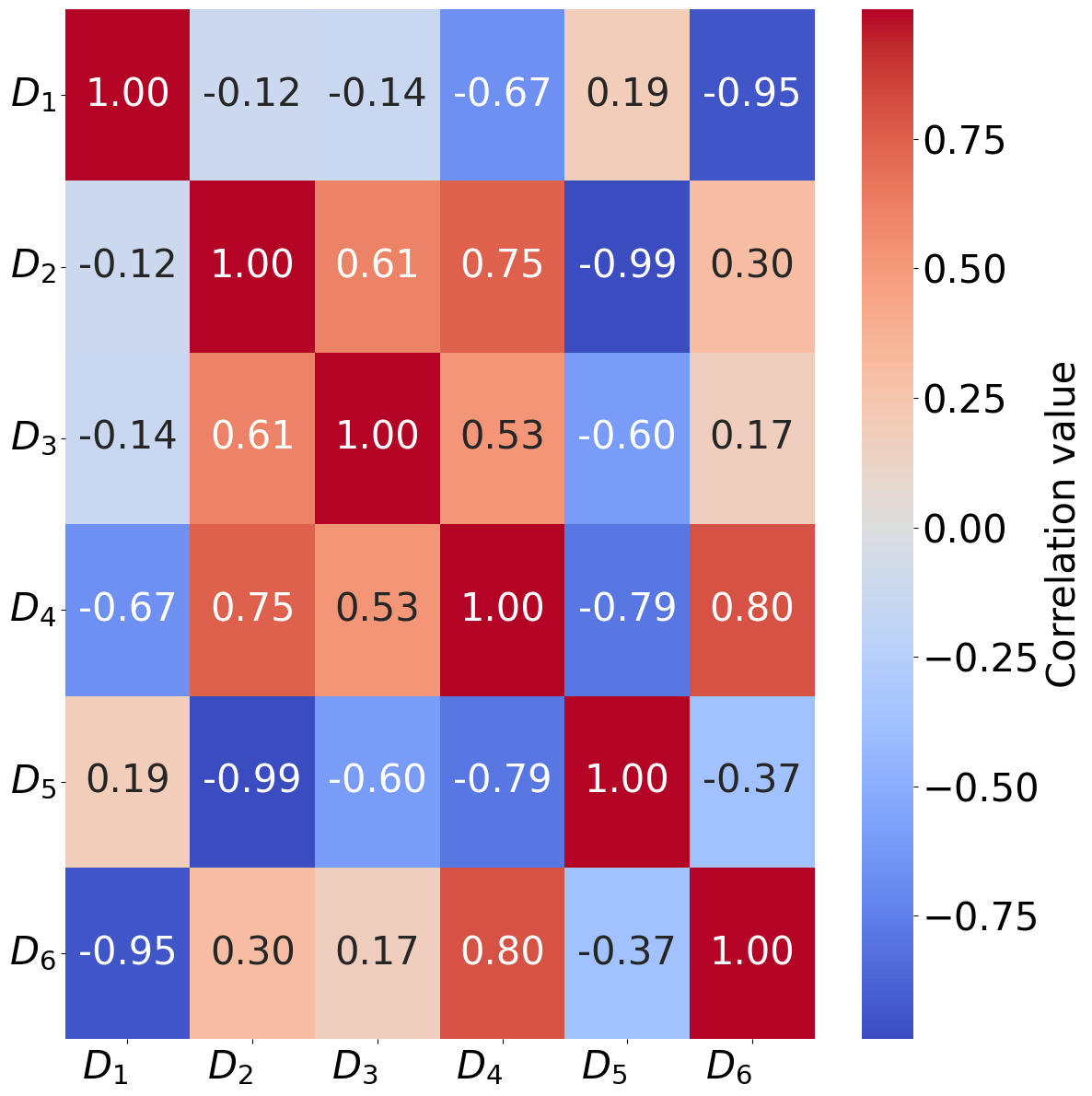}
        \caption*{(B)\quad $D^K$}
    \end{minipage}
    \hfill
    \begin{minipage}{0.23\textwidth}
        \centering
        \includegraphics[width=\linewidth]{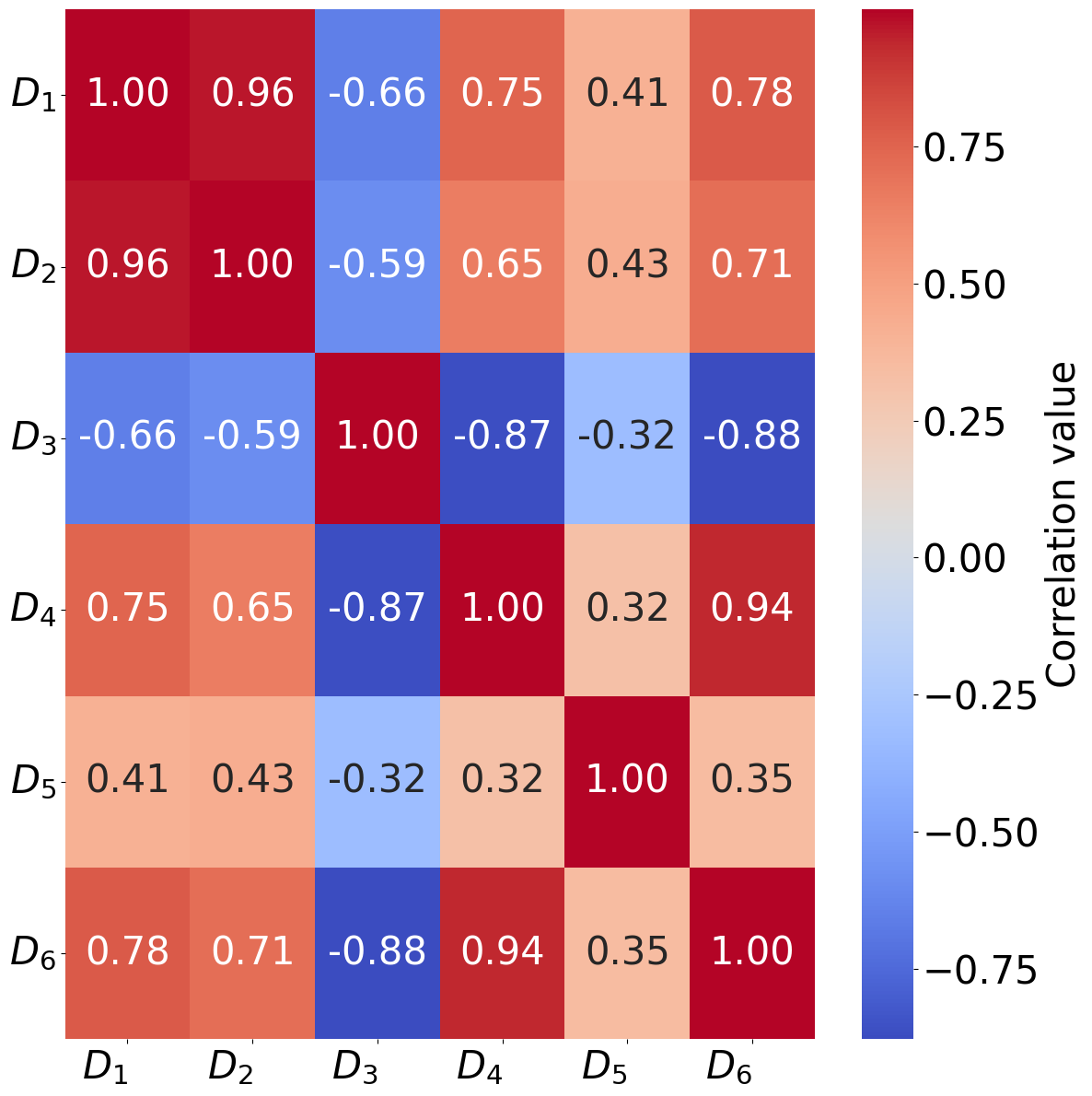}
        \caption*{(C)\quad $D^V$}
    \end{minipage}
    \hfill
    \begin{minipage}{0.23\textwidth}
        \centering
        \includegraphics[width=\linewidth]{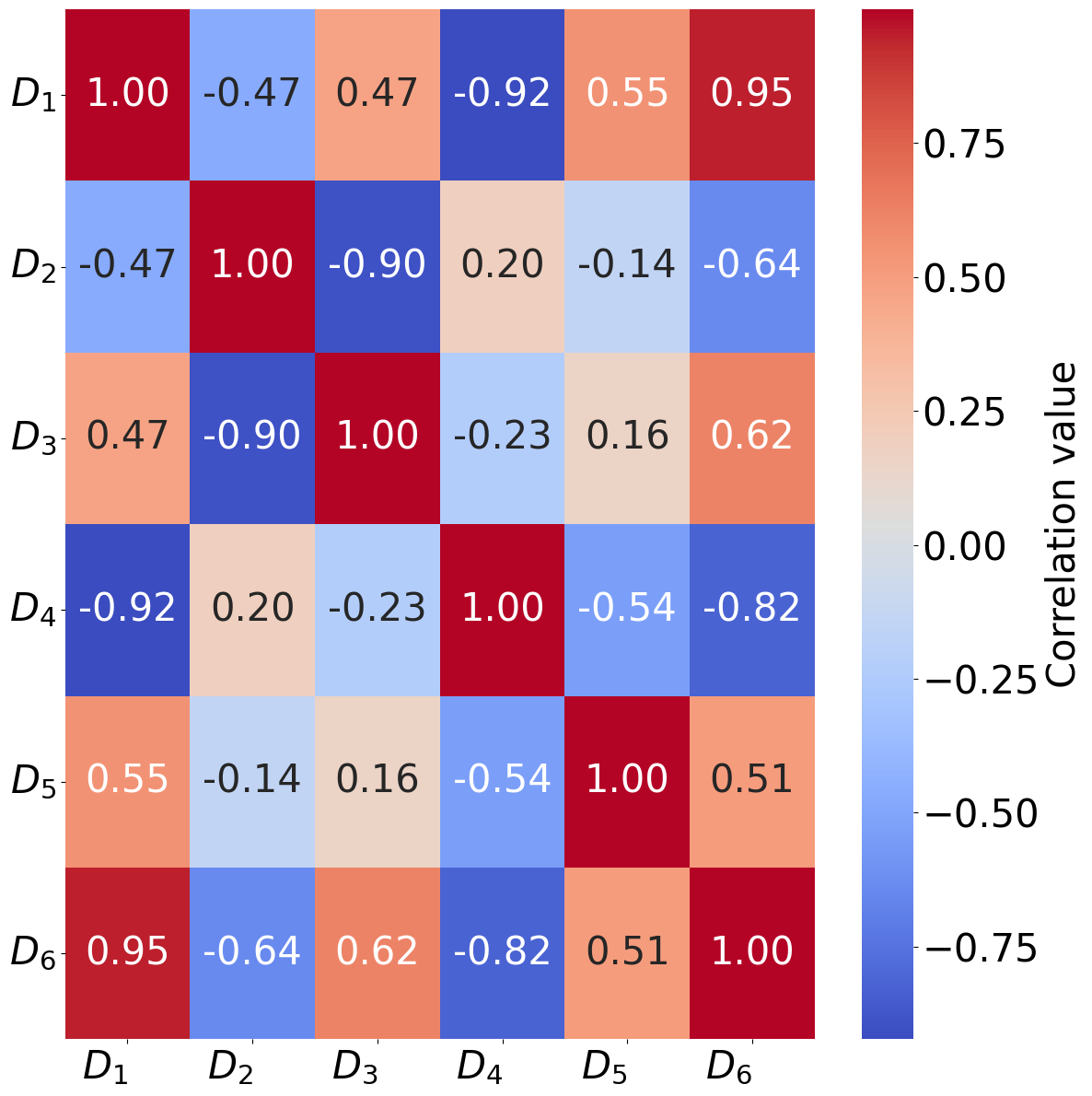}
        \caption*{(D)\quad $D^O$}
    \end{minipage}
    \caption{Cosine similarity between atoms in each dictionary for Q, K, V, and O projections (left to right). Higher absolute values indicate stronger atom correlations. Results shown for MASA-QKVO (small transformer, S=6)}
    \label{fig:correlation_basis_6_qkvo}
    \vspace{-3mm}
\end{figure*}

\begin{figure*}[!h]
    \centering
    \begin{minipage}{0.23\textwidth}
        \centering
        \includegraphics[width=\linewidth]{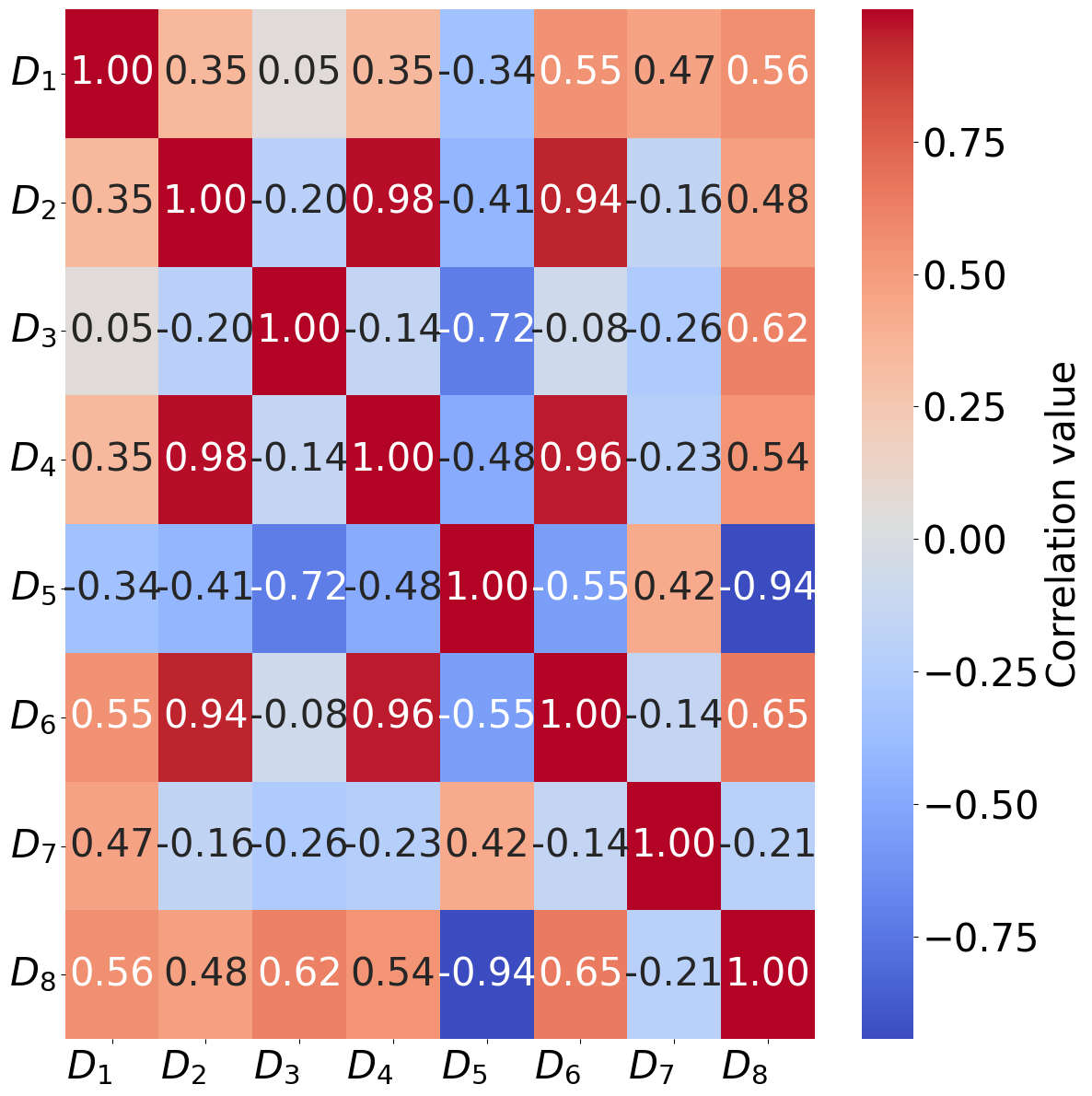}
        \caption*{(A)\quad $D^Q$}
    \end{minipage}
    \hfill
    \begin{minipage}{0.23\textwidth}
        \centering
        \includegraphics[width=\linewidth]{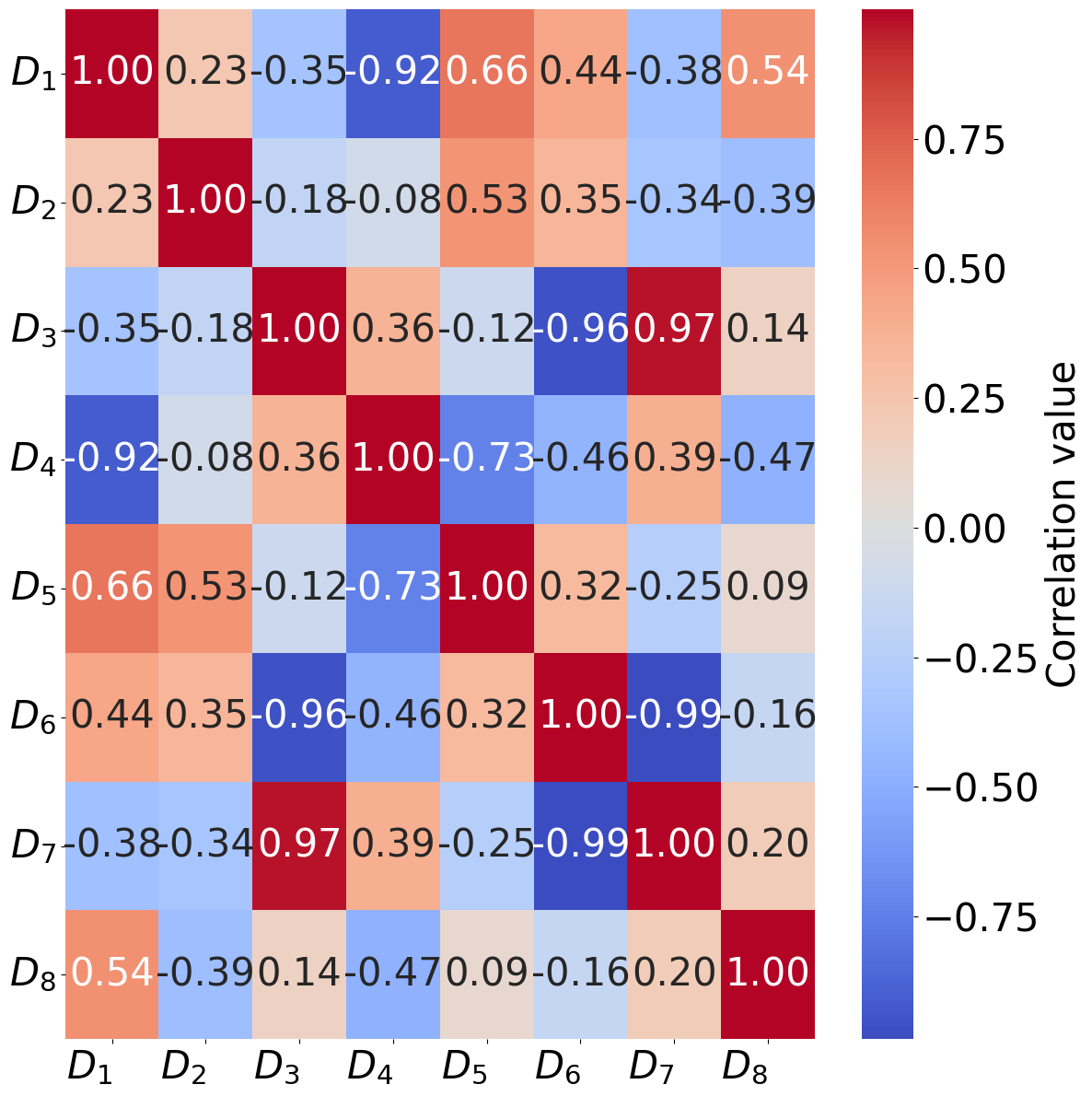}
        \caption*{(B)\quad $D^K$}
    \end{minipage}
    \hfill
    \begin{minipage}{0.23\textwidth}
        \centering
        \includegraphics[width=\linewidth]{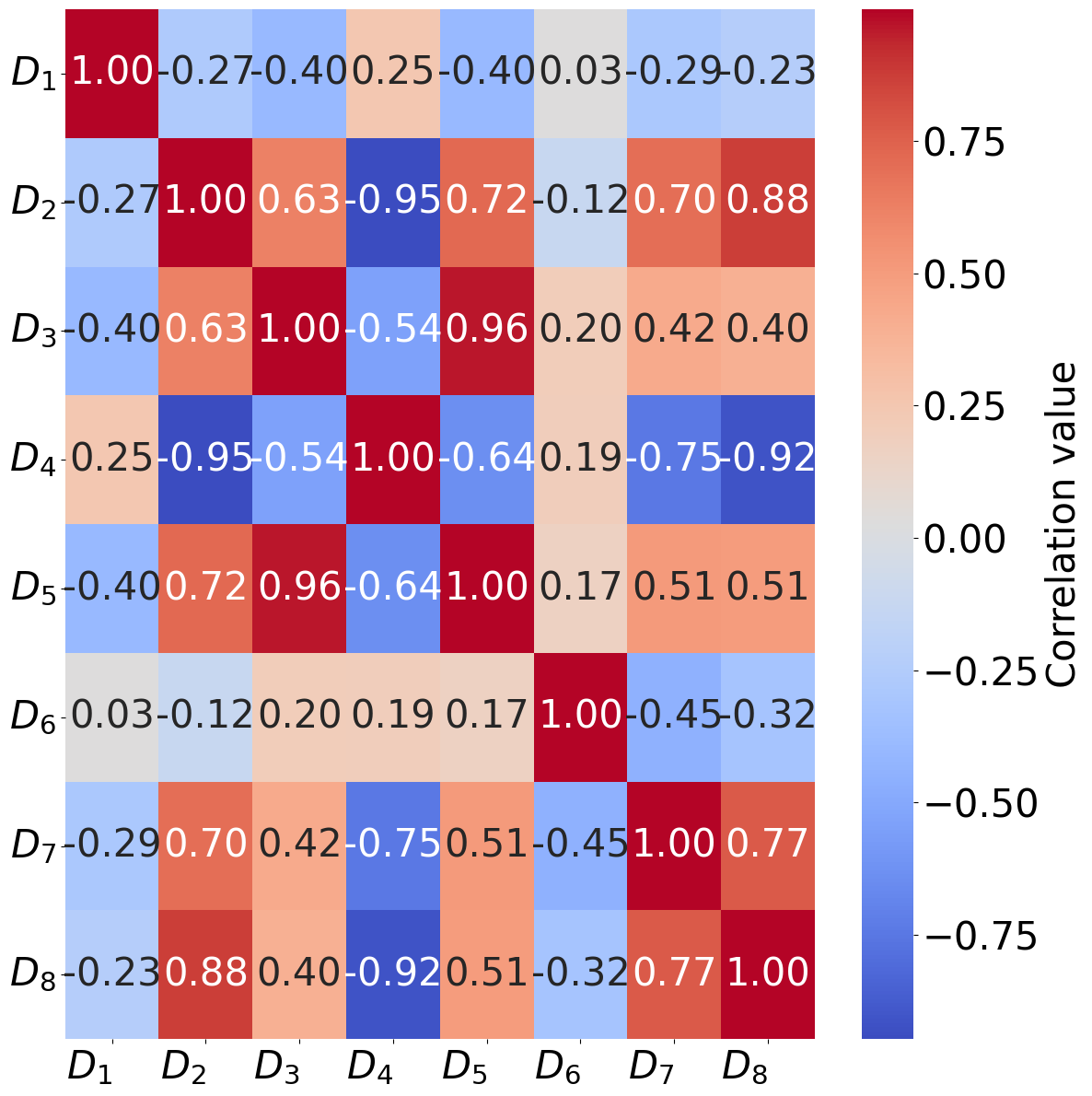}
        \caption*{(C)\quad $D^V$}
    \end{minipage}
    \hfill
    \begin{minipage}{0.23\textwidth}
        \centering
        \includegraphics[width=\linewidth]{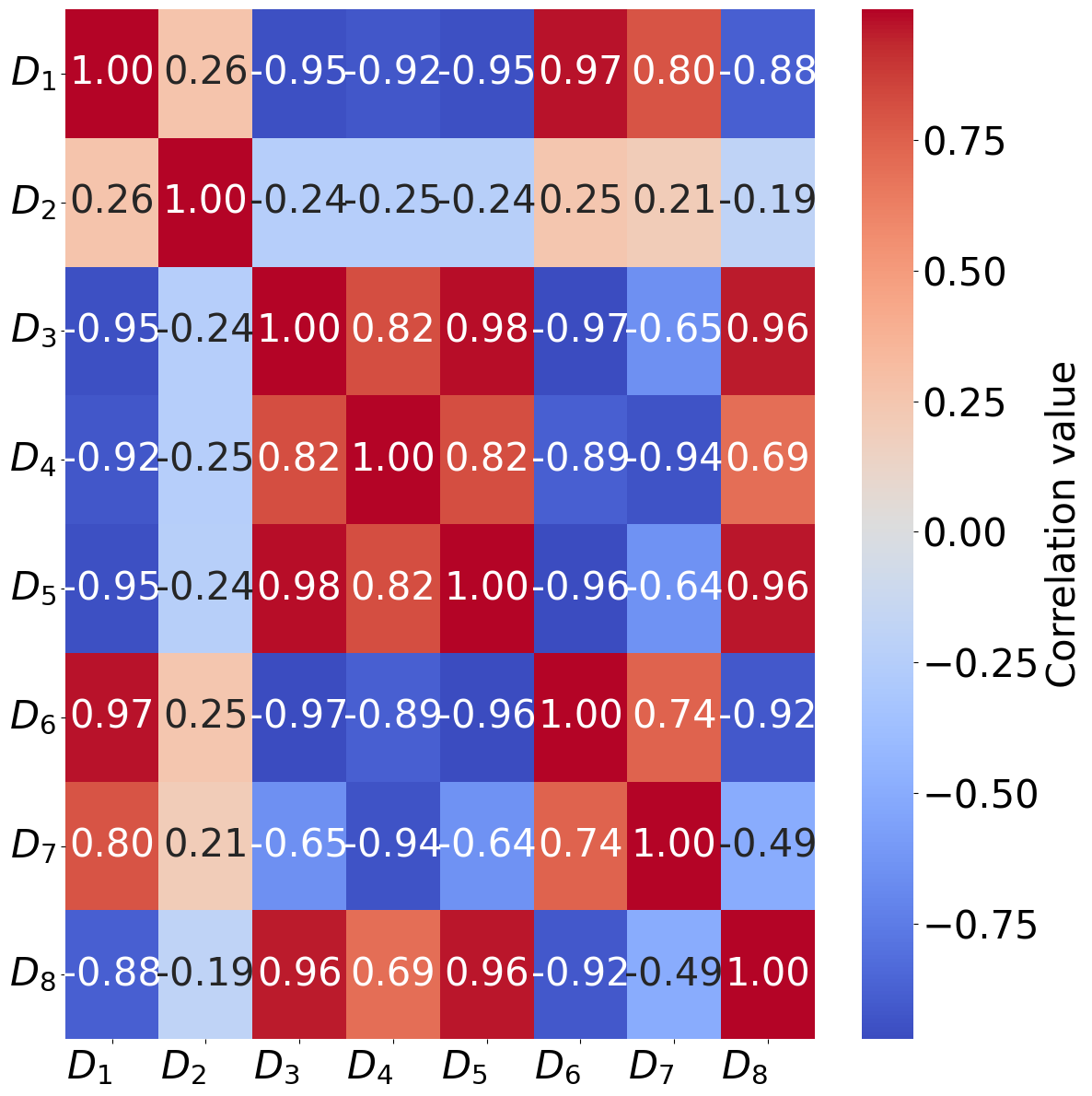}
        \caption*{(D)\quad $D^O$}
    \end{minipage}
    \caption{Cosine similarity between atoms in each dictionary for Q, K, V, and O projections (left to right). Higher absolute values indicate stronger atom correlations. Results shown for MASA-QKVO (small transformer, S=8)}
    \label{fig:correlation_basis_8_qkvo}
    \vspace{-3mm}
\end{figure*}

\subsection{Visualization of Mixing Coefficients}
\label{sec:ablation_weight_scalars}
\begin{figure*}[!h]
    \centering
    \begin{minipage}{0.23\textwidth}
        \centering
        \includegraphics[width=\linewidth]{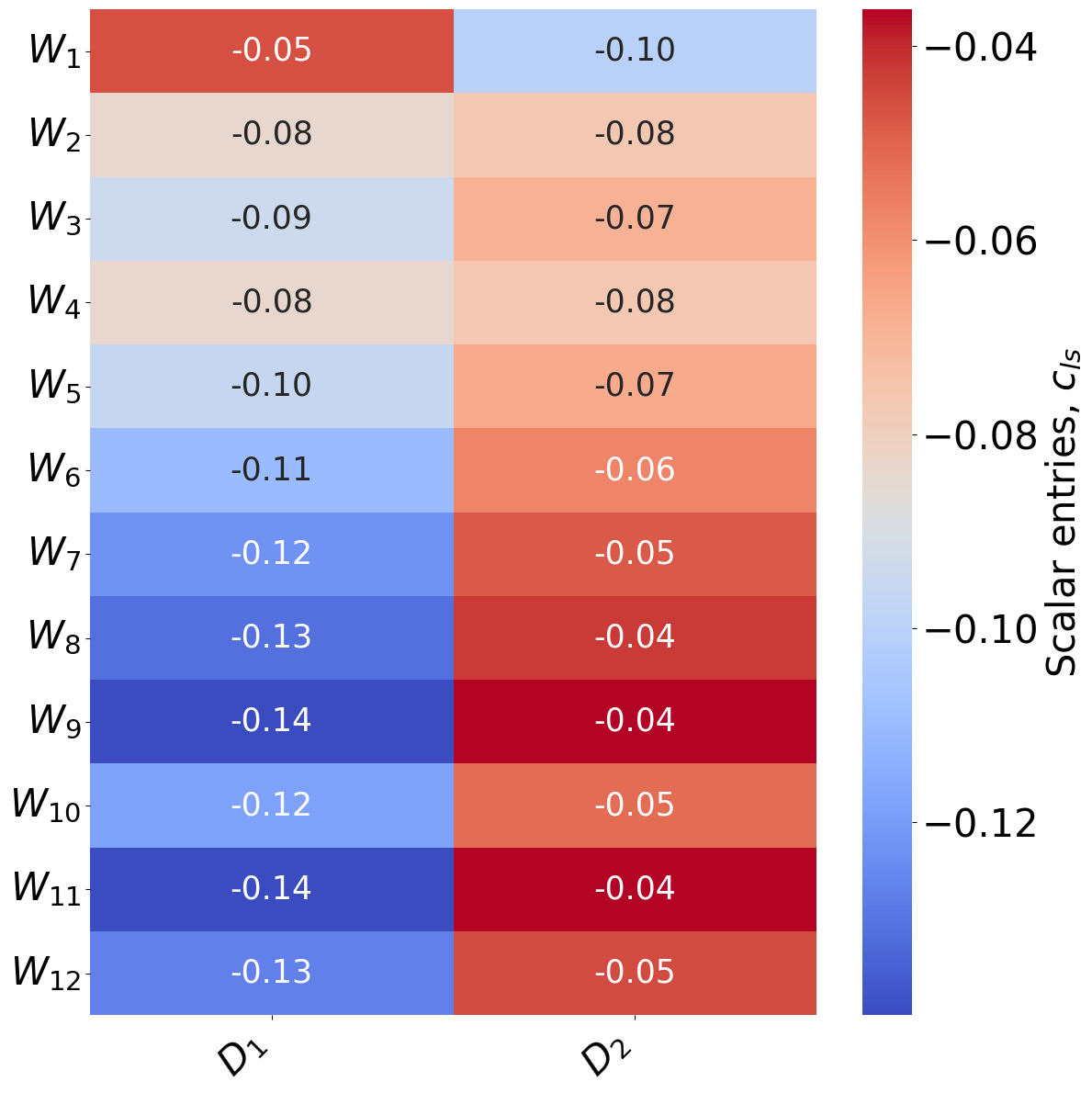}
        \caption*{(A)\quad $Query$}
    \end{minipage}
    \hfill
    \begin{minipage}{0.23\textwidth}
        \centering
        \includegraphics[width=\linewidth]{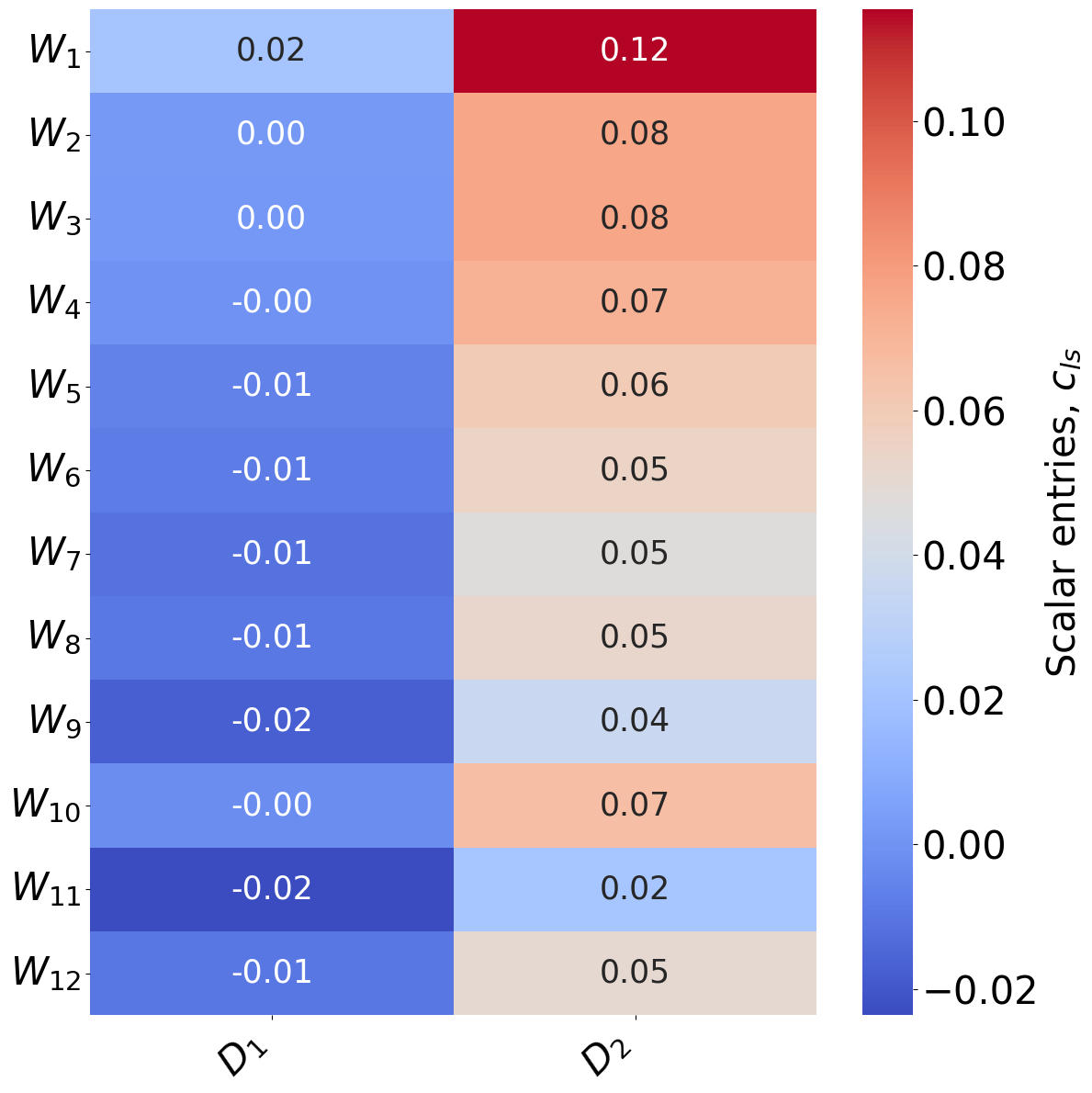}
        \caption*{(B)\quad $Key$}
    \end{minipage}
    \hfill
    \begin{minipage}{0.23\textwidth}
        \centering
        \includegraphics[width=\linewidth]{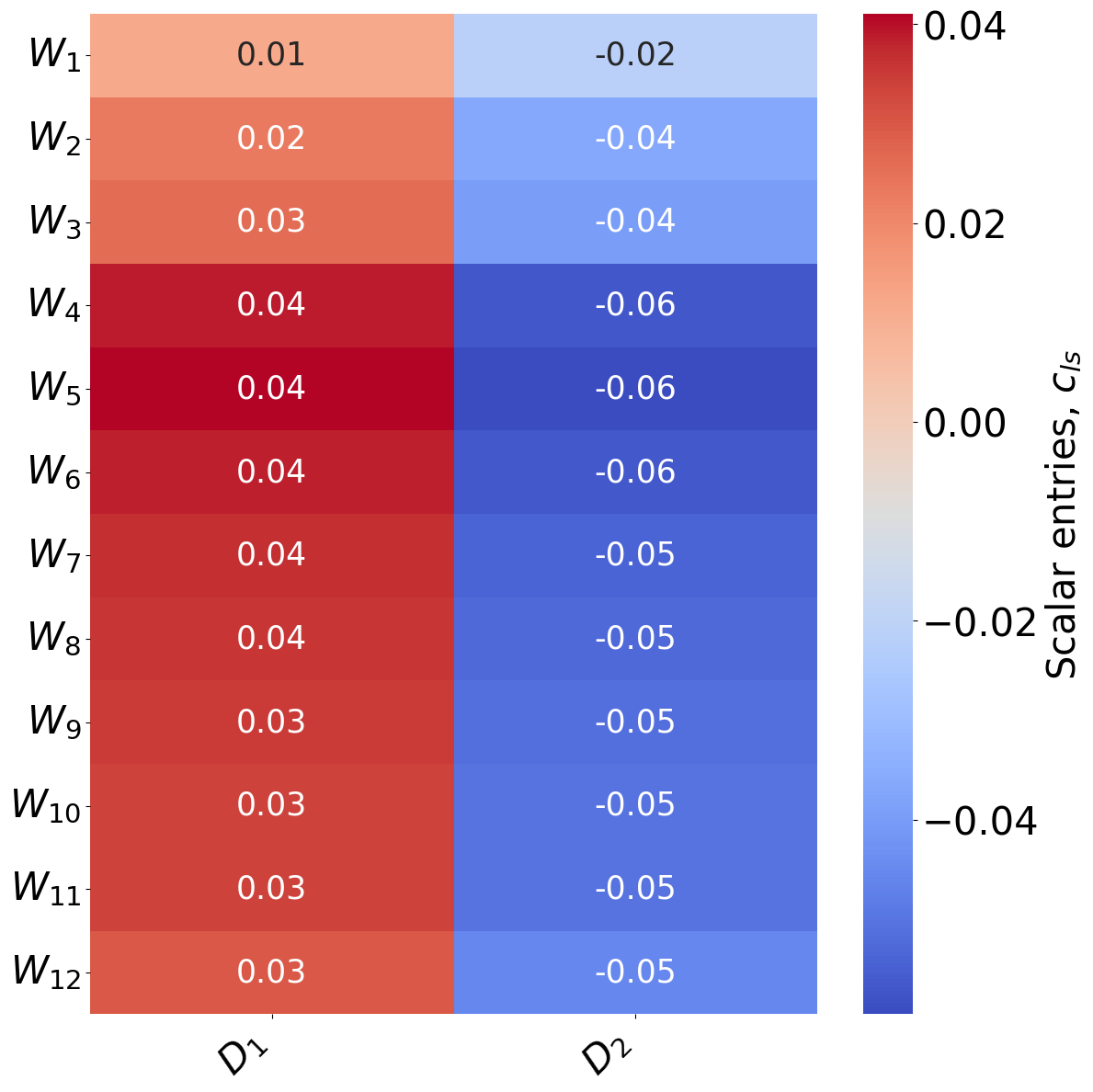}
        \caption*{(C)\quad $Values$}
    \end{minipage}
    \hfill
    \begin{minipage}{0.23\textwidth}
        \centering
        \includegraphics[width=\linewidth]{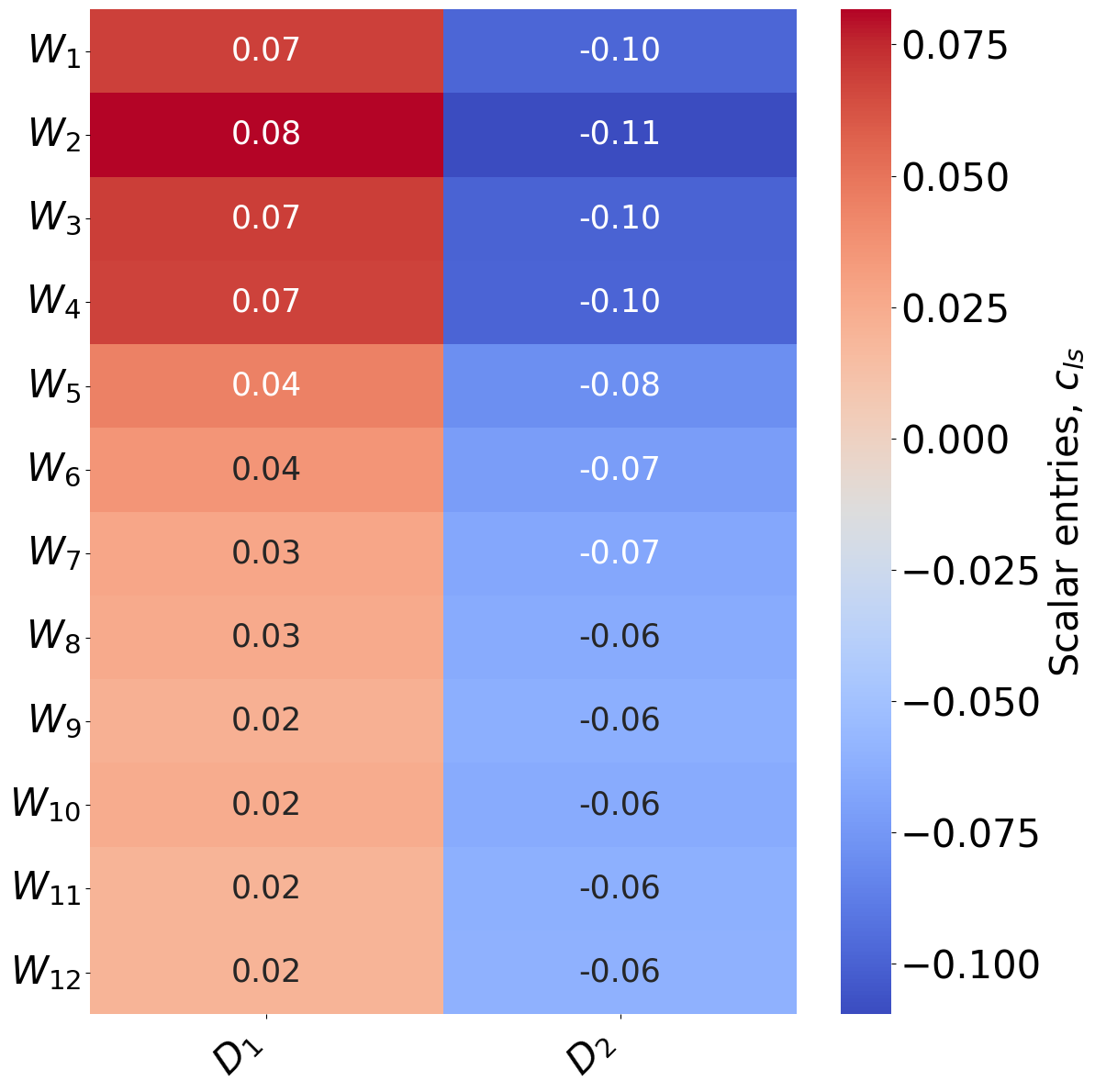}
        \caption*{(D)\quad $Out$}
    \end{minipage}
    \caption{Weight coefficients $\m C$ for each layer and atom in each dictionary for Q, K, V, and O projections (left to right). Results shown for MASA-QKVO (small transformer, S=2)}
    \label{fig:correlation_weights_2_qkvo}
    \vspace{-3mm}
\end{figure*}

\begin{figure*}[!h]
    \centering
    \begin{minipage}{0.23\textwidth}
        \centering
        \includegraphics[width=\linewidth]{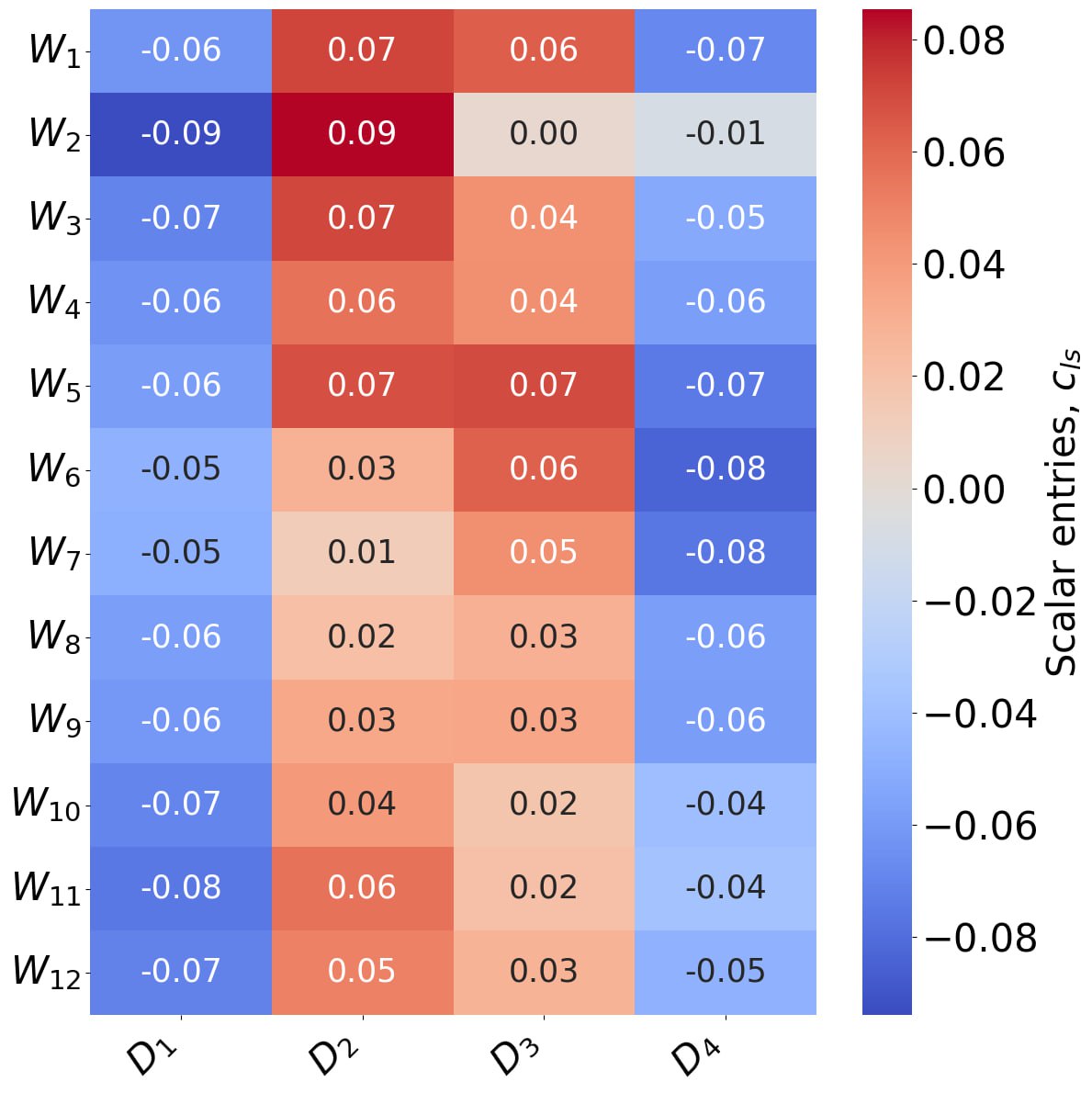}
        \caption*{(A)\quad $Query$}
    \end{minipage}
    \hfill
    \begin{minipage}{0.23\textwidth}
        \centering
        \includegraphics[width=\linewidth]{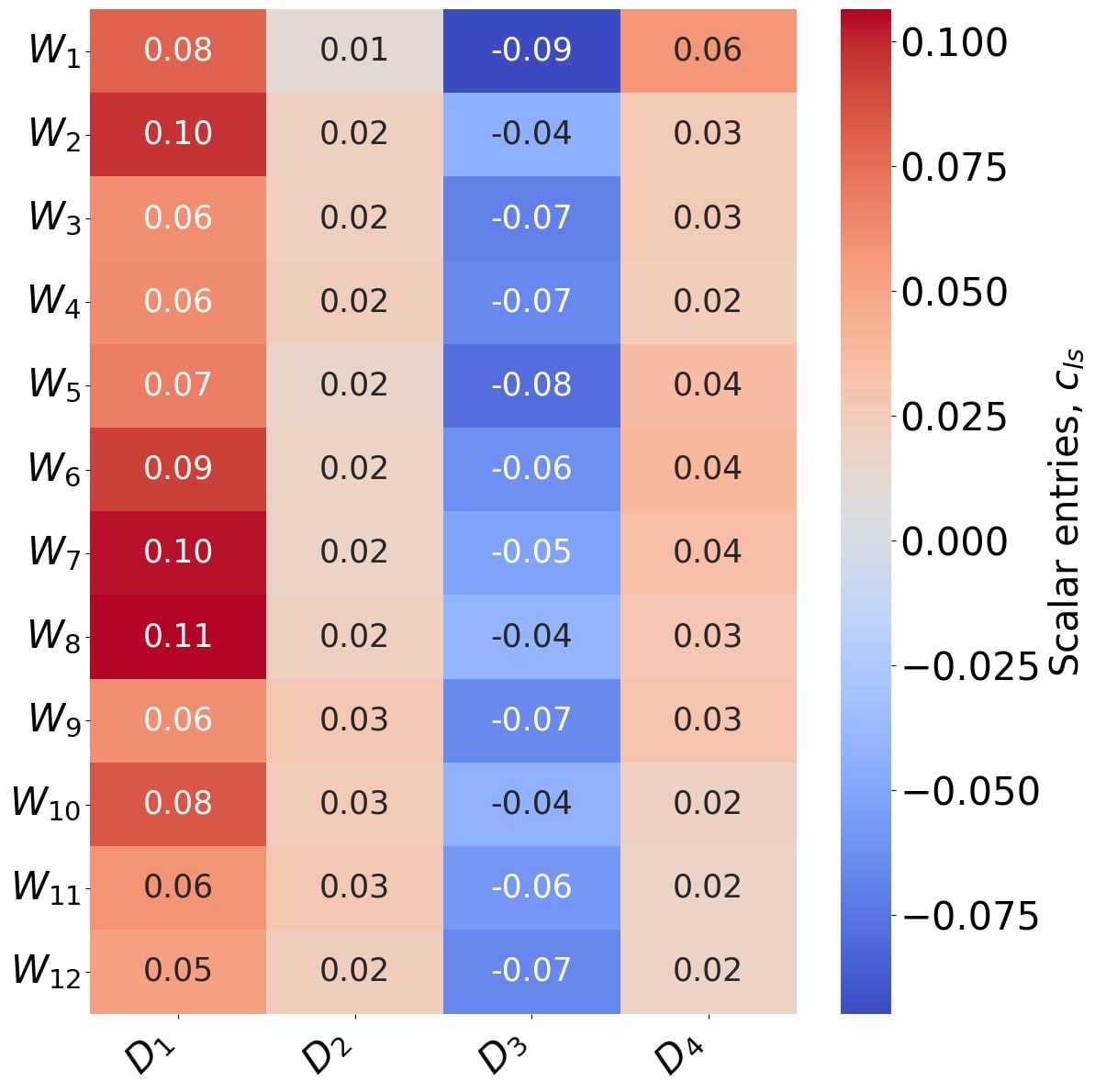}
        \caption*{(B)\quad $Key$}
    \end{minipage}
    \hfill
    \begin{minipage}{0.23\textwidth}
        \centering
        \includegraphics[width=\linewidth]{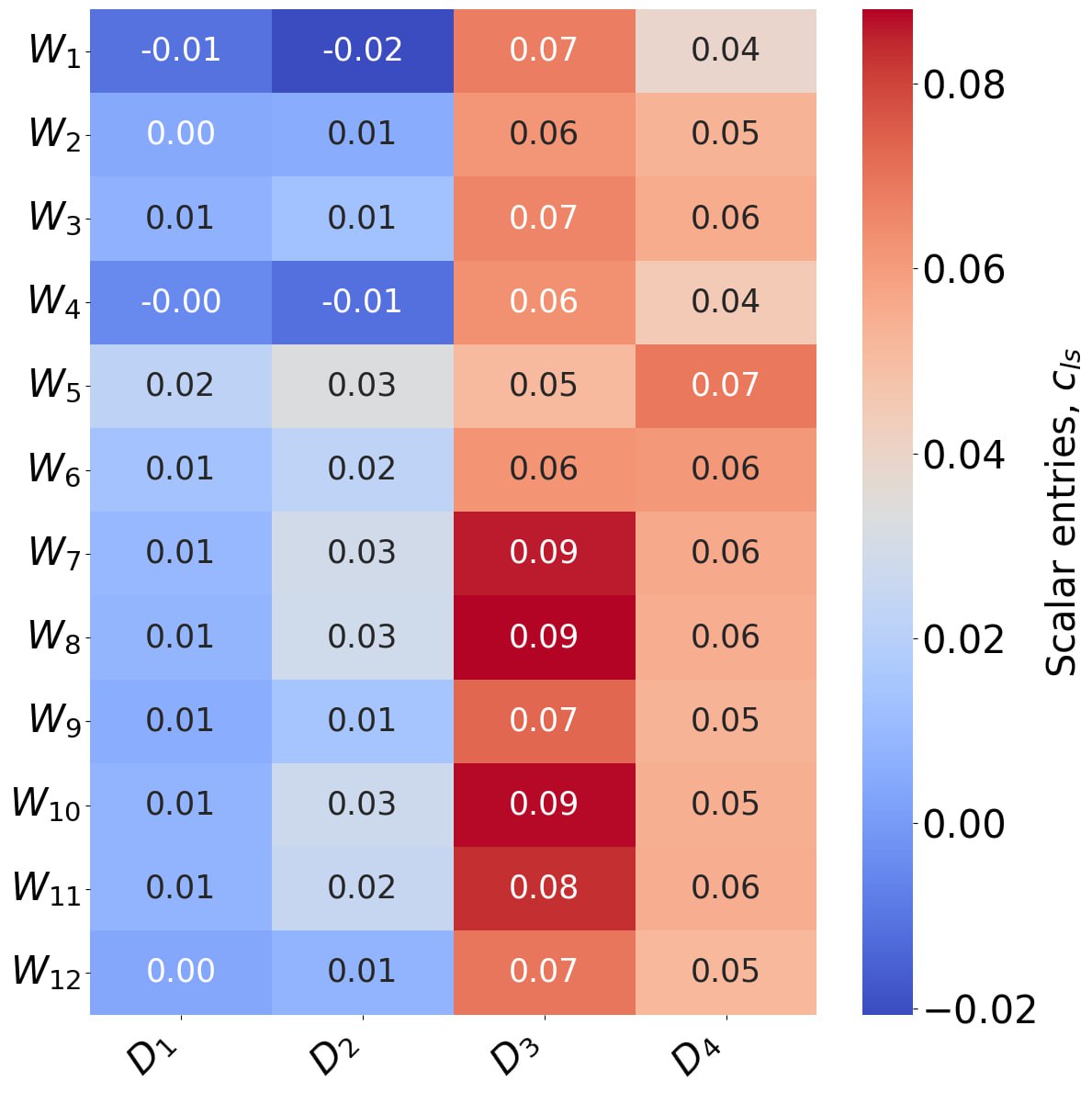}
        \caption*{(C)\quad $Value$}
    \end{minipage}
    \hfill
    \begin{minipage}{0.23\textwidth}
        \centering
        \includegraphics[width=\linewidth]{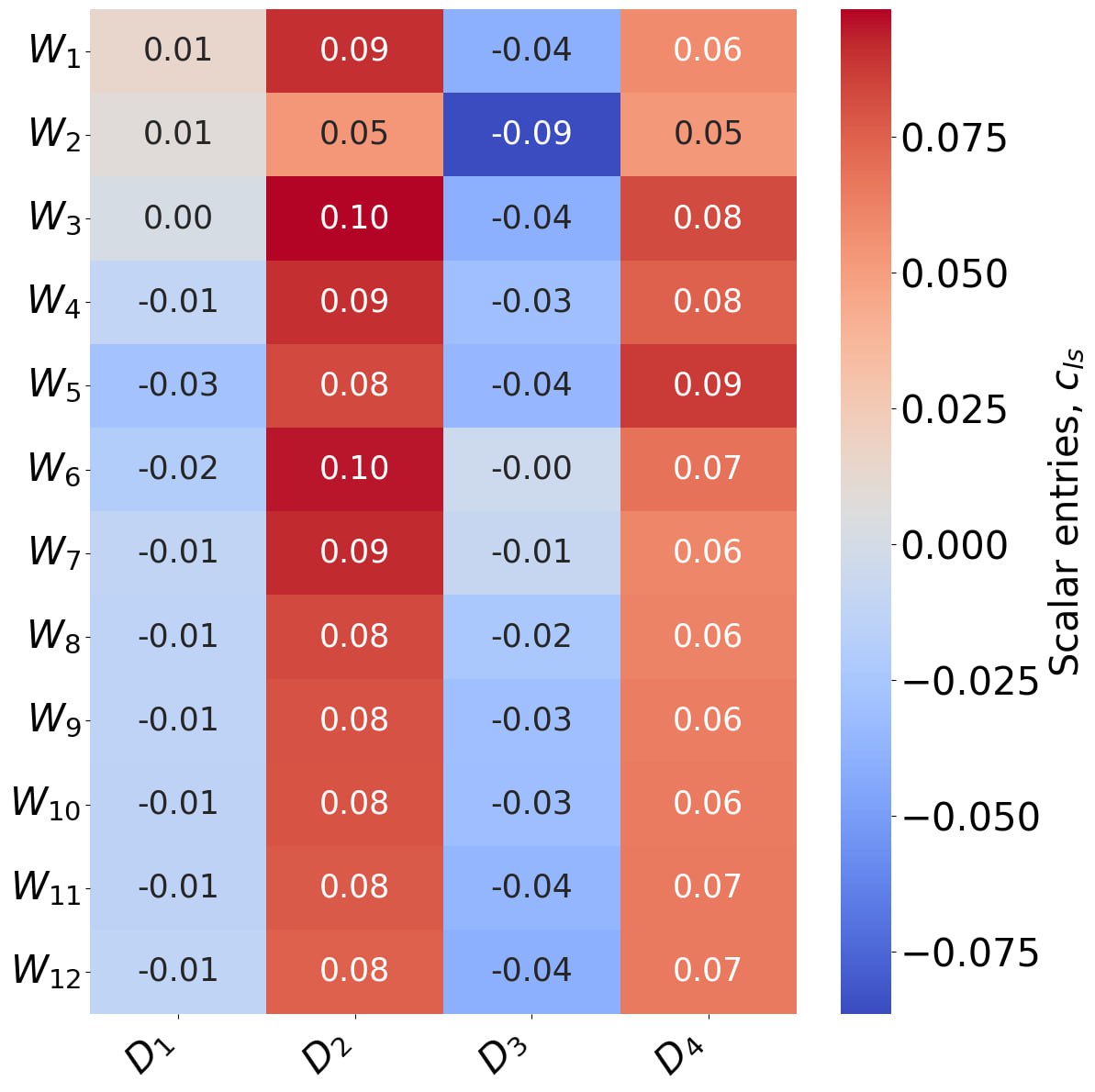}
        \caption*{(D)\quad $Out$}
    \end{minipage}
    \caption{Weight coefficients $\m C$ for each layer and atom in each dictionary for Q, K, V, and O projections (left to right). Results shown for MASA-QKVO (small transformer, S=4)}
    \label{fig:correlation_weights_4_qkvo}
    \vspace{-3mm}
\end{figure*}

\begin{figure*}[!h]
    \centering
    \begin{minipage}{0.23\textwidth}
        \centering
        \includegraphics[width=\linewidth]{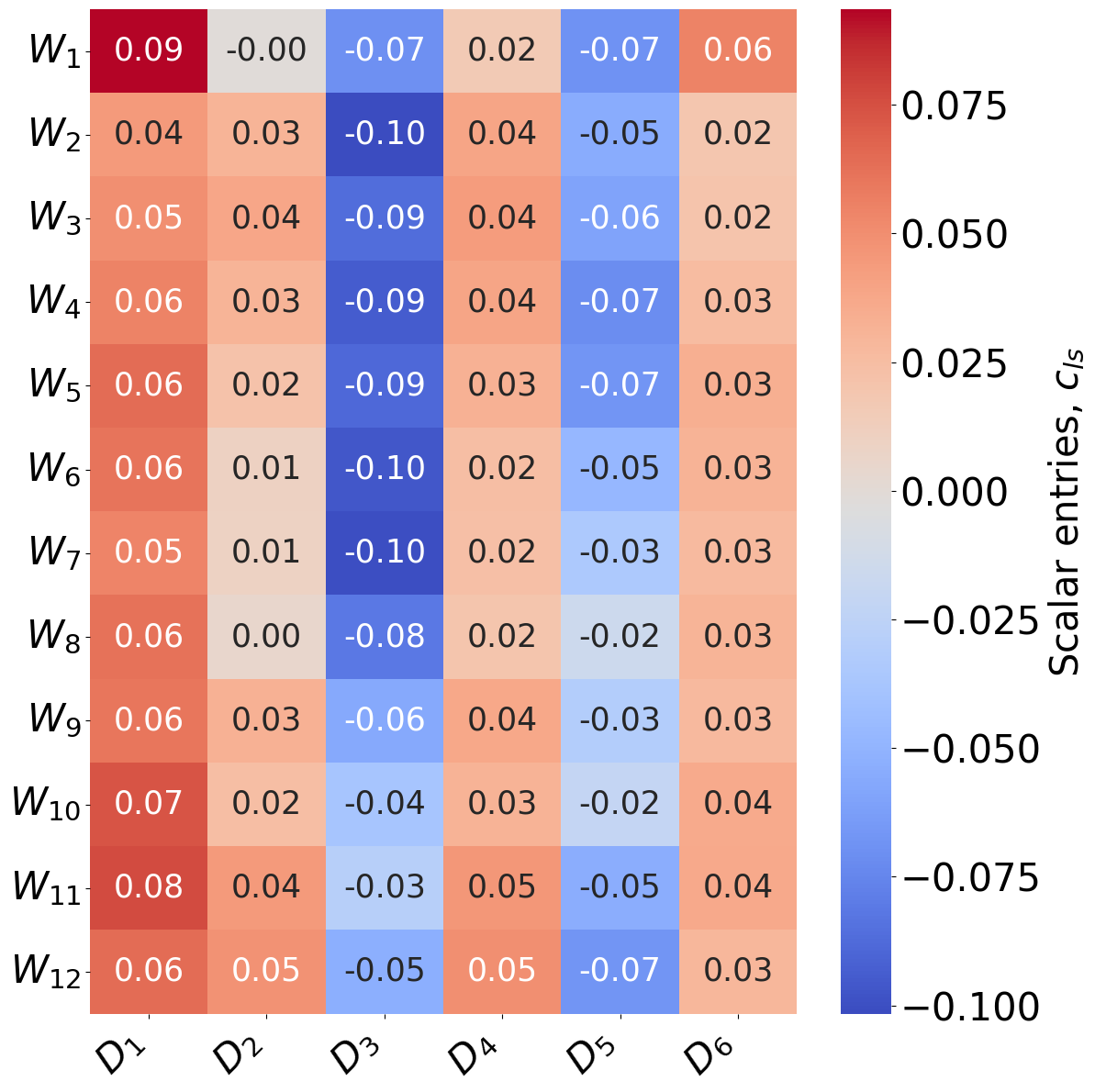}
        \caption*{(A)\quad $Query$}
    \end{minipage}
    \hfill
    \begin{minipage}{0.23\textwidth}
        \centering
        \includegraphics[width=\linewidth]{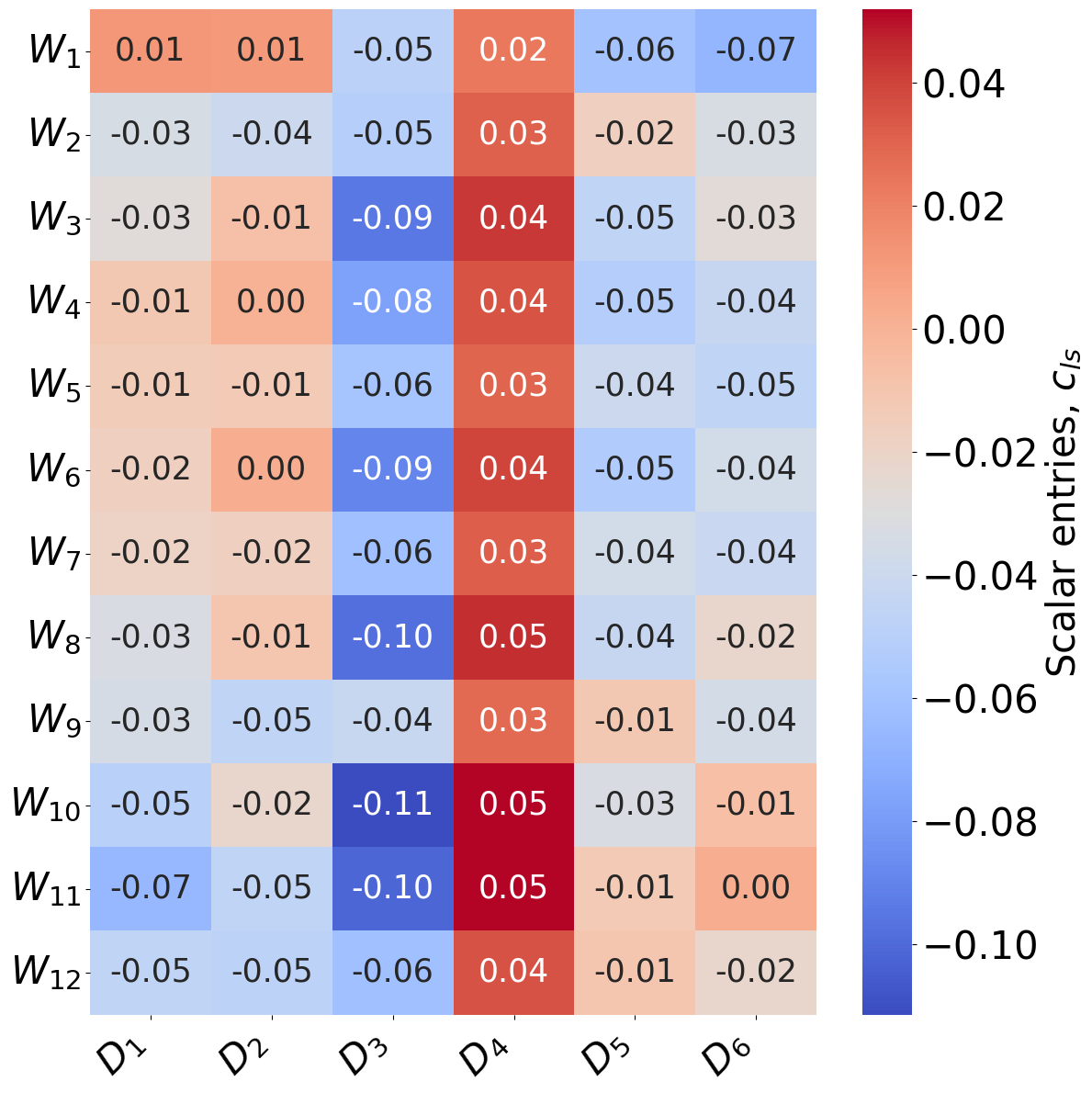}
        \caption*{(B)\quad $Key$}
    \end{minipage}
    \hfill
    \begin{minipage}{0.23\textwidth}
        \centering
        \includegraphics[width=\linewidth]{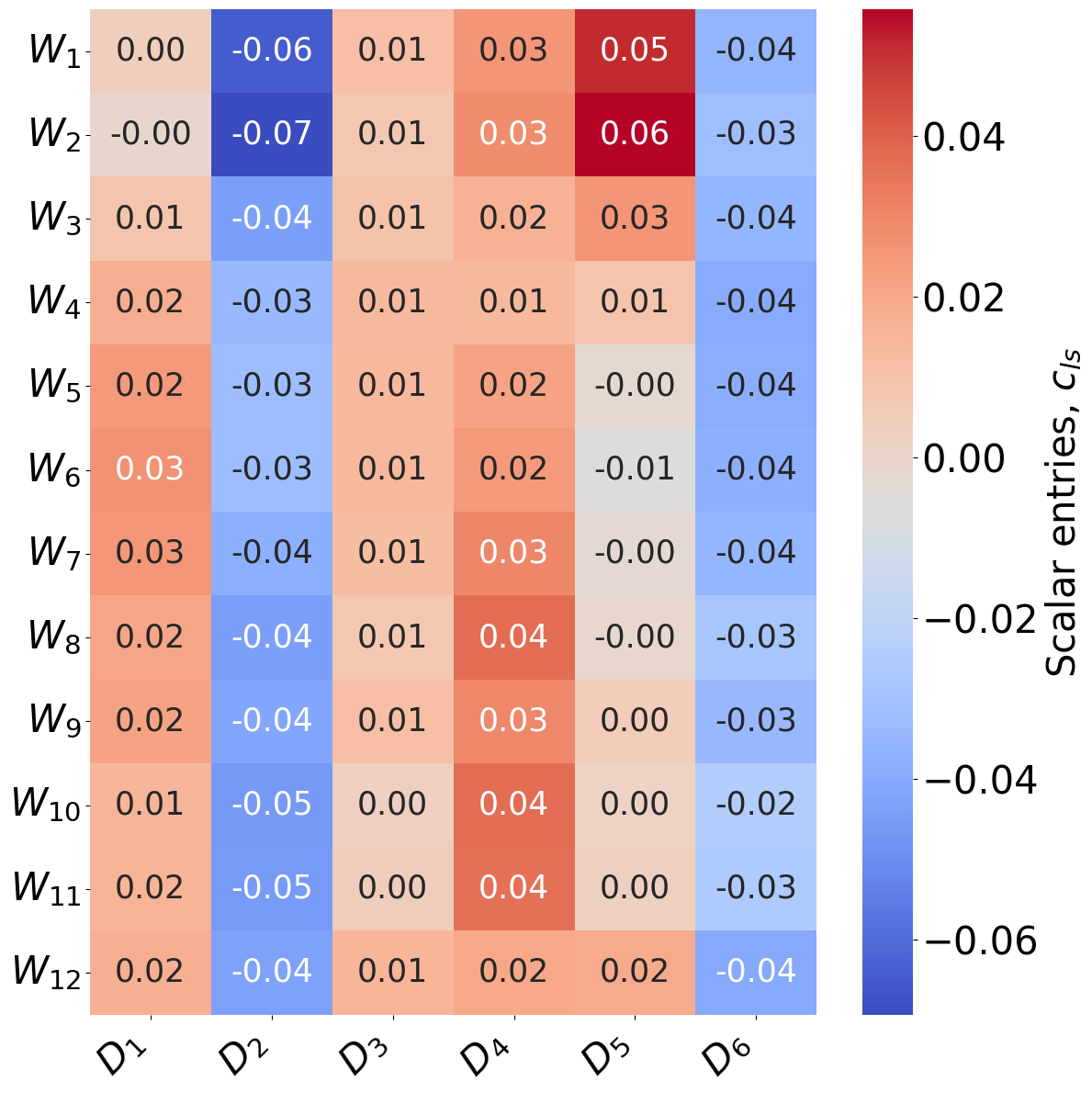}
        \caption*{(C)\quad $Value$}
    \end{minipage}
    \hfill
    \begin{minipage}{0.23\textwidth}
        \centering
        \includegraphics[width=\linewidth]{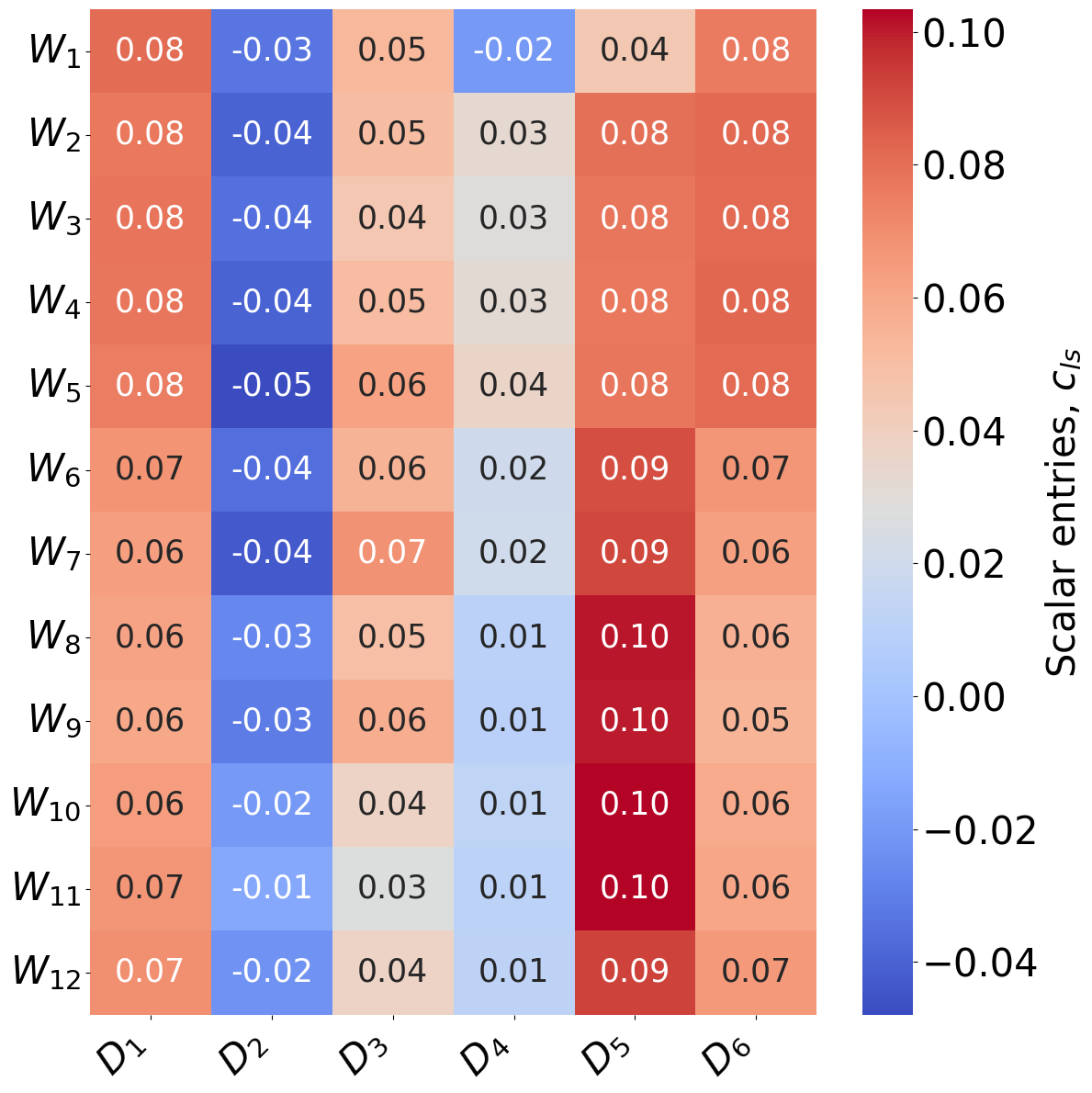}
        \caption*{(D)\quad $Out$}
    \end{minipage}
    \caption{Weight coefficients $\m C$ for each layer and atom in each dictionary for Q, K, V, and O projections (left to right). Results shown for MASA-QKVO (small transformer, S=6)}
    \label{fig:correlation_weights_6_qkvo}
    \vspace{-3mm}
\end{figure*}

\begin{figure*}[!h]
    \centering
    \begin{minipage}{0.23\textwidth}
        \centering
        \includegraphics[width=\linewidth]{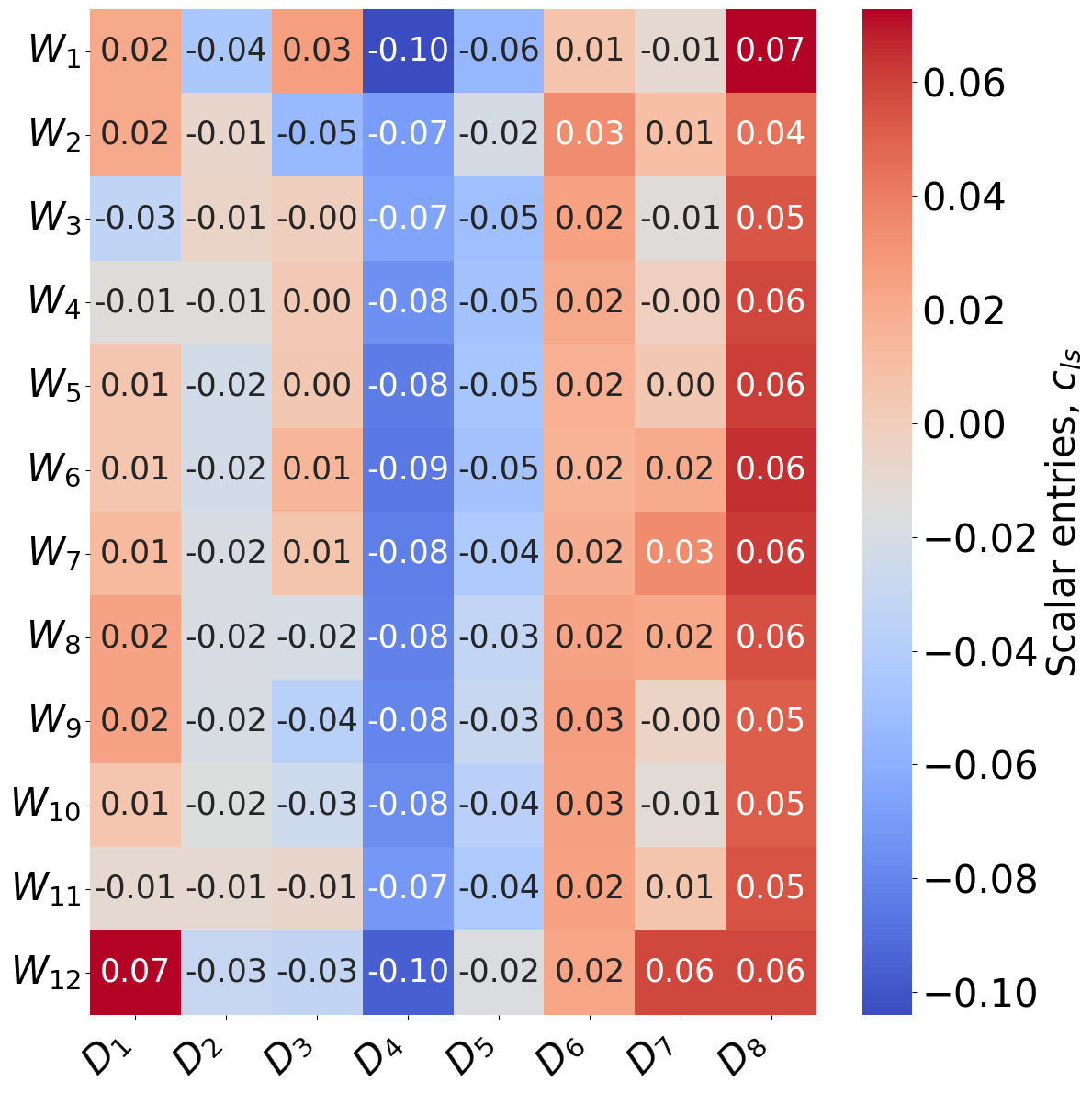}
        \caption*{(A)\quad $Query$}
    \end{minipage}
    \hfill
    \begin{minipage}{0.23\textwidth}
        \centering
        \includegraphics[width=\linewidth]{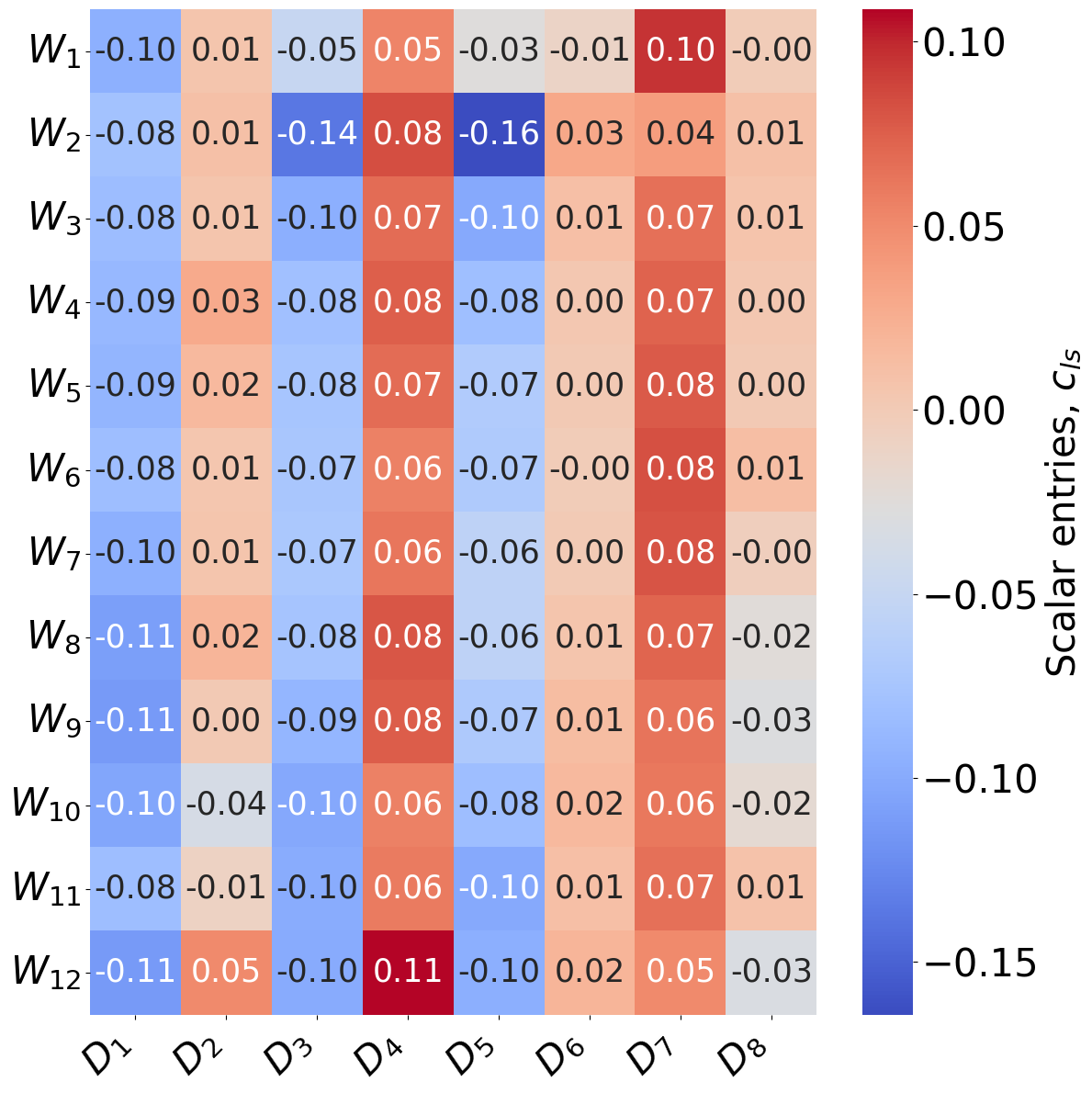}
        \caption*{(B)\quad $Key$}
    \end{minipage}
    \hfill
    \begin{minipage}{0.23\textwidth}
        \centering
        \includegraphics[width=\linewidth]{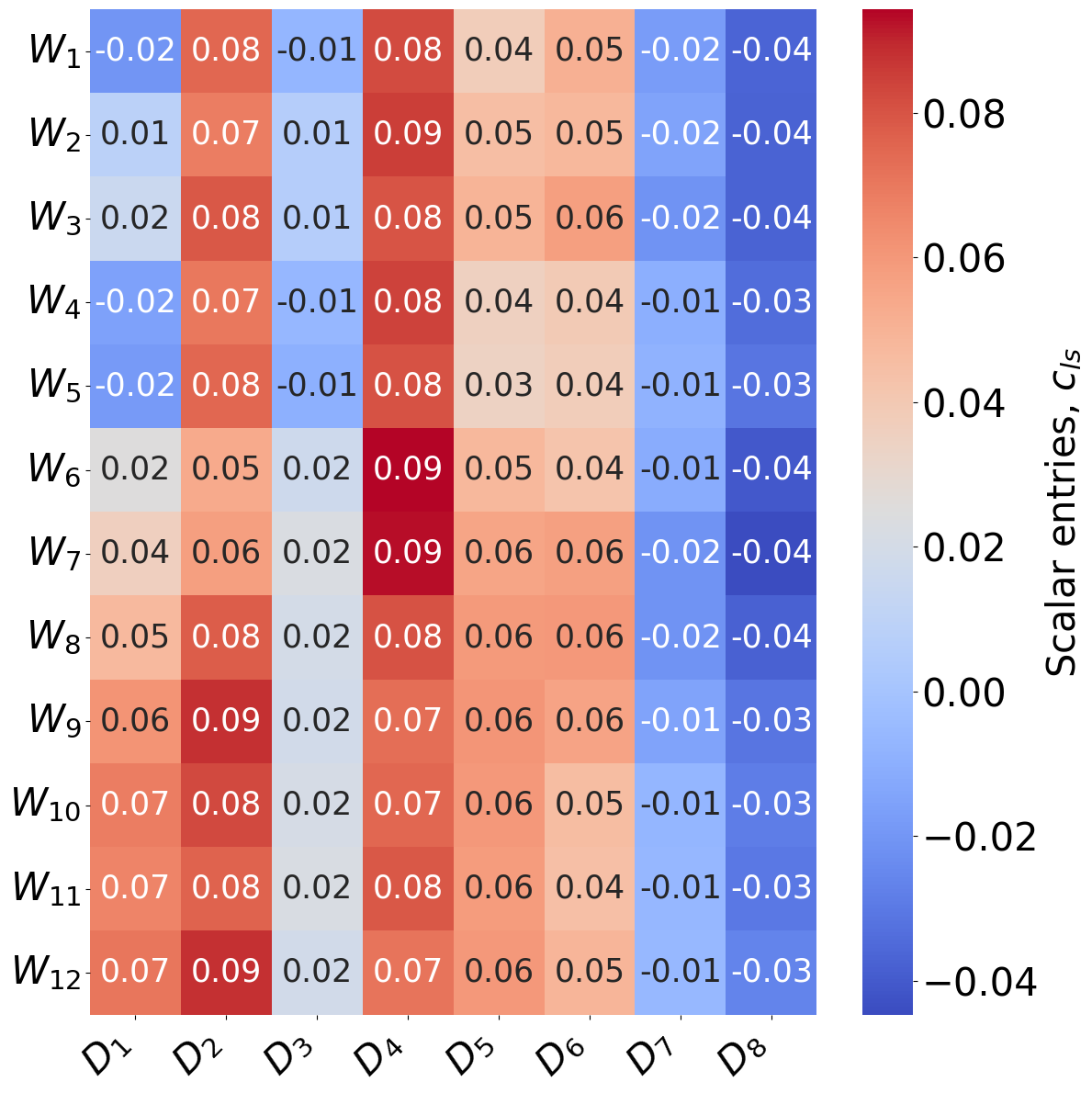}
        \caption*{(C)\quad $Value$}
    \end{minipage}
    \hfill
    \begin{minipage}{0.23\textwidth}
        \centering
        \includegraphics[width=\linewidth]{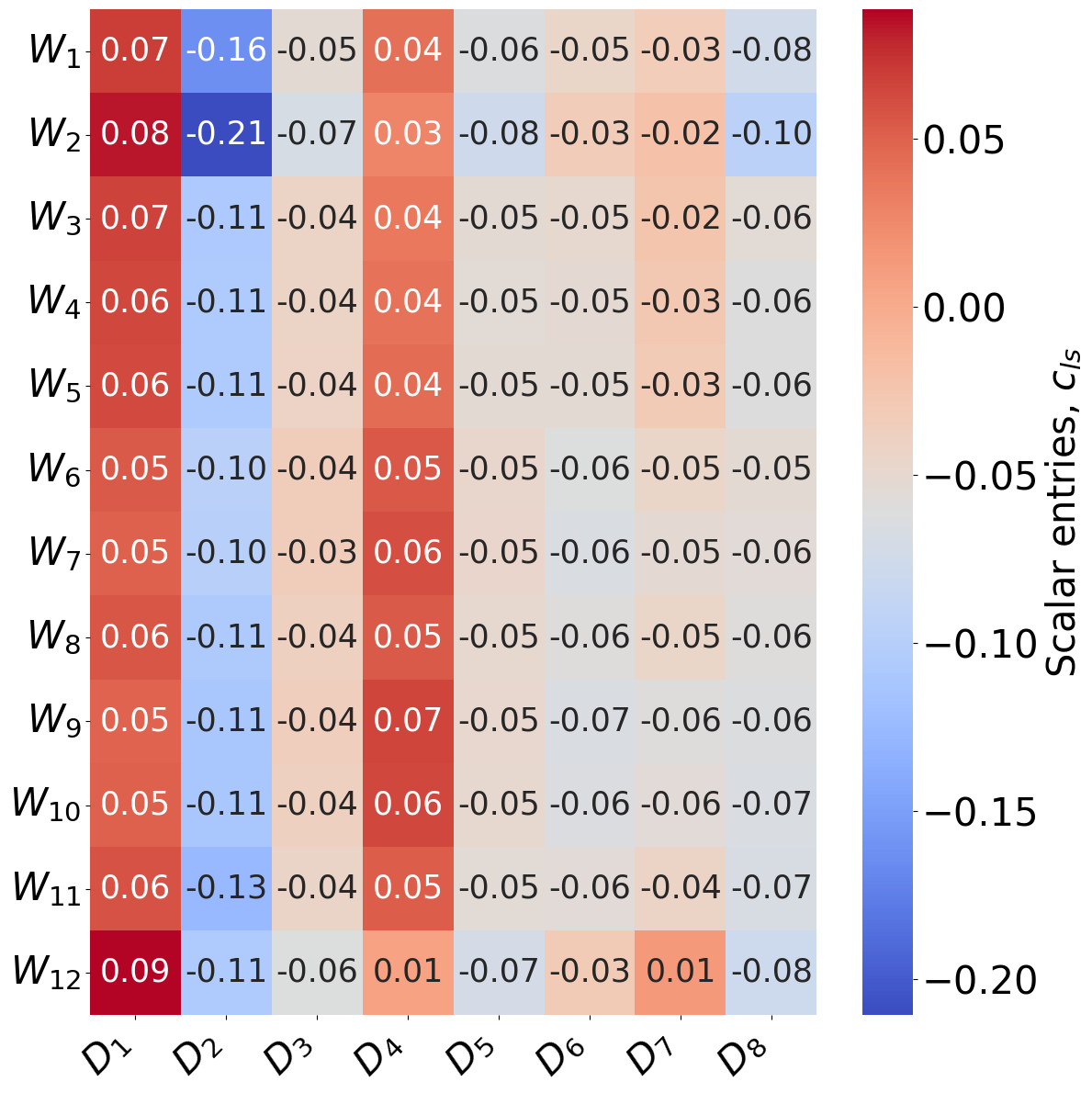}
        \caption*{(D)\quad $Out$}
    \end{minipage}
    \caption{Weight coefficients $\m C$ for each layer and atom in each dictionary for Q, K, V, and O projections (left to right). Results shown for MASA-QKVO (small transformer, S=8)}
    \label{fig:correlation_weights_8_qkvo}
    \vspace{-3mm}
\end{figure*}

In this section, we analyze the learned coefficients $\m C$ across layers of the small MASA-QKVO model for varying dictionary sizes $S = 2, 4, 6, \text{and } 8$. The results are visualized in Figures~\ref{fig:correlation_weights_2_qkvo} to~\ref{fig:correlation_weights_8_qkvo}. Each vertical line corresponds to a Transformer layer, and each row represents the contribution of a specific dictionary atom. These heatmaps reveal how different atoms are utilized across the network depth, highlighting patterns of specialization, redundancy, and layer-wise adaptivity in the shared weight reconstruction.

\subsection{Large Data-scale Training}
To assess the model's performance under large-scale training regimes, we compare MASA-QKV with a Transformer-S model trained on a RefinedWeb dataset 600 times larger than the model size. We use exactly the same training hyper-parameters (effective batch size, learning rate, tokenizer etc.) but increase the training data to 65B tokens. This allows us to evaluate whether the parameter efficiency of MASA-QKV preserves competitiveness as data scale increases, or if architectural compression becomes a bottleneck in data-rich settings. According to Table~\ref{tab:ablation_large_training}, our proposed method is negligibly behind in terms of average accuracy (-$\textbf{0.23}\%$), but outperforms in terms of WikiText perplexity. In conclusion, the results show that MASA-QKV maintains strong performance under large data training, despite having significantly fewer attention parameters.

\subsection{Extension to Vision Tasks}
To evaluate the generalization ability of our training-based MASA framework, we investigate its applicability to Vision Transformers (ViTs) on image \textbf{classification} and image \textbf{detection} tasks.

\paragraph{Image Classification} While the main paper focuses on CIFAR-100, we include additional results on CIFAR-10 (see Figure~\ref{cifar10_results}) and  on TinyImageNet (see Figure~\ref{tinyimagenet_results}) to further demonstrate the robustness and transferability of MASA across different data regimes. 

All models were trained for $100$ epochs on the CIFAR-10 dataset at $32×32$ image resolution, with evaluation on the test split. During training we  utilized a batch size of $512$, an initial learning rate of $0.001$ , and the ReduceLROnPlateau scheduler (factor: $0.2$ , patience: $3$ ). We employed the Adam optimizer and followed the official Vision Transformer (ViT) implementation. Experiments were conducted on an A100 (40GB) GPU.

\paragraph{Image Detection} For this experiment, we adopt the RT-DETR-Large architecture and follow the same training protocol as~\cite{zhao2024detrs}, with the exception that the input images are resized to a fixed resolution of $256 \times 256$. We also apply MASA on Q, K, V, and O projections of the attention modules in RT-DETR, where we learn two matrix atoms ($S=2$) shared across the six decoder layers of the original model. We implement the training pipeline using the Ultralytics library~\cite{ultralytics} and train both models (vanilla and MASA-QKVO) for 300 epochs with standard data augmentation and optimization settings. As shown in Table~\ref{tab:detr_results}, MASA achieves competitive performance compared to the full RT-DETR model, demonstrating that dictionary-based sharing incurs negligible accuracy loss.

\begin{table}[!h]
    \centering
\caption{Ablation study of MASA-QKVO applied to RT-DETR-Large architecture. Performance is evaluated on COCO val2017 with input resolution $256 \times 256$.}
    \resizebox{0.48\textwidth}{!}{%
    \renewcommand{\arraystretch}{1.2}
    \begin{tabular}{l|c|c|c}
        \hline 
        Model & Num. Weights, $S$& $mAP_{95}$ & $mAP_{50}$ \\ \hline 
        RT-DETR    & N/A & 0.368 & 0.530 \\
        MASA-QKVO  & 2   & 0.357 & 0.520 \\ \hline 
    \end{tabular}
    }
\label{tab:detr_results}
\vspace{-3mm}
\end{table}

\usetikzlibrary{positioning}

\begin{figure*}[!ht]
\centering
\begin{tikzpicture}
\begin{axis}[
    width=15cm, height=8cm,
    xlabel={Hidden Dimension \& Depth},
    ylabel={Accuracy (\%)},
    ymajorgrids=true,
    ylabel style={black},
    xmin=0.75, xmax=3.25,
    ymin=66, ymax=76,
    xtick={1,2,3},
    xticklabels={
        {64 \_ (12 vs MASA)},
        {256 \_ (12 vs MASA)},
        {256 \_ (24 vs MASA)}
    },
    legend style={at={(0.02,0.95)},anchor=north west,font=\fontsize{16}{17}\selectfont},
    tick label style={font=\small},
    label style={font=\bfseries\large},
    title={Vanilla Attention vs MASA on CIFAR10},
    title style={yshift=10pt,font=\bfseries\Large},
    ytick distance=2,
    grid=both,
]

\addplot [
    color=blue, mark=o, solid, thick
]
coordinates {
    (1,67.8)
    (2,71.66)
    (3,71.71)
};

\addplot [
    color=green!70!black, mark=triangle*, solid, thick
]
coordinates {
    (1,69.69)
    (2,73.43)
    (3,74.28)
};
  
\node at (axis cs:1,68.4) {\footnotesize 67.80};
\node at (axis cs:1,70.3) {\footnotesize 69.69};
\node at (axis cs:2,72.2) {\footnotesize 71.66};
\node at (axis cs:2,74.0) {\footnotesize 73.43};
\node at (axis cs:3,72.3) {\footnotesize 71.71};
\node at (axis cs:3,74.9) {\footnotesize 74.28};
\addlegendentry{Vanilla 12 layers}
\addlegendentry{MASA-QKVO (S=4)}
\addlegendimage{color=green!70!black, mark=triangle*, solid, thick}
\end{axis}

\begin{axis}[
    width=15cm, height=8cm,
    axis y line*=right,
    axis x line=none, 
    ylabel={Number of Parameters (M)},
    ylabel style={gray},
    ymin=0, ymax=11,
    xmin=0.75, xmax=3.25,
    ytick={0,2,4,6,8,10},
    yticklabel style={gray, /pgf/number format/fixed, /pgf/number format/precision=1},
    xtick=\empty,
    hide x axis,
    y tick label style={font=\small, color=gray}]

\addplot [
    color=blue!70!black, mark=o, dotted, thick
]
coordinates {
    (1,0.6)
    (2,4.77)
    (3,9.5195)
};

\addplot [
    color=green!70!black, mark=triangle*, dotted, thick
]
coordinates {
    (1,0.328)
    (2,4.304)
    (3,7.476)
};

\node [gray] at (axis cs:1,1.1) {\scriptsize 0.600};
\node [gray] at (axis cs:0.91,0.52) {\scriptsize 0.328};
\node [gray] at (axis cs:2,5.1) {\scriptsize 4.770};
\node [gray] at (axis cs:2,4.0) {\scriptsize 4.304};
\node [gray] at (axis cs:3,8.7) {\scriptsize 9.520};
\node [gray] at (axis cs:3,7.69) {\scriptsize 7.476};

\end{axis}
\end{tikzpicture}
\caption{Evaluation results of different ViT models trained from scratch on CIFAR10 train data, the blue solid plot represents the Top1-Accuracy of the vanilla attention models, the green solid plot represents the Top1-Accuracy of MASA, the dotted lines represent the parameter count of the full models respectivly.} 
\label{cifar10_results}
\end{figure*}
\begin{figure*}[!h]
    \centering
    \begin{tikzpicture}
    \begin{axis}[
        width=15cm, height=8cm,
        xlabel={Hidden Dimension \& Depth},
        ylabel={Accuracy (\%)},
        ymajorgrids=true,
        ylabel style={black},
        xmin=0.75, xmax=3.25,
        ymin=25, ymax=35,
        xtick={1,2,3},
        xticklabels={
            64\_ (12 vs MASA),
            256\_ (12 vs MASA),
            256\_ (24 vs MASA)
        },
        legend style={at={(0.02,0.95)},anchor=north west,font=\fontsize{16}{17}\selectfont},
        tick label style={font=\small},
        label style={font=\bfseries\large},
        title={Vanilla Attention vs MASA on TinyImageNet},
        title style={yshift=10pt,font=\bfseries\Large},
        ytick distance=2,
        grid=both,
    ]
    
    \addplot [
        color=blue, mark=o, solid, thick
    ]
    coordinates {
        (1,30.5)
        (2,31.91)
        (3,30.95)
    };
    
    \addplot [
        color=green!70!black, mark=triangle*, solid, thick
    ]
    coordinates {
        (1,29.8)
        (2,31.38)
        (3,33.46)
    };
      
    \node at (axis cs:1,31.4) {\footnotesize 30.5};
    \node at (axis cs:1,30.3) {\footnotesize 29.8};
    \node at (axis cs:2,32.8) {\footnotesize 31.91};
    \node at (axis cs:2,31.5) {\footnotesize 31.38};
    \node at (axis cs:3,31.3) {\footnotesize 30.95};
    \node at (axis cs:3,33.2) {\footnotesize 33.46};
    \addlegendentry{Vanilla 12 layers}
    \addlegendentry{MASA-QKVO (S=4)}
    \addlegendimage{color=green!70!black, mark=triangle*, solid, thick}
\end{axis}
    
    \begin{axis}[
        width=15cm, height=8cm,
        axis y line*=right,
        axis x line=none, 
        ylabel={Number of Parameters (M)},
        ylabel style={gray},
        ymin=0, ymax=11,
        xmin=0.75, xmax=3.25,
        ytick={0,2,4,6,8,10},
        yticklabel style={gray, /pgf/number format/fixed, /pgf/number format/precision=1},
        xtick=\empty,
        hide x axis,
        y tick label style={font=\small, color=gray}]

    \addplot [
        color=blue!70!black, mark=o, dotted, thick
    ]
    coordinates {
        (1,0.6)
        (2,4.77)
        (3,9.5195)
    };
    
    \addplot [
        color=green!70!black, mark=triangle*, dotted, thick
    ]
    coordinates {
        (1,0.328)
        (2,4.304)
        (3,7.476)
    };
    
    \node [gray] at (axis cs:1,1.1) {\scriptsize 0.600};
    \node [gray] at (axis cs:0.91,0.47) {\scriptsize 0.328};
    \node [gray] at (axis cs:2,5.1) {\scriptsize 4.770};
    \node [gray] at (axis cs:2,4.0) {\scriptsize 4.304};
    \node [gray] at (axis cs:3,10) {\scriptsize 9.520};
    \node [gray] at (axis cs:3,7.69) {\scriptsize 7.476};

    \end{axis}
    \end{tikzpicture}
    \caption{Evaluation results of different ViT models trained from scratch on TinyImageNet train data, the blue solid plot represents the Top1-Accuracy of the vanilla attention models, the green solid plot represents the Top1-Accuracy of MASA, the dotted lines represent the parameter count of the full models respectivly.} 
    \label{tinyimagenet_results}
\end{figure*}

\end{document}